\documentclass[11pt]{article}

\usepackage[a4paper,margin=1in]{geometry}
\usepackage[utf8]{inputenc}
\usepackage[T1]{fontenc}
\usepackage{setspace}
\usepackage{parskip}
\usepackage{graphicx}
\usepackage{xcolor}
\usepackage{titlesec}
\usepackage{fancyhdr}
\usepackage{hyperref}
\usepackage{lmodern}
\usepackage{tcolorbox}
\usepackage{csquotes}
\usepackage{lmodern}
\usepackage{graphicx}
\usepackage{xspace}
\usepackage[backend=biber,style=numeric,sorting=nyt]{biblatex}
\usepackage{microtype}
\usepackage{url}
\usepackage{booktabs}
\usepackage{subfigure}

\usepackage{enumitem}
\newenvironment{itemize*}%
 {\leftmargini=20pt\begin{itemize}%
  \setlength{\itemsep}{3pt}%
  \setlength{\parskip}{0pt}%
  }%
 {\end{itemize}} 
\newenvironment{enumerate*}%
 {\begin{enumerate}%
  \setlength{\itemsep}{0pt}%
  \setlength{\parskip}{0pt}}%
 {\end{enumerate}}

\usepackage{amsmath}
\usepackage{cleveref}
\usepackage{listings}
\usepackage{multirow}
\usepackage{makecell}
\usepackage{subcaption} 
\usepackage{cuted}
\usepackage{wrapfig}

\usepackage{array}       
\usepackage{caption}     
\usepackage{amssymb}
\usepackage{dsfont}

\definecolor{midnightgreen}{rgb}{0.0, 0.29, 0.33}
\definecolor{deepgreen}{HTML}{0aa344}
\definecolor{deeppurple}{HTML}{7030a0}
\definecolor{deepblue}{HTML}{171d91}
\definecolor{brown}{HTML}{843c0c}
\definecolor{shadered}{HTML}{ffe5e5}
\definecolor{shadegreen}{HTML}{e5f7ed}
\definecolor{msftBlack}{RGB}{0,0,0}
\definecolor{lightred}{RGB}{255,163,163}
\definecolor{deepred}{RGB}{146,0,0}

\usepackage{pifont}

\NewDocumentCommand{\heng}
{ mO{} }{\textcolor{red}{\textsuperscript{\textit{Heng}}\textsf{\textbf{\small[#1]}}}}
\NewDocumentCommand{\cheng}
{ mO{} }{\textcolor{orange}{\textsuperscript{\textit{Cheng}}\textsf{\textbf{\small[#1]}}}}

\definecolor{sfblue}{HTML}{00A1E0}      
\definecolor{sfnavy}{HTML}{032D60}      
\definecolor{sfgray}{HTML}{706E6B}      
\definecolor{sflightblue}{HTML}{EAF5FC} 

\hypersetup{
  colorlinks=true,
  linkcolor=sfblue,
  citecolor=sfnavy,
  urlcolor=sfblue
}

\fancypagestyle{plain}{
  \fancyhf{} 
  \fancyfoot[C]{\textcolor{sfgray}{\thepage}} 
  
}

\makeatletter
\renewcommand{\maketitle}{
  \thispagestyle{plain} 
  \noindent
  \vspace*{-10pt}
  \noindent\raisebox{0pt}[0pt][0pt]{\includegraphics[height=1cm]{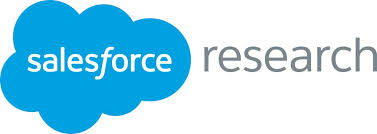}}

  \vspace{-5pt}
  \color{sfgray}\rule{\linewidth}{0.6pt}

  \vspace{10pt}
  {\Huge\bfseries\color{sfnavy} \@title \par}
  \vspace{0.5em}
  {\large \@author \par}
  \vspace{0.2em}
}
\makeatother

\titleformat{\section}{\large\bfseries\color{sfnavy}}{\thesection}{1em}{}
\titleformat{\subsection}{\normalsize\bfseries\color{sfnavy}}{\thesubsection}{1em}{}

\title{\huge xRouter: Training Cost-Aware LLMs Orchestration System via Reinforcement Learning}
\author{
\textbf{Cheng Qian}$^{1,2*}$, 
\textbf{Zuxin Liu}$^{1*\diamondsuit}$, 
\textbf{Shirley Kokane}$^{1\dagger\diamondsuit}$, 
\textbf{Akshara Prabhakar}$^{1\dagger}$, 
\textbf{Jielin Qiu}$^{1\dagger}$, 
\textbf{Haolin Chen}$^{1}$, 
\textbf{Zhiwei Liu}$^{1}$, 
\textbf{Heng Ji}$^{2}$, 
\textbf{Weiran Yao}$^{1}$, 
\textbf{Shelby Heinecke}$^{1}$, 
\textbf{Silvio Savarese}$^{1}$, 
\textbf{Caiming Xiong}$^{1}$, 
\textbf{Huan Wang}$^{1}$\\[7pt]
$^{1}$Salesforce AI Research \quad $^{2}$University of Illinois Urbana-Champaign\\
\small $^{*}$Co-first authors \quad $^{\dagger}$Core contributors \quad $^{\diamondsuit}$Work done while at Salesforce AI Research
}

\date{\today}

\newcommand{\xrouter}{\textsc{xRouter}}
\newcommand{\github}{\raisebox{-1.5pt}{\includegraphics[height=1.05em]{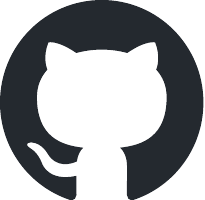}}\xspace}

\begin{document}
\maketitle
\begin{tcolorbox}[colback=sflightblue!60,
                  colframe=sfblue,
                  boxrule=0.6pt,
                  arc=2mm,
                  left=6pt,right=6pt,top=6pt,bottom=6pt]
\textbf{Abstract.} Modern LLM deployments confront a widening cost–performance spectrum: premium models deliver strong reasoning but are expensive, while lightweight models are economical yet brittle on complex tasks. Static escalation rules and keyword heuristics under-utilize this spectrum and fail to adapt across task types. We present \textbf{{\xrouter}}, a tool-calling–based routing system in which a learned router can either answer directly or invoke one or more external models. The router is trained end-to-end with reinforcement learning using an explicit, cost-aware reward that encodes cost--performance trade-offs, eliminating the need for hand-engineered routing rules. Our implementation encompasses the full reinforcement learning framework, including reward and cost accounting, as well as the deployment and evaluation pipelines. Across diverse benchmarks, {\xrouter} achieves strong cost–performance trade-offs (e.g., substantial cost reductions at comparable task completion rates) and provides empirical insights into what reliably helps learned routing and what does not, ranging from model trainability to the difficulty of eliciting sophisticated orchestration behaviors in small open models. We hope these findings and our open implementation will serve as a practical substrate for advancing learned, cost-aware LLM orchestration.

\vspace{1em}
\begin{center}
\begin{tabular}{rll}
    \github & \textbf{\small{Code}} &
    \href{https://github.com/SalesforceAIResearch/xRouter}{\small{\texttt{https://github.com/SalesforceAIResearch/xRouter}}}
\end{tabular}
\end{center}
\end{tcolorbox}
\color{black}
\section{Introduction}

The proliferation of large language models (LLMs) has turned single-model inference into a multi-model selection problem \cite{ding2024hybrid,stripelis2024tensoropera,zhang2025router,chuang2025learning,woisetschlager2025dynamically,dekoninck2025unifiedapproachroutingcascading,ong2025routellm}. In practice, queries arrive with unpredictable difficulty and domain variation; no single model optimally spans this space given the steep gradient in capability \cite{shnitzer2023largelanguagemodelrouting}, latency, and price \cite{chen2023frugalgptuselargelanguage}. Naïve strategies, such as ``using expensive/bigger models for hard queries, the cheaper ones for easy queries'' \cite{wang2023tabi,vsakota2024fly,song2025irt}, are brittle; and hand-crafted routing trees \cite{ding2024hybrid} rarely transfer across domains, providers or evolving APIs. The result is a persistent gap between what deployments pay for and what they need on a per-request basis.

We approach routing as decision-making under uncertainty with explicit economic constraints. Our system, {\xrouter}, trains a tool-calling \emph{router} that can either respond directly or delegate to external models (and, when beneficial, coordinate multiple calls). Rather than encoding escalation logic by hand, we formulate routing as a reinforcement-learning problem with a cost-aware objective that rewards successful task completion while penalizing unnecessary spend. Concretely, the reward is success-contingent and cost-sensitive (more intuitively, \emph{no success, no reward; on success, cheaper is better}), which encourages judicious use of premium models without deterring exploration when difficulty warrants it.

Building a practical learned router requires more than a clever reward \cite{lu2023routing,zhang2025router}. We implement the end-to-end pipeline needed for training and evaluation: data preprocessing that exposes difficulty diversity, explicit and auditable cost accounting across model calls, and an RL training framework that scales to realistic orchestration settings. We then conduct systematic experiments to probe what works and what does not in learned routing: are raw instruct models trainable as routers, how routing behaviors evolve over the training process, when orchestration beyond single-model selection actually emerges, and where it stubbornly does not. These investigations also surfaced limitations, such as training instabilities for certain architectures and the non-triviality of eliciting sophisticated multi-step orchestration in small open models, for which we document in detail to guide the future research.

To summarize, our contributions are fourfold:
\begin{itemize}[topsep=-3pt, partopsep=-3pt, leftmargin=*, itemsep=-1.5pt]
\item We design a tool-calling based routing system that lets the router answer on its own or call external models flexibly, enabling direct answers when their ability suffices and delegation when necessary.
\item We deliver a complete reward/cost accounting and RL training framework demonstrating that routing behavior can be learned with explicit cost--performance trade-offs, rather than hand-engineered rules. 
\item Through initial explorations, we characterize practical constraints and failure modes of learned routing, from model trainability to the surprising difficulty of getting sophisticated orchestration behaviors to emerge naturally on small open models. From these observations, we also distill actionable insights for future systems. 
\item To catalyze progress, we release our complete implementation and evaluation framework so others can extend our results.
\end{itemize}
Empirically, {\xrouter} attains favorable operating points on the cost--performance Pareto frontier while making transparent the scenarios where learned routing pays off and where a single strong model may remain simpler and more predictable. Together, these contributions position {\xrouter} as a first step toward principled, economically grounded orchestration of LLMs, advancing the path from ad-hoc heuristics to generalizable and cost-efficient deployment.

\section{Related Work}
\label{sec:related}

Our work bridges several active research areas in machine learning and natural language processing. We situate {\xrouter} within the broader contexts of model routing systems, cost-aware optimization, multi-model orchestration, reinforcement learning for adaptive model control, and economic perspectives on AI deployment.

\subsection{Model Routing and Mixture of Experts}

The idea of intelligent model selection originates in ensemble learning and mixture of experts (MoE) architectures \cite{jacobs1991adaptive,jordan1994hierarchical}. Classical MoE systems learn to assign inputs to specialized subnetworks, with a gating network determining the contribution of each expert based on input features. Modern large language models reinterpret this concept at scale—different models or variants act as experts, and routing decisions determine which are activated for a given input.

Recent advances have explored routing efficiency from multiple perspectives. For instance, \cite{jiang2024mixtral} shows that activating only the relevant subset of parameters can yield competitive accuracy with lower computation. LoraHub \cite{huang2024lorahub}, LoRA Soups \cite{prabhakar2024lorasoupsmergingloras} explore adaptive model composition but focuses primarily on combining LoRA adapters rather than orchestrating fundamentally different model architectures. However, such approaches generally emphasize architectural sparsity or static escalation strategies rather than \emph{learned} routing behavior that adapts dynamically to cost or task context. Our work extends this line of thought by explicitly learning routing policies across heterogeneous models with varying costs and capabilities.

\subsection{Cost-Aware Machine Learning}

Cost-aware optimization has become increasingly central as practitioners aim to balance accuracy and efficiency. Early research targeted resource-constrained environments—optimizing for memory and latency on mobile and embedded devices \cite{kang2017neurosurgeon,han2015deep}. Subsequent work generalized these principles to cloud-based inference, emphasizing server efficiency and multi-tenant fairness. For example, \cite{crankshaw2017clipper} proposed adaptive batching and caching mechanisms to reduce inference costs, while \cite{kumar2018resource} developed resource allocation algorithms for shared ML systems. \cite{kapoor2024ai} examine inference costs and show that independent sampling outperforms the methods proposed by \cite{shinn2023reflexion} under equivalent sampling budgets. Yet, most of these approaches treat cost merely as a constraint, rather than as an explicit optimization objective jointly considered with performance.

Recent work has begun to view cost itself as a central axis of optimization. FrugalGPT \cite{chen2023frugalgptuselargelanguage} introduced a cascade-based approach that escalates queries from lightweight to more capable models, while FORC \cite{vsakota2024fly} extended this concept to cost-efficient selection among multiple LLMs. TO-Router \cite{stripelis2024tensoropera} unifies domain-specific experts through a single interface that dynamically dispatches based on task semantics. Router-R1 \cite{zhang2025router} pushes this further by instantiating the router as a reasoning LLM capable of alternating between ``think'' and ``route'' actions, illustrating the promise of learned decision policies for orchestration. Similarly, RouteLLM \cite{ong2025routellm} leverages human preference signals to optimize routing between strong and weak models, achieving adaptive trade-offs between cost and quality.

The rise of paid API-based language model services has amplified the need for cost-sensitive optimization. \cite{wang2023cost} and \cite{zhang2023cost} analyze cost-performance trade-offs for prompting and fine-tuning, while \cite{qian2025smart} and \cite{wang2025acting} address inefficiencies such as tool overuse through reward shaping and cost-aware reinforcement learning. Yet, these efforts focus on optimizing single-model usage rather than orchestrating multiple models under explicit cost--performance trade-offs, a gap that {\xrouter} aims to address.

\subsection{Multi-Model Orchestration and Ensemble Systems}

Multi-model orchestration extends ensemble learning by coordinating diverse models to exploit complementary strengths. Recent approaches have shown how structured collaboration among models can improve reasoning, factuality, and generalization. For example, \cite{wang2023selfconsistency} uses multiple model calls for consistency-based voting, while \cite{yao2023tree} proposes tree-of-thought reasoning that benefits from specialization across reasoning steps. Likewise, \cite{zheng2023judging} employs stronger models to evaluate and select responses from weaker ones, establishing early frameworks for response selection and validation.

Tool-using agents further broaden this paradigm. \cite{schick2023toolformer} and \cite{qin2023tool} demonstrate how LLMs can learn to invoke external APIs, creating opportunities for dynamic multi-model pipelines. \cite{paranjape2023art} explores autonomous reasoning and tool-use, while \cite{qian2023creator, qian2024toolink} investigate the automatic creation and selection of specialized tools. Despite these advances, orchestration in current systems is largely heuristic or rule-based, lacking mechanisms to learn optimal coordination policies under cost constraints. The integration of reinforcement learning into multi-model orchestration thus remains an open and promising direction.

\subsection{Reinforcement Learning for Language Models}

Reinforcement learning (RL) has proven to be a powerful framework for aligning and enhancing large language models. Techniques such as Reinforcement Learning from Human Feedback (RLHF) \cite{ouyang2022training,christiano2017deep} established that language models can be optimized beyond supervised objectives by learning from preference-based reward signals. Building on this foundation, RL has been applied to specific language tasks such as summarization \cite{stiennon2020learning}, factual QA \cite{nakano2021webgpt}, mathematical reasoning \cite{cobbe2021training}, user-centric personalization \cite{qian2025userrl}, and code generation \cite{le2022coderl}. Beyond these, RL has facilitated advances in tool-use \cite{qian2025toolrl}, reward modeling \cite{chen2025rm}, and verification \cite{he2025veri}, underscoring its versatility and generalizability.

More recently, RL has been used to enable adaptive and meta-level behaviors in LLMs. \cite{carta2024grounding} studies RL for grounding models in interactive environments, while \cite{yao2023retroformer} applies RL to optimize retrieval strategies. These works highlight how RL can enable adaptive control within single-model systems. However, its application to cross-model routing and multi-model orchestration, particularly when cost efficiency is a factor, remains underexplored. Our work extends this frontier by leveraging RL to learn routing policies that explicitly balance performance gains with inference cost.

\subsection{Economic Models in AI Systems}

The economic dimensions of AI system design have gained importance as models grow larger and more expensive to train and deploy. Foundational analyses by \cite{strubell2020energy} and \cite{patterson2021carbon} quantify the environmental and economic impact of large-scale training. Subsequent work has developed theoretical and practical frameworks to optimize compute utilization. \cite{brown2020language} and \cite{kaplan2020scaling} analyze scaling laws and their cost implications, while \cite{hoffmann2022training} proposes optimal compute allocation strategies that jointly consider training and inference efficiency.

While these studies provide valuable macro-level insights into model economics, they primarily address training dynamics or static scaling behavior. In contrast, the economic optimization of \emph{operational} inference, particularly in multi-model systems where cost structures are heterogeneous and dynamic, remains relatively unexplored. Our work contributes to this growing area by introducing an explicit cost--performance trade-off framework for real-time multi-model routing and orchestration.

\section{Preliminaries}

\paragraph{Problem setting.}
We study \emph{learned routing} for multi-model LLM systems under explicit economic constraints.
At inference time, inputs arrive with heterogeneous difficulty and domain shift, while the system has access to a catalog of models that vary in capability, latency, and price.
Rather than fixing a rule-based escalation tree, we cast routing as sequential decision-making: a router observes the user query and conversational context, reasons over the available model catalog, and decides either to answer directly or to delegate one or more calls to external models.%
\footnote{Throughout, we use ``model'' broadly to include vendor APIs and local models.}
This view makes room for genuine cost–performance trade-offs: spending more is sometimes warranted, but the router should avoid unnecessary calls when a cheaper path suffices.

\paragraph{Cost model and accounting.}
Each routed interaction incurs a cost that aggregates per-call token prices (or equivalent metering) across all external invocations, plus fixed overheads (e.g., retrieval or formatting). We track costs at two granularities: per turn (single routing decision) and per episode (a short conversation or task), since routing that is locally frugal can still lead to globally expensive workflows. For stability across datasets and budgets, we optionally normalize costs by a configurable per-turn cap so rewards are comparable across training batches and evaluation suites. All accounting is auditable: we log selected models, prompts, token counts, and wall-clock latencies alongside success indicators.

\paragraph{Success metrics and trade-offs.}
Following practical deployments, we evaluate both \emph{task success} (e.g., pass@1, exact match, rubric-based correctness) and \emph{operational cost} (token-denominated spend or budget proxy). This highlights where learned routing yields clear wins (maintaining success at lower cost) and where a single strong model remains simpler and more predictable. These preliminaries set up the learning objective and system we introduce next.

\section{Methodology}

\subsection{System Overview}

\xrouter{} is designed to balance cost-efficiency and model capability through learned routing and orchestration. It comprises two core components:

\emph{(1) The router agent}, a fine-tuned language model (e.g., Qwen2.5-7B-Instruct), observes the user query and conversational context and produces either a direct answer or a tool call. These tool calls are lightweight schema that specify which external models to invoke, along with configuration hints (such as prompt style or temperature).

\emph{(2) The orchestration engine}, a model-agnostic execution layer, receives these tool calls, issues requests to the selected models (via APIs or local inference endpoints), and gathers responses. It supports both simple one-shot routing (e.g., ``pick one model and return'') and light orchestration (e.g., ``query multiple models and fuse'').

In practice, the engine handles infrastructure complexity, including timeouts, retries, caching, response validation, logging, so the router can focus on the routing policy alone. This disentanglement ensures extensibility and abstracts model-specific details from routing logic.

\subsection{Learning Objective}

We frame routing as a reinforcement learning problem with a reward function that jointly encodes success and cost-awareness. For each turn (or episode), we define the reward as:

\[
R_{\text{final}} \;=\; R_{\text{binary}} \times \left(K - \lambda \, C \right),
\]

where $R_{\text{binary}} \in \{0, 1\}$ indicates task success, $C$ is the total cost of all model invocations (e.g., normalized token-based spend), $K$ is a fixed success bonus, and $\lambda$ controls the strength of the cost penalty.

This reward is intentionally gated: no success means zero reward regardless of cost. On success, lower-cost strategies are preferred. This incentivizes the router to experiment with cheap paths (including answering directly), but escalate to more expensive models when needed. For multi-turn dialogs, we also compute episode-level cost and success to discourage greedy, short-term savings that increase downstream difficulty.

We train the policy using a GRPO-style algorithm: DAPO~\cite{yu2025dapo}, but the algorithm itself is not our contribution. Rather, our focus is the cost-sensitive reward formulation and the infrastructure for scalable, reproducible training.

\subsection{Training Data and Signal Shaping}

To train a useful router, the dataset must expose genuine diversity in query difficulty and model behavior. We construct our training set by sampling tasks from the Reasoning360 benchmark~\cite{cheng2025revisiting}, which includes reasoning-intensive, multi-format question sets. Each sample is annotated with a difficulty estimate using the pass@$k$ rate of a strong model (Qwen3-32B). We stratify the data into easy, medium, and hard tiers, and sample proportionally across these buckets to promote robust generalization.

We also augment the dataset with simpler queries, including chit-chat, small retrieval, and factual lookups, to ensure the router encounters scenarios where it can safely respond on its own. Without this, policies may over-rely on delegation and miss opportunities to reduce cost.

The training corpus includes the description and costs of multiple models from different capability tiers (e.g., budget vs. premium), allowing the router to observe quality differences and their associated prices. To avoid memorization, we periodically refresh the model catalog and simulate cost perturbations. Failed attempts (e.g., wrong answers from expensive models, redundant calls, or missed self-answer opportunities) still incur cost and result in zero reward, thus reinforcing the need for efficient routing behavior.

This setup produces a clean signal: \textbf{correctness gates reward, while cost shapes routing decisions}. The router must learn to balance the trade-off between performance and expenditure without relying on brittle heuristics.

\subsection{Implementation Details}

The router emits structured tool calls in a minimal OpenAI-compatible function-calling format, allowing easy extension to new models, prompt templates, or sampling strategies. Each tool call contains a selected model name, an optional system prompt override, and sampling parameters (e.g., temperature). These are passed to the orchestration engine, which is stateless and configurable at runtime.

To keep latency and cost under control, we disable fan-out (number of models called simultaneously) and call depth (number of cascaded calls in multi-turn or self-refinement settings). We also debounce retries to prevent runaway spending under uncertain model behavior.

We evaluate as in training, reporting cost and success per episode. In early tests, we also explored dedicated multi-turn modeling heads, but found little empirical lift compared to our simpler success-gated objective, especially for short or loosely structured conversations. As such, we keep the turn-level router lightweight and focus on rewards and data design to capture context sensitivity.

\paragraph{Outcome.}
The result is a system that learns to answer directly when safe, escalate when necessary, and balance performance with budget in a principled way. The router can be tuned via two intuitive hyperparameters ($K$ and $\lambda$), and adapts to different environments with minimal reconfiguration. This setup enables us to study the emergence (and failure) of sophisticated orchestration behaviors under realistic, cost-constrained conditions.

\section{Experimental Evaluation}
\label{sec:experiment}

This section presents a comprehensive experimental evaluation of {\xrouter} across diverse tasks and deployment scenarios. We evaluate both the effectiveness of our cost-aware routing approach and the practical performance characteristics essential for production deployment.

\subsection{Experimental Setup}

\paragraph{Overall Settings.}
Using the reward shaping strategy described above, we implement our algorithm on the \texttt{Qwen2.5-7B-Instruct} model to train the router component. Specifically, we limit the maximum interaction turns to three and employ the DAPO algorithm within the \textsc{Verl} framework for optimization. To study the effect of cost–performance trade-offs, we train three router variants with different $\lambda$ values, which are further analyzed in the results section. The resulting models serve as the core routers within the {\xrouter} system, demonstrating strong decision-making capabilities in balancing computational cost and task performance.

We evaluate {\xrouter} across a diverse suite of benchmarks covering multiple reasoning domains to assess its generalization and robustness. These include mathematical reasoning tasks such as \textit{AIME}, code generation benchmarks such as \textit{Codeforces}, and logical reasoning evaluations such as \textit{GPQA}. Please refer to the main results table below for the complete list of tasks. For all benchmarks, we follow the task-specific evaluation protocols and report both performance and cost metrics accordingly.

For all experiments involving the router, the following pool of models is available for task offloading:
\textit{{GPT-5, GPT-5-mini, GPT-5-nano, GPT-4o, GPT-4.1, o3, o3-Pro, o4-mini, GPT-OSS-120B, GPT-OSS-20B, Gemini-2.5-Pro, Gemini-2.5-Flash-Lite}}.
We further perform an ablation study to evaluate how varying the composition of this model pool affects routing performance in later analysis.

\paragraph{Baseline Systems.}
We compare {\xrouter} against several baseline approaches to demonstrate the effectiveness of our cost-aware routing methodology:
\begin{itemize}[topsep=-1pt, partopsep=-3pt, leftmargin=*, itemsep=0pt]
    \item \textbf{Single-Model Baselines:} Instead of employing the routing system, we use a single model as the baseline for evaluation. This setup allows us to assess the benefits of our routing framework compared to directly using individual models without any routing mechanism.
    \item \textbf{Static Routing Baselines:} Instead of using our trained router model, we use existing models to perform routing decisions directly during inference. This comparison highlights the effectiveness of our trained router in learning adaptive routing strategies over static or heuristic-based approaches.
\end{itemize}

\paragraph{Evaluation Metrics.}
We employ a comprehensive set of metrics to evaluate both the performance and cost effectiveness of our approach:
\begin{itemize}[topsep=-1pt, partopsep=-3pt, leftmargin=*, itemsep=0pt]
    \item \textbf{Performance Metric:} We measure domain-specific task performance. Since each task uses its own evaluation metric (e.g., accuracy, F1, etc.), the absolute performance scores are not directly comparable across tasks.
    \item \textbf{Cost Metric:} We average cost per question incurred during the evaluation of benchmarks. Because different tasks vary in their complexity (e.g., number of turns, expected answer length, etc.), the total evaluation cost across different tasks is also not directly comparable.
\end{itemize}

\subsection{Main Results}

\begin{table*}[!t]
\begin{center}
\small
\renewcommand{\arraystretch}{1.2}
\tabcolsep=0.005\linewidth
\resizebox{\linewidth}{!}{
\begin{tabular}{
l
cc cc cc cc cc cc cc cc
}
\toprule
\multirow{2}{*}{\textbf{Model}} &
\multicolumn{2}{c}{\textbf{Minerva}} &
\multicolumn{2}{c}{\textbf{MATH-500}} &
\multicolumn{2}{c}{\textbf{Olympiad Bench}} &
\multicolumn{2}{c}{\textbf{AIME-24}} &
\multicolumn{2}{c}{\textbf{AMC-23}} &
\multicolumn{2}{c}{\textbf{Codeforces}} &
\multicolumn{2}{c}{\textbf{Code-Contests}} &
\multicolumn{2}{c}{\textbf{Human-EvalPlus}} \\
\cmidrule(lr){2-3}
\cmidrule(lr){4-5}
\cmidrule(lr){6-7}
\cmidrule(lr){8-9}
\cmidrule(lr){10-11}
\cmidrule(lr){12-13}
\cmidrule(lr){14-15}
\cmidrule(lr){16-17}
 & \textbf{Acc.} & \textbf{Cost} & 
   \textbf{Acc.} & \textbf{Cost} & 
   \textbf{Acc.} & \textbf{Cost} & 
   \textbf{Acc.} & \textbf{Cost} & 
   \textbf{Acc.} & \textbf{Cost} & 
   \textbf{Acc.} & \textbf{Cost} & 
   \textbf{Acc.} & \textbf{Cost} & 
   \textbf{Acc.} & \textbf{Cost} \\
\midrule
\multicolumn{17}{c}{\textit{Existing Model as Router}} \\
\midrule
GPT-4o & 0.14 & 0.016660 & 0.67 & 0.021832 & 0.46 & 0.023937 & 0.22 & 0.027271 & 0.29 & 0.020637 & 0.51 & 0.016514 & \underline{0.64} & 0.018202 & 0.92 & 0.024512 \\
GPT-5-nano & 0.22 & 0.001439 & 0.94 & 0.001311 & 0.82 & 0.001808 & 0.77 & 0.004069 & 0.87 & 0.003984 & 0.49 & 0.005172 & \underline{0.64} & 0.006316 & \underline{0.94} & 0.001026 \\
GPT-5-mini & 0.17 & 0.004276 & \underline{0.95} & 0.004006 & \underline{0.83} & 0.005172 & \textbf{0.93} & 0.010321 & \textbf{0.96} & 0.006220 & \textbf{0.55} & 0.015665 & 0.60 & 0.013026 & 0.90 & 0.016481 \\
GPT-5 & 0.22 & 0.023585 & \textbf{0.97} & 0.021243 & \textbf{0.84} & 0.028735 & \underline{0.89} & 0.057474 & \underline{0.94} & 0.035875 & 0.51 & 0.015580 & \underline{0.64} & 0.014523 & \textbf{0.95} & 0.021353 \\
\midrule
GPT-OSS-20B & 0.24 & 0.000594 & 0.87 & 0.000429 & 0.67 & 0.000595 & 0.41 & 0.000965 & 0.73 & 0.000588 & 0.41 & 0.016793 & 0.51 & 0.016350 & 0.83 & 0.000357 \\
Qwen2.5-7B-Instruct & 0.23 & 0.007638 & 0.74 & 0.004634 & 0.57 & 0.004919 & 0.19 & 0.008521 & 0.48 & 0.003836 & 0.28 & 0.016507 & 0.18 & 0.027208 & 0.45 & 0.002276 \\
Qwen3-8B-Instruct & \textbf{0.42} & 0.006753 & 0.90 & 0.003984 & 0.73 & 0.005594 & 0.59 & 0.020233 & 0.77 & 0.012196 & 0.44 & 0.033924 & 0.49 & 0.038872 & 0.86 & 0.004950 \\
\midrule
\multicolumn{17}{c}{\textit{Our Trained Router}} \\
\midrule
{\xrouter}-7b-$\lambda$1 & 0.25 & 0.003050 & 0.93 & 0.003277 & 0.81 & 0.004586 & 0.81 & 0.009145 & 0.91 & 0.005768 & 0.50 & 0.012872 & 0.58 & 0.013740 & 0.92 & 0.002400 \\
{\xrouter}-7b-$\lambda$2 & \underline{0.28} & 0.002567 & 0.94 & 0.002224 & \underline{0.83} & 0.003230 & 0.74 & 0.006810 & 0.92 & 0.005819 & \underline{0.52} & 0.011134 & \textbf{0.65} & 0.012433 & 0.93 & 0.002128 \\
{\xrouter}-7b-$\lambda$3 & 0.24 & 0.002793 & 0.91 & 0.002908 & 0.80 & 0.004129 & 0.78 & 0.010000 & 0.90 & 0.005136 & 0.51 & 0.013007 & \underline{0.64} & 0.011583 & 0.91 & 0.002400 \\
\bottomrule
\end{tabular}
}
\end{center}
\caption{\textbf{Static Routing Baseline:} Comparison of trained {\xrouter} with baseline router models across multiple domains. Each task reports accuracy and average cost per query. Our trained routers achieve competitive performance while maintaining relatively lower cost.}
\label{tab:main}
\end{table*}

\paragraph{Main Results on Router Model Choice.}
Table~\ref{tab:main} demonstrates the strong effectiveness of our trained {\xrouter} than other static router models across diverse benchmarks. First, the {\xrouter} trained with Qwen2.5-7B-Instruct significantly outperforms its untrained counterpart, indicating that end-to-end router optimization substantially enhances decision quality beyond naïve model selection. Second, across most domains, {\xrouter} consistently surpasses purely open-source models of comparable scale, validating the advantage of learned cost–performance trade-offs. Third, {\xrouter} attains performance on par with top-tier proprietary systems such as GPT-5, while frequently achieving this at a fraction of their cost. For instance, {\xrouter}-7B-$\lambda$2 achieves near-GPT-5 accuracy on \textit{Olympiad Bench} with only about one-eighth of the original evaluation cost. These findings highlight that a trained routing model can make substantially more efficient allocation decisions than static or heuristic routing strategies, achieving high task performance while maintaining strong cost efficiency.

\begin{table*}[!t]
\begin{center}
\small
\renewcommand{\arraystretch}{1.2}
\tabcolsep=0.01\linewidth
\resizebox{\linewidth}{!}{
\begin{tabular}{lcccccccccccc}
\toprule
\multirow{2}{*}{\textbf{Model}} &
\multicolumn{2}{c}{\textbf{LiveCodeBenchv5}} &
\multicolumn{2}{c}{\textbf{GPQADiamond}} &
\multicolumn{2}{c}{\textbf{AIME25}} &
\multicolumn{2}{c}{\textbf{MTBench}} &
\multicolumn{2}{c}{\textbf{IFEval}} &
\multicolumn{2}{c}{\textbf{LiveBench}} \\
\cmidrule(lr){2-3}
\cmidrule(lr){4-5}
\cmidrule(lr){6-7}
\cmidrule(lr){8-9}
\cmidrule(lr){10-11}
\cmidrule(lr){12-13}
 & \textbf{Acc.} & \textbf{Cost} &
   \textbf{Acc.} & \textbf{Cost} &
   \textbf{Acc.} & \textbf{Cost} &
   \textbf{Acc.} & \textbf{Cost} &
   \textbf{Acc.} & \textbf{Cost} &
   \textbf{Acc.} & \textbf{Cost} \\
\midrule
\multicolumn{13}{c}{\textit{Direct Test on Existing Model}} \\
\midrule
GPT-4.1 & 0.4480 & 0.005953 & 0.3384 & 0.000568 & 0.4000 & 0.025765 & \textbf{9.4465} & 0.006725 & 0.906/0.9389 & 0.000440 & 57.2776 & 10.782766 \\
GPT-5-mini & 0.2832 & 0.006656 & \underline{0.7677} & 0.004484 & \textbf{0.8333} & 0.011523 & \underline{9.3031} & 0.004810 & 0.904/0.9311 & 0.000551 & \underline{78.5805} & 6.576172 \\
GPT-5 & 0.1756 & 0.049634 & \textbf{0.8586} & 0.033716 & 0.7333 & 0.048160 & 9.3019 & 0.033150 & \textbf{0.948/0.9662} & 0.003693 & \textbf{83.8888} & 35.863581 \\
\midrule
GPT-OSS-20B & 0.3190 & 0.000363 & 0.4848 & 0.000281 & 0.1333 & 0.000405 & 8.6761 & 0.000462 & 0.786/0.8309 & 0.000038 & 47.6848 & 0.405241 \\
Kimi-K2 & 0.4480 & 0.003089 & 0.6313 & 0.004225 & 0.2333 & 0.005865 & 9.2438 & 0.002626 & \underline{0.918/0.9454} & 0.000237 & 57.7387 & 3.783430 \\
Deepseek-R1 & 0.0609 & 0.015368 & 0.197 & 0.014830 & 0.7333 & 0.014701 & 7.7531 & 0.023351 & 0.806/0.8674 & 0.001216 & 35.3873 & 16.602912 \\
Qwen3-235B-Instruct & 0.4265 & 0.000473 & 0.4848 & 0.000148 & 0.2333 & 0.001199 & 9.2906 & 0.000847 & \underline{0.918/0.9454} & 0.000046 & 47.7381 & 0.924759 \\
Qwen3-235B-Thinking & 0.0251 & 0.006353 & 0.2121 & 0.006180 & 0.7333 & 0.006232 & 7.8469 & 0.009962 & 0.350/0.4889 & 0.000607 & 17.0317 & 7.745709 \\
\midrule
\multicolumn{13}{c}{\textit{{\xrouter} System}} \\
\midrule
{\xrouter}-7b-$\lambda$1 & \underline{0.6344} & 0.007825 & 0.7121 & 0.004872 & \underline{0.7667} & 0.010513 & 8.0227 & 0.007064 & 0.778/0.8479 & 0.000573 & 57.0392 & 6.424311 \\
{\xrouter}-7b-$\lambda$2 & \textbf{0.6774} & 0.007166 & 0.7172 & 0.004548 & \underline{0.7667} & 0.015377 & 7.9780 & 0.006981 & 0.784/0.8518 & 0.000569 & 56.9508 & 6.411576 \\
{\xrouter}-7b-$\lambda$3 & 0.4229 & 0.006447 & 0.6061 & 0.001320 & 0.5000 & 0.008646 & 8.3165 & 0.005970 & 0.806/0.8635 & 0.000496 & 61.4092 & 7.873074 \\
\bottomrule
\end{tabular}
}
\end{center}
\caption{\textbf{Single Model Baseline:} Comparison of {\xrouter} system with baseline single model evaluation across multiple domains. Each task reports accuracy and average cost per query. Our trained system achieves competitive performance while maintaining relatively lower cost.}
\label{tab:main_single}
\end{table*}

\paragraph{Single-Model Baseline Analysis.}
Table~\ref{tab:main_single} presents a comprehensive comparison between the proposed {\xrouter} framework and a range of leading single-model baselines across diverse benchmarks. First, despite operating at a similar or lower computational scale, the {\xrouter}-7B variants deliver performance that rivals or exceeds that of significantly larger proprietary and open-source models. Notably, {\xrouter}-7B-$\lambda$3 achieves the highest average accuracy on \textit{LiveCodeBenchv5}, while maintaining a moderate cost profile, demonstrating that cost-aware training effectively balances model capability and inference efficiency. Second, the consistently strong results of {\xrouter}-7B variants across \textit{AIME25} and \textit{LiveBench} tasks suggest that learned routing priors enable the system to generalize across heterogeneous domains without explicit retraining. Third, compared with large proprietary systems such as GPT-5, {\xrouter} usually attains 80–90\% of accuracy while consuming less than one-fifth of the cost, such as in \textit{GPQA}. Overall, these findings affirm that the {\xrouter} architecture not only narrows the gap between open and closed systems but also redefines the efficiency frontier of large-scale LM utilization.

\subsection{Analysis}

\paragraph{Effect of Cost Penalty (\boldmath$\lambda$).}
Across both tables, the cost penalty $\lambda$ governs a trade-off between accuracy and computational efficiency: theoretically, smaller penalties should encourage the router to rely more heavily on expensive experts to maximize performance, whereas larger penalties constrain spending but risk underutilizing model capacity. Empirically, a moderate setting (\(\lambda=2\)) tends to yield the most balanced results. In Table~\ref{tab:main}, {\xrouter}-7B-$\lambda2$ maintains strong accuracy across reasoning and coding tasks while keeping overall cost low, whereas reducing the penalty (\(\lambda=1\)) generally leads to higher computational spending, as observed in tasks such as \textit{Minerva} and \textit{MATH-500}. Interestingly, our experiments reveal that higher expenditure does not necessarily translate to better performance, nor does increasing $\lambda$ always result in lower cost. This non-monotonic behavior may arise because extreme $\lambda$ values impose overly rigid cost constraints that reduce model accuracy; to preserve the accuracy, which is prioritized in reward shaping, the training process may inadvertently adjust cost dynamics in less predictable ways.

In Table~\ref{tab:main_single}, we observe a similar pattern: $\lambda=2$ often provides the most favorable cost–performance balance, using less computation than $\lambda=1$ while often achieving higher accuracy than $\lambda=3$. These results highlight the importance of careful $\lambda$ tuning: an appropriately chosen penalty enables the router to make economically efficient allocation decisions while sustaining near-optimal accuracy, avoiding both the wastefulness of small penalties and the conservatism of excessively large ones.

\begin{figure*}
    \centering
    \includegraphics[width=\linewidth]{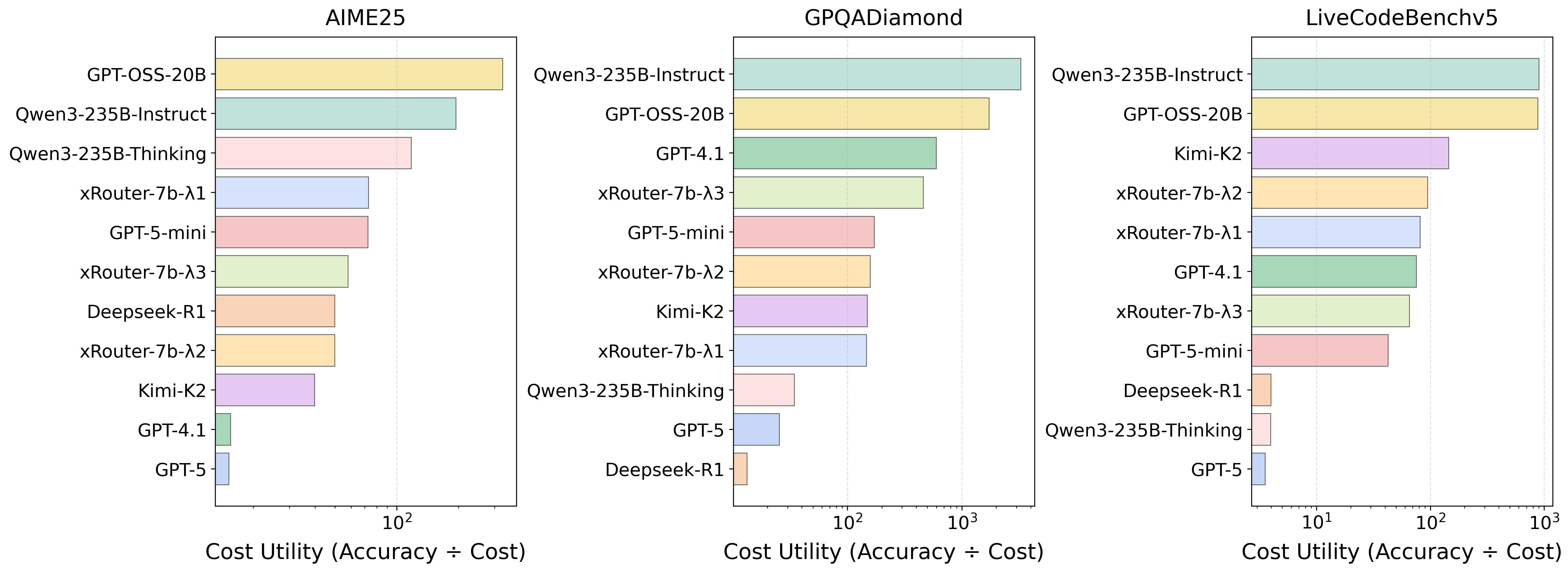}
    \caption{The cost utility calculation on \textit{AIME25}, \textit{GPQADiamond} and \textit{LiveCodeBenchv5}.}
    \label{fig:cost_utility}
    \vspace{-2mm}
\end{figure*}

\paragraph{Cost Utility Calculation.}
To better analyze the balance between cost and performance, we introduce a new metric called \textit{cost utility}, defined as the ratio of accuracy to cost. This metric reflects the level of performance a model or system can achieve per unit cost. Thus, a higher cost utility indicates greater efficiency in attaining strong performance at lower cost.

We compare the {\xrouter} series with other single-model baselines in \Cref{fig:cost_utility}, focusing on three benchmarks: \textit{AIME25}, \textit{GPQADiamond}, and \textit{LiveCodeBenchv5}. Our results reveal that many open-source models tend to exhibit higher cost utility due to their significantly lower API prices compared to proprietary systems such as GPT-5. Consequently, {\xrouter} may appear to lag behind these models in cost utility. However, as shown in \Cref{tab:main_single}, models with higher cost utility often achieve lower overall accuracy---an undesirable trade-off in many practical applications where performance is of primary importance.

We also observe that models like GPT-5 and Deepseek-R1 typically demonstrate the lowest cost utility, whereas the {\xrouter} models achieve a more balanced profile: offering higher accuracy than low-cost models and better cost efficiency than expensive ones. This balance underscores the central advantage of {\xrouter}: its ability to maintain strong performance while remaining economically efficient.

\begin{figure*}
    \centering
    \includegraphics[width=\linewidth]{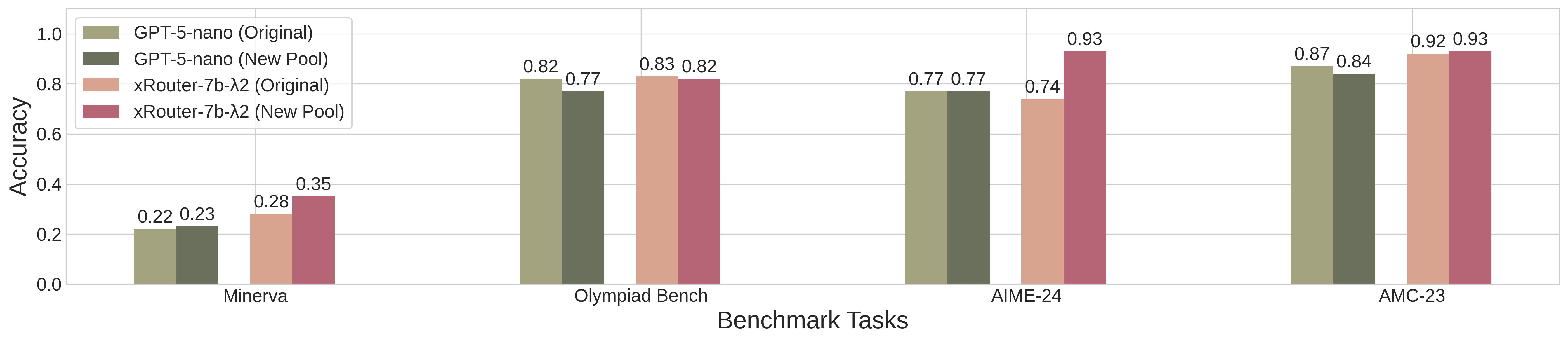}
    \caption{Compare of router performance under different selection model pool.}
    \label{fig:compare}
    \vspace{-2mm}
\end{figure*}

\paragraph{Impact of Offloading Model Pool.}
In the previous analysis, all routing systems shared the same downstream model pool from which the router could select. To further examine the impact of the model pool composition, we expanded the original pool by including additional models:
\textit{{GPT-4.1-mini, GPT-4.1-nano, o4-mini, Qwen3-Coder-480B, Kimi-K2}}.
Introducing more models provides routers with greater flexibility during inference, but also increases decision complexity, thereby presenting a stronger test of the router’s adaptability and robustness.

As shown in \Cref{fig:compare}, we evaluate this effect across four representative tasks, comparing {\xrouter}-7b-$\lambda2$ with GPT-5-nano as the router model. The results reveal that when the model pool is enlarged, our trained {\xrouter} generally exhibits improved or stable performance, whereas GPT-5-nano, used as an off-the-shelf router, shows similar or even degraded accuracy. This suggests that {\xrouter} is more robust to changes in the routing environment. We attribute this advantage to our dynamic training paradigm, where the downstream model pool continuously varies across training instances. Consequently, {\xrouter} learns to reason over model capabilities contextually and make adaptive routing decisions, rather than overfitting to static model patterns. This property enables it to generalize better when exposed to an expanded or evolving set of candidate models.

Another noteworthy observation is that the newly added models generally lack strong domain-specific capabilities, such as mathematical reasoning (what we mainly test on). Consequently, the decline in GPT-5-nano’s performance may be attributed to the introduction of noise from these weaker or less specialized models. In contrast, the performance improvement observed for {\xrouter} appears counterintuitive at first glance. Upon further inspection, we find that the overall computational cost increased compared to the original pool. This indicates that under the expanded pool, {\xrouter} tends to select stronger but also more expensive models, which may partly explain the observed performance gains. This phenomenon highlights an important insight: router systems may exhibit sensitivity to the composition of the model pool, leading to variability in routing behavior. While the performance improvements appear positive, such fluctuations also suggest potential risks of instability or inconsistency when the available downstream models change.

\begin{figure*}[t]
    \centering
    \setlength{\tabcolsep}{1pt} 
    \renewcommand{\arraystretch}{1.5} 

    \begin{tabular}{cccc}
        \includegraphics[width=0.24\linewidth]{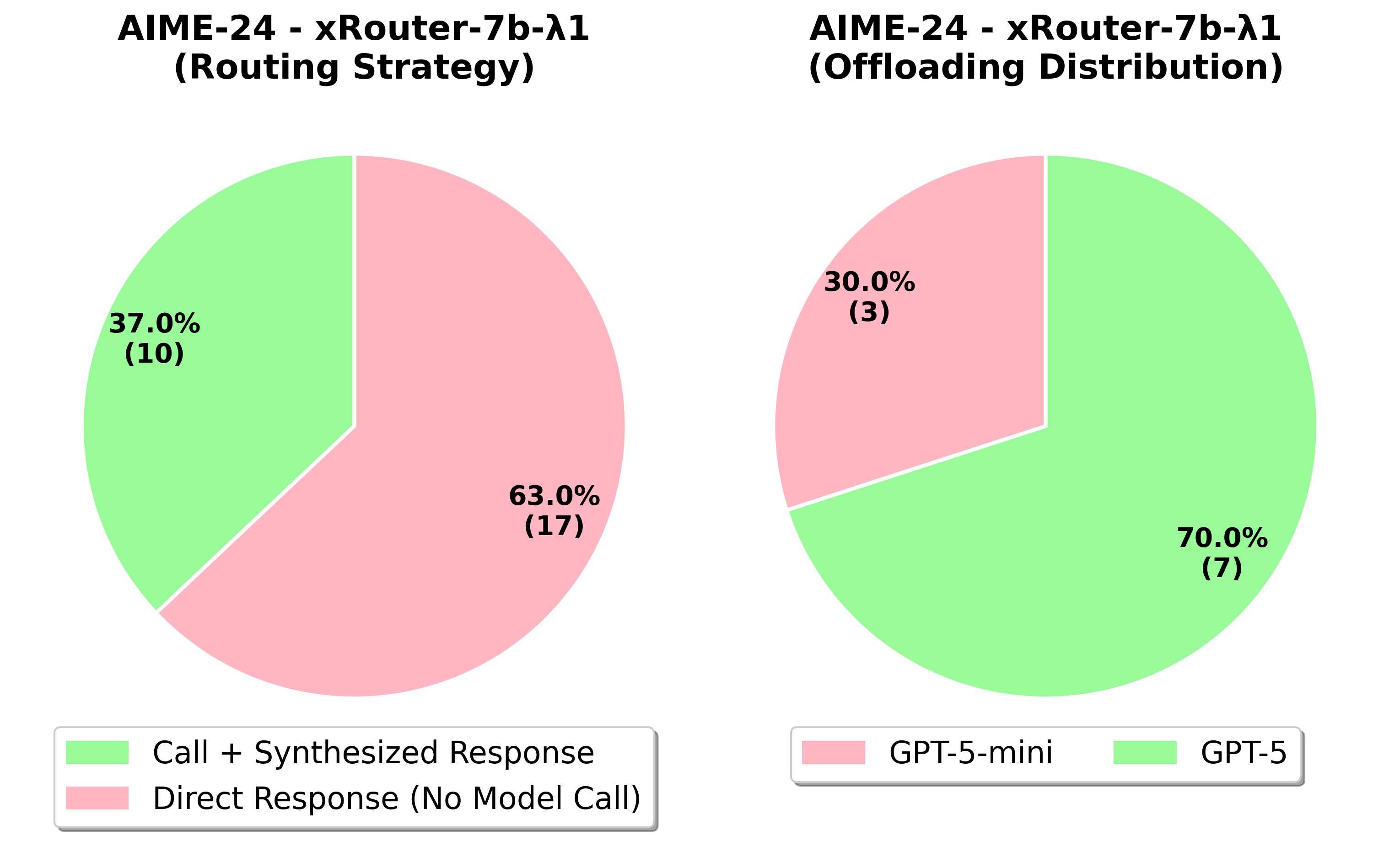} &
        \includegraphics[width=0.24\linewidth]{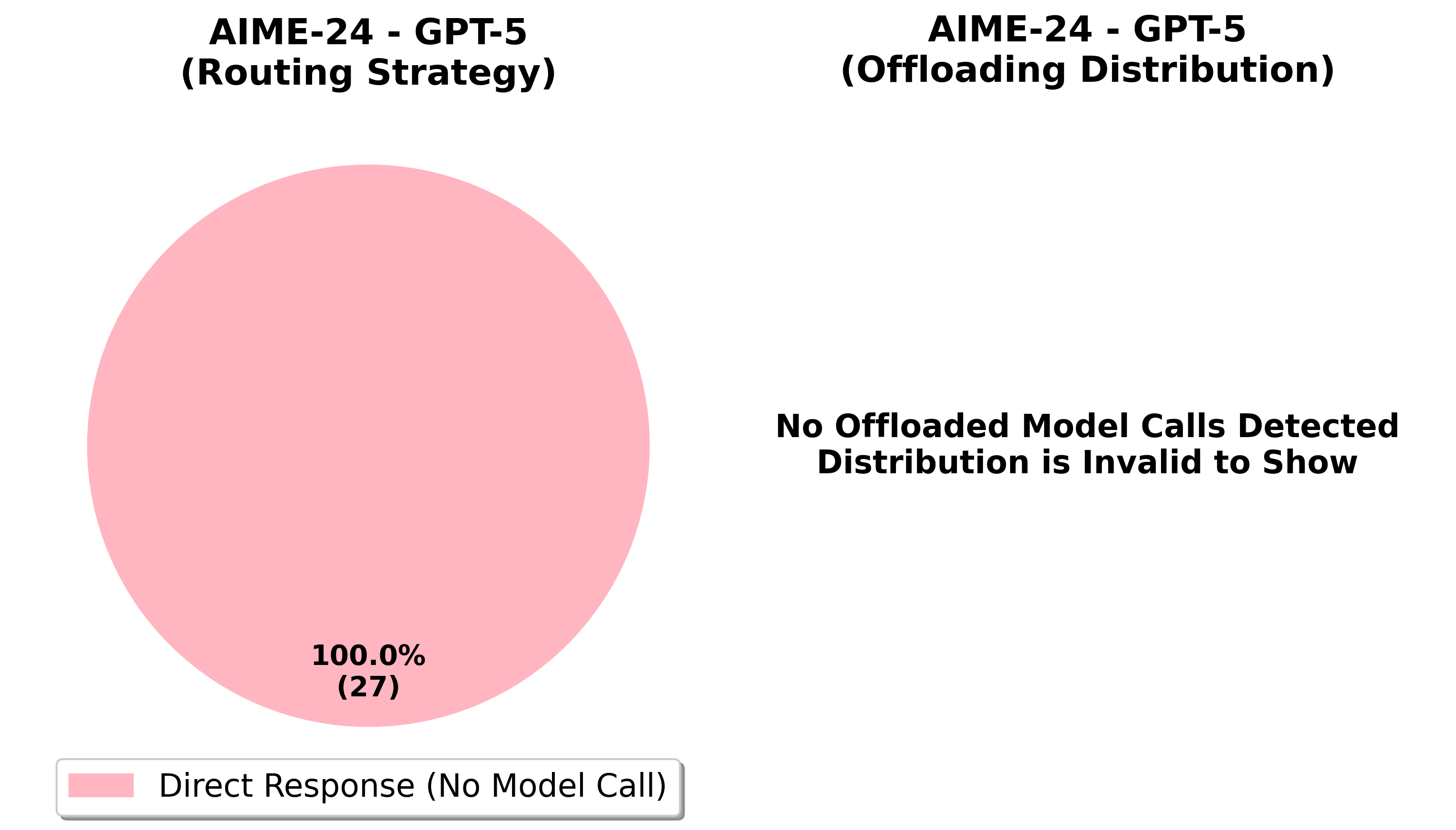} &
        \includegraphics[width=0.24\linewidth]{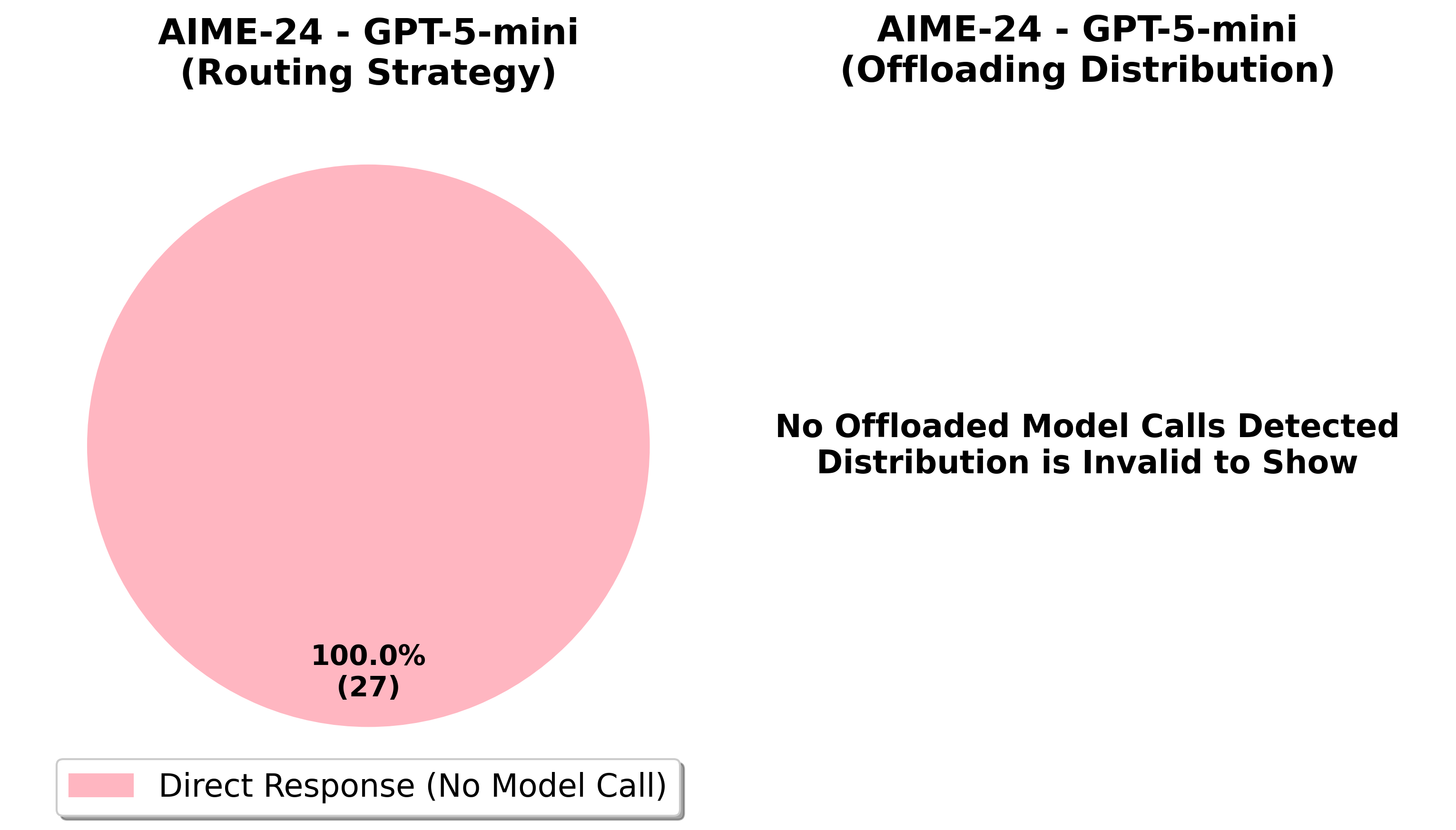} &
        \includegraphics[width=0.24\linewidth]{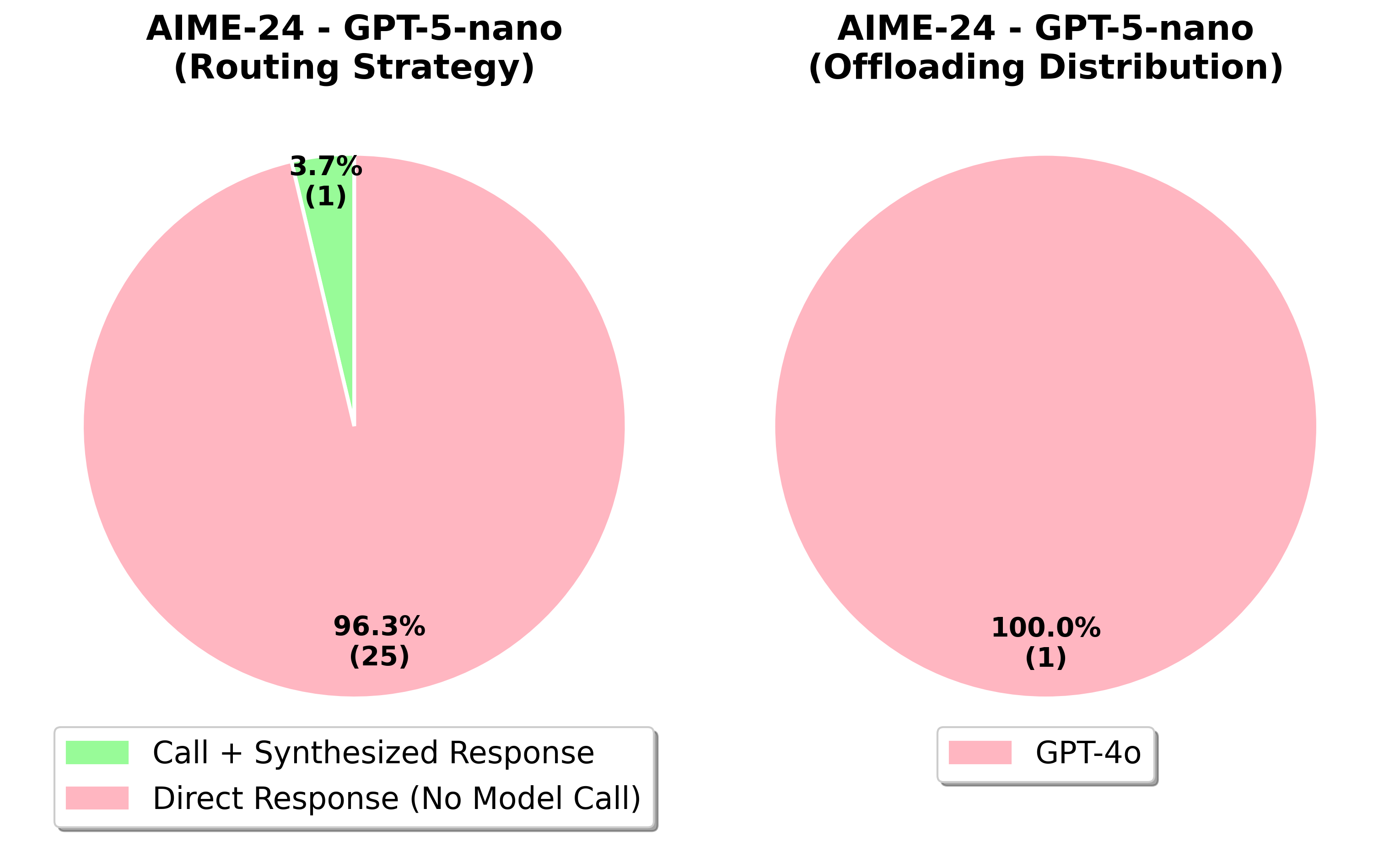} \\
        \includegraphics[width=0.24\linewidth]{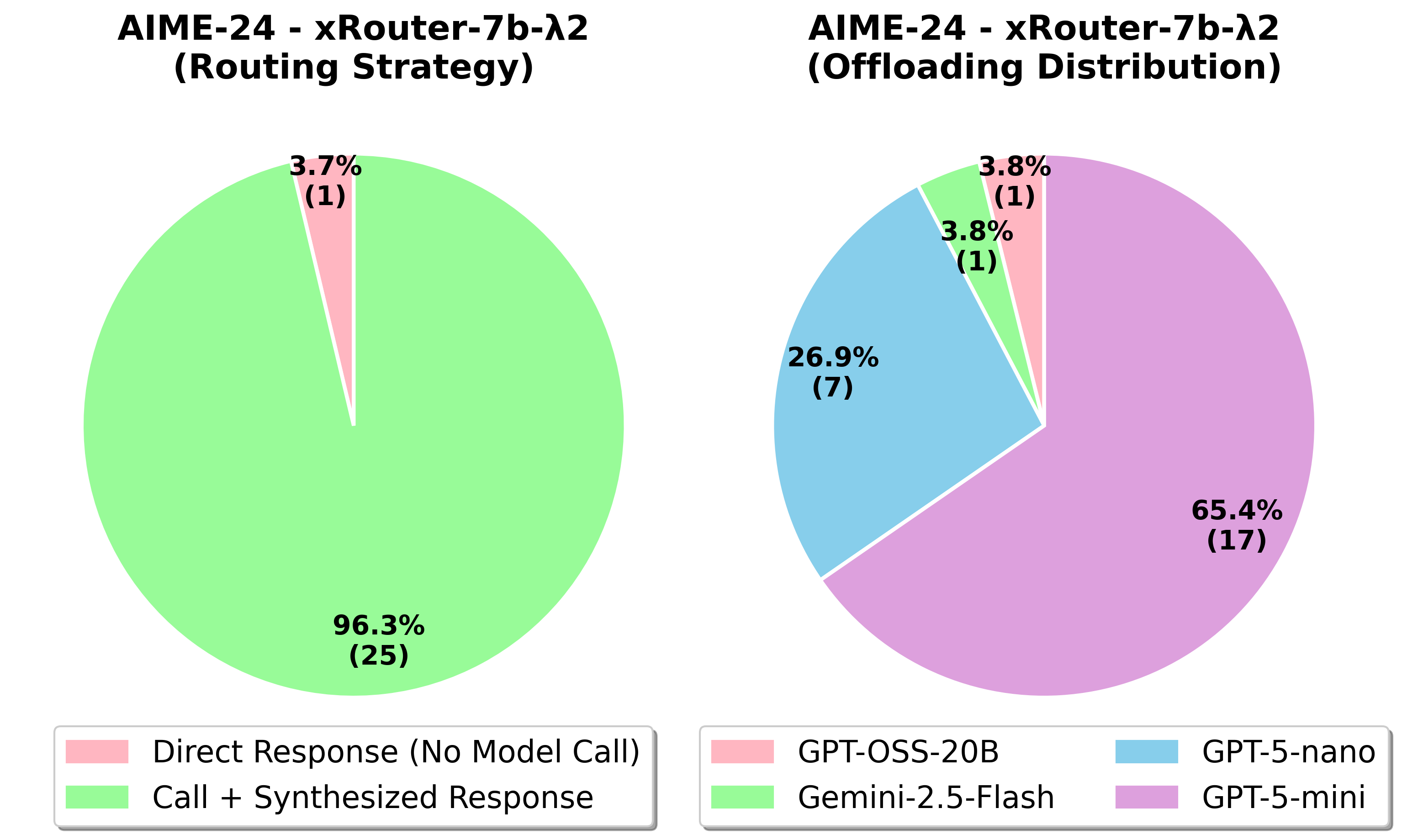} &
        \includegraphics[width=0.24\linewidth]{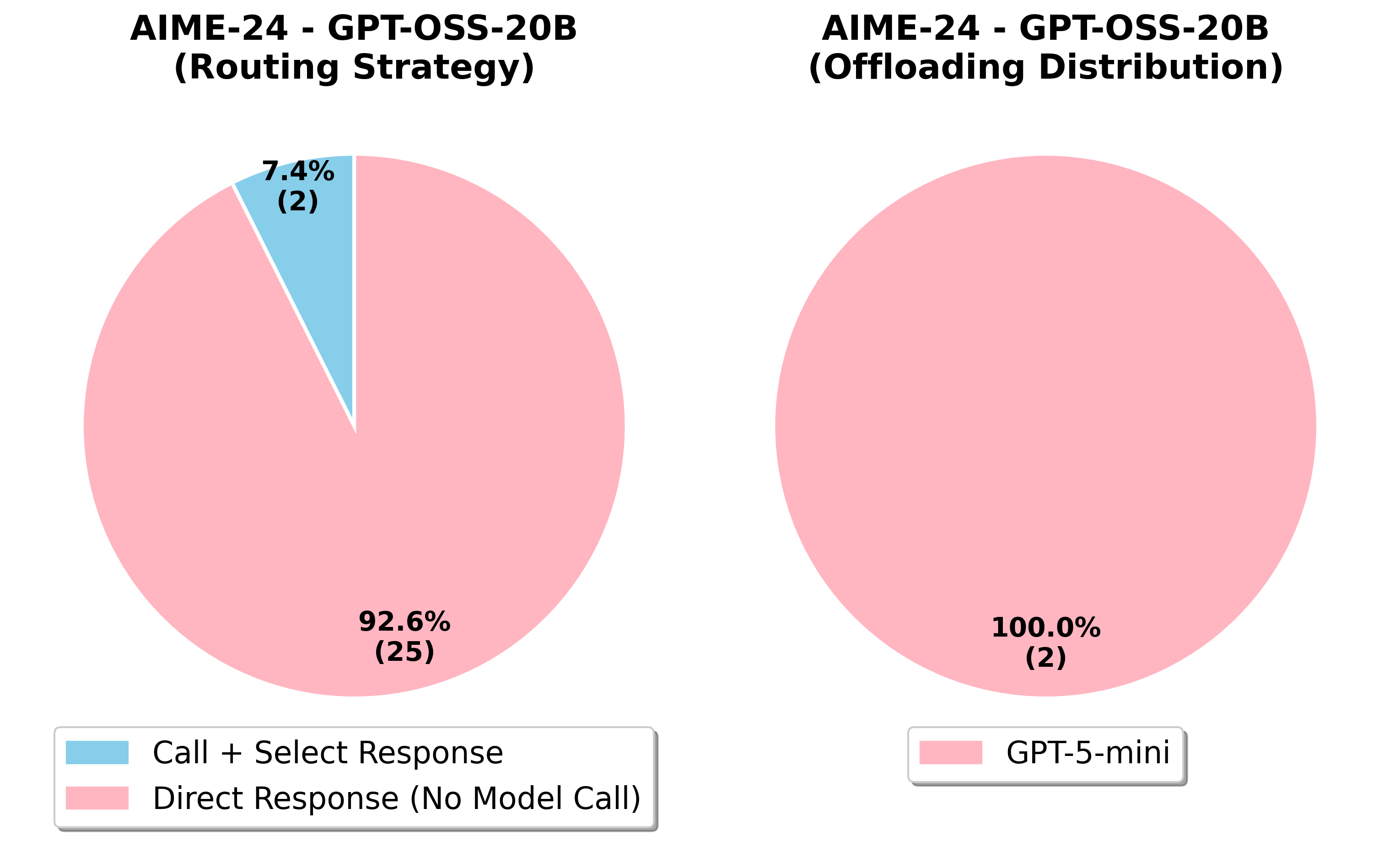} &
        \includegraphics[width=0.24\linewidth]{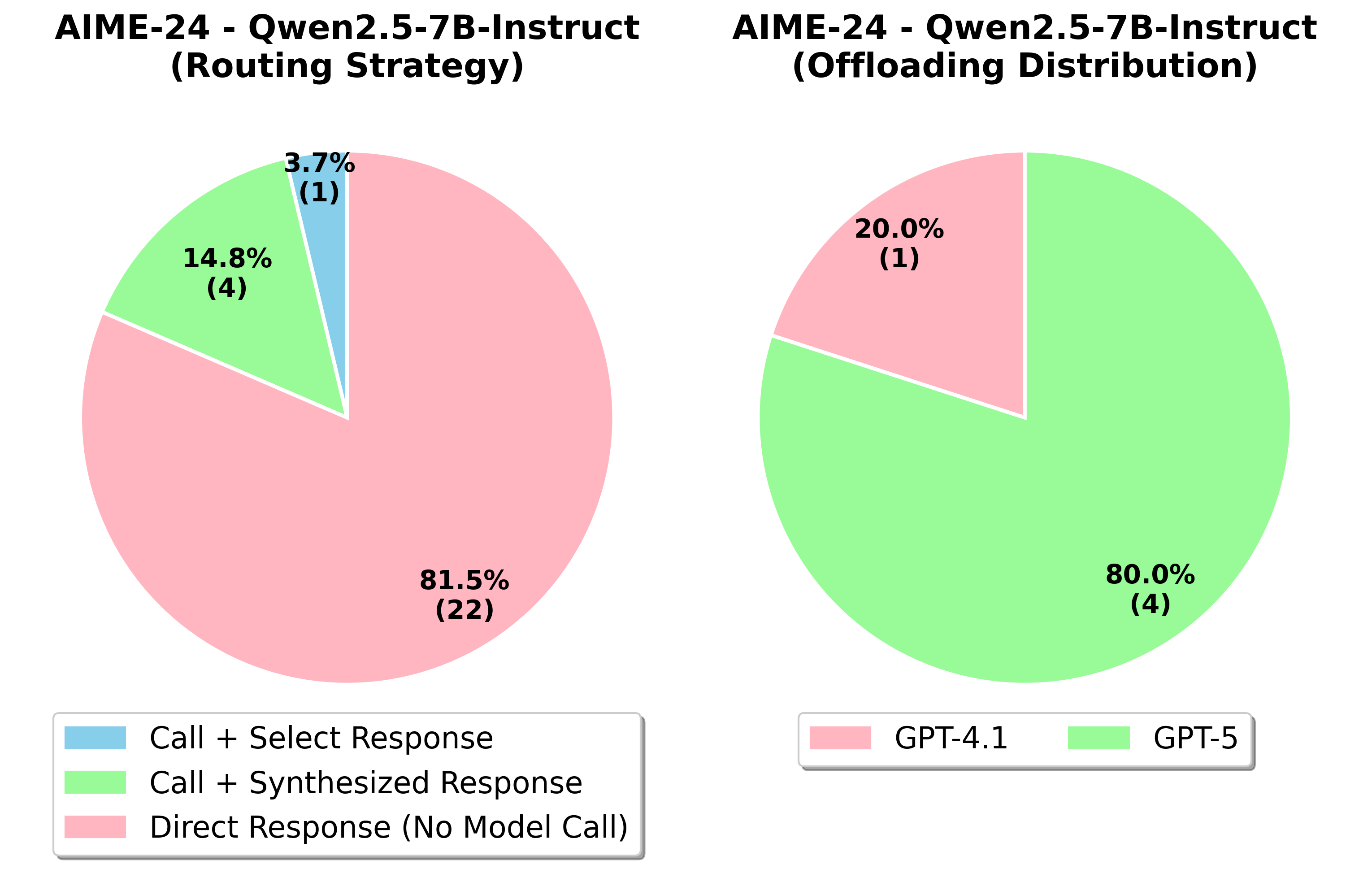} &
        \includegraphics[width=0.24\linewidth]{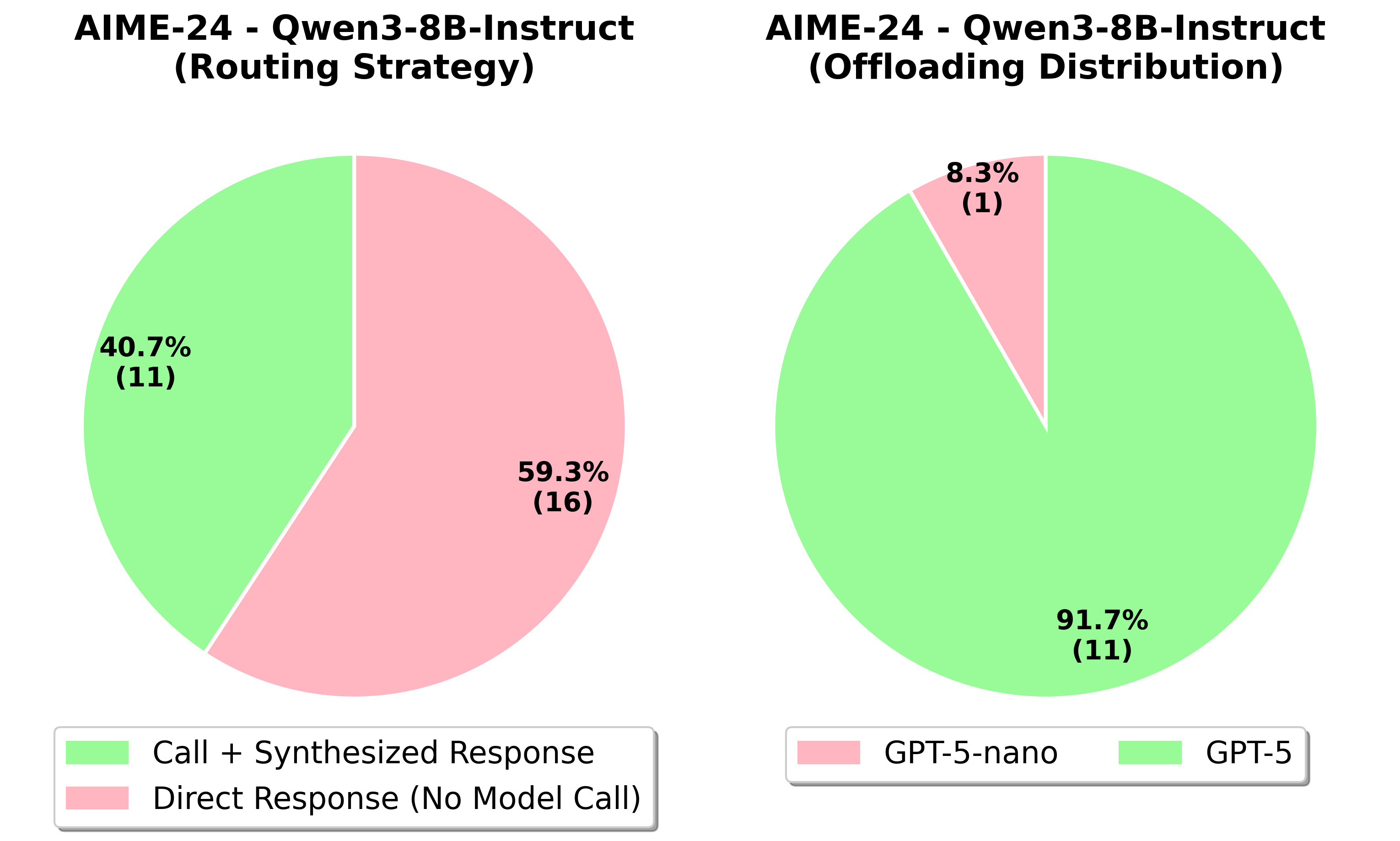} \\

        \includegraphics[width=0.24\linewidth]{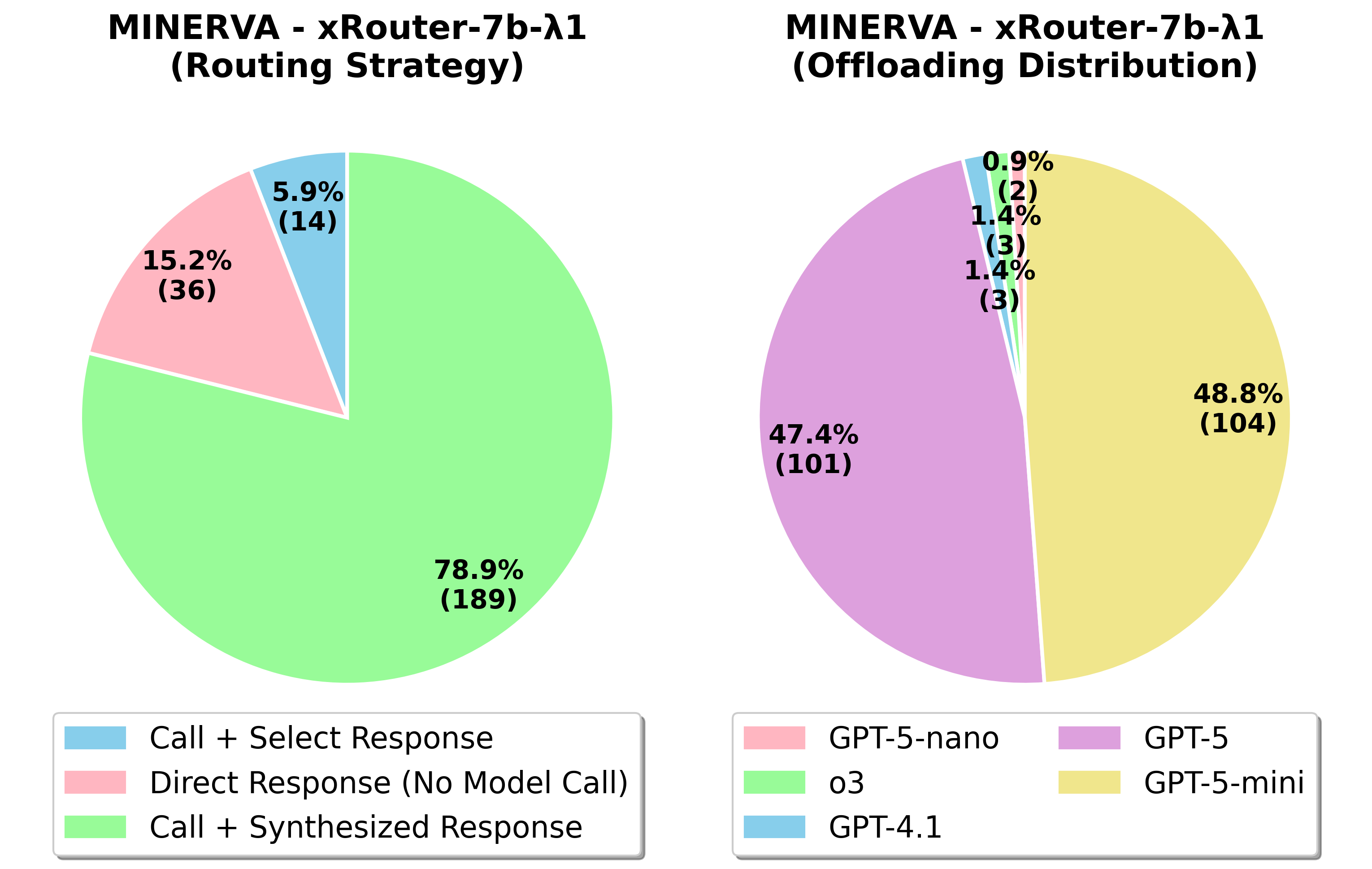} &
        \includegraphics[width=0.24\linewidth]{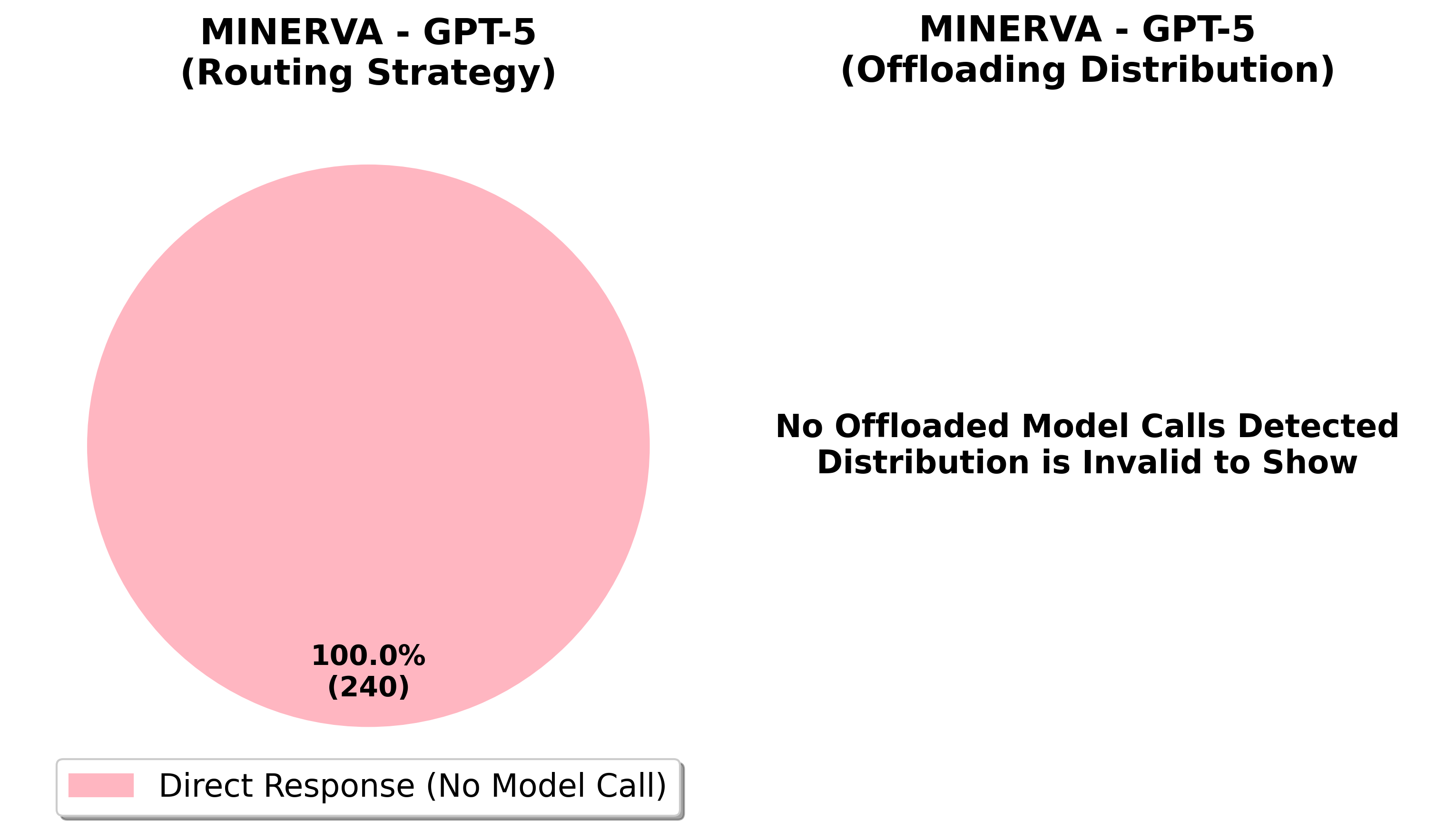} &
        \includegraphics[width=0.24\linewidth]{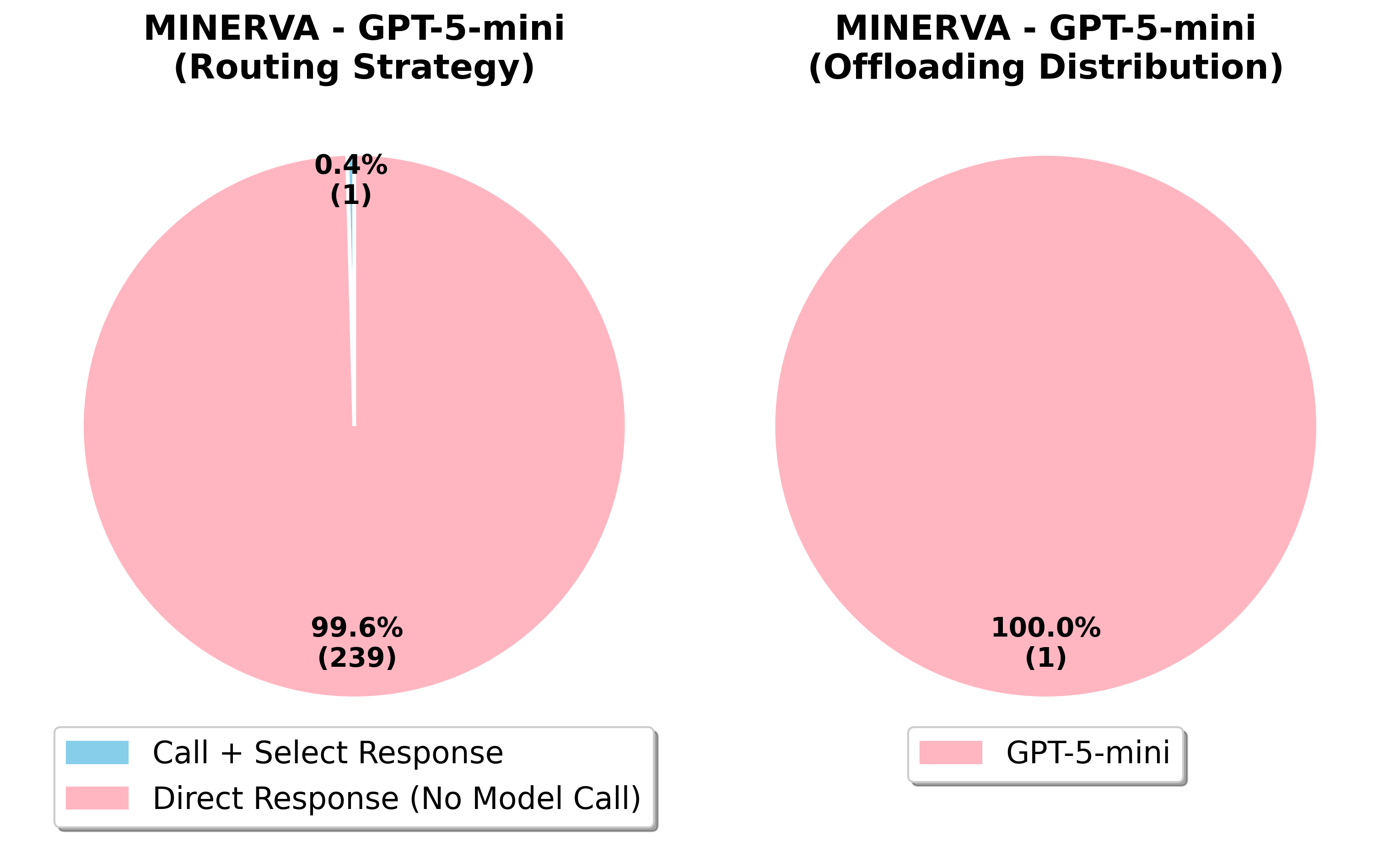} &
        \includegraphics[width=0.24\linewidth]{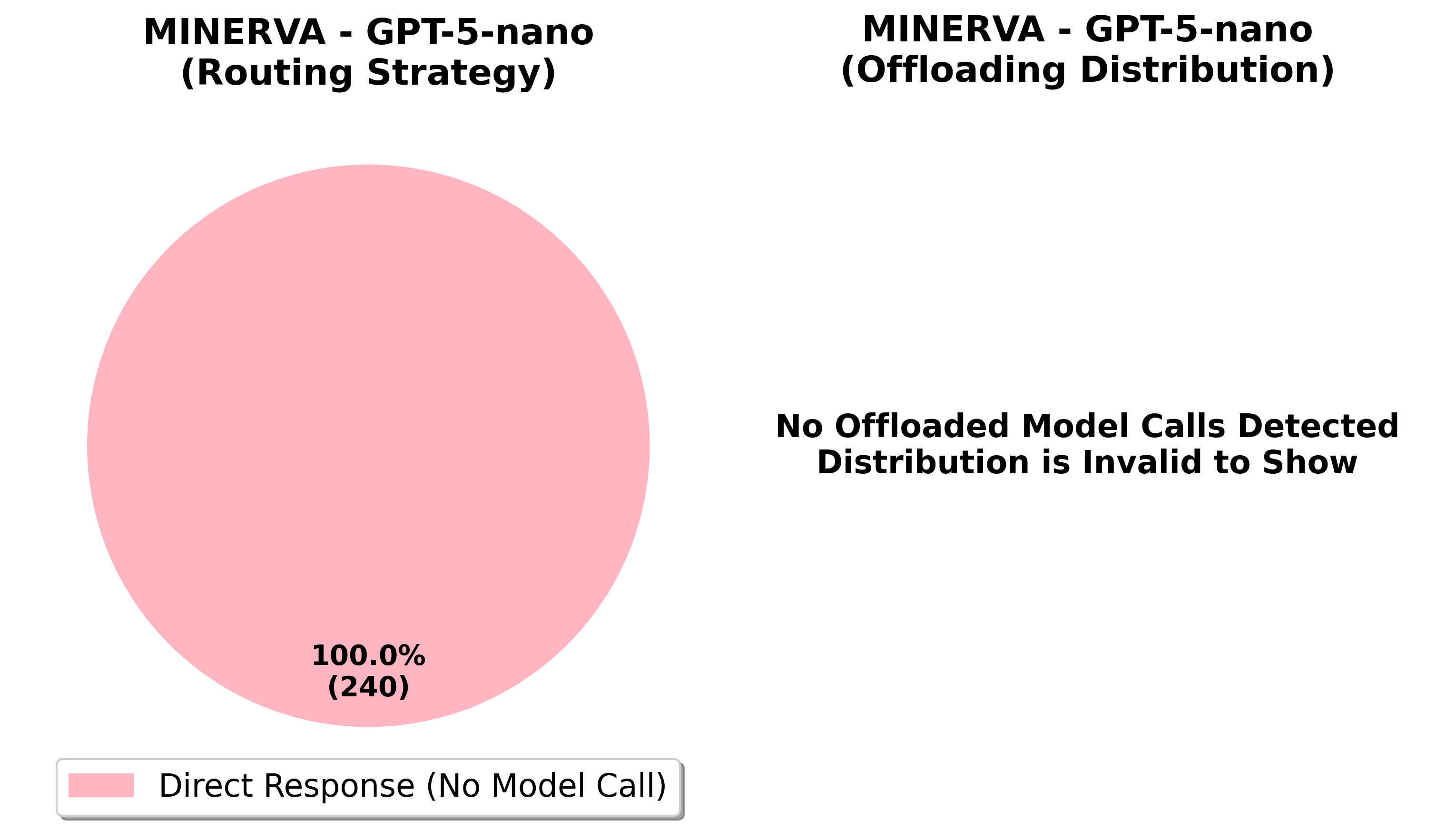} \\
        \includegraphics[width=0.24\linewidth]{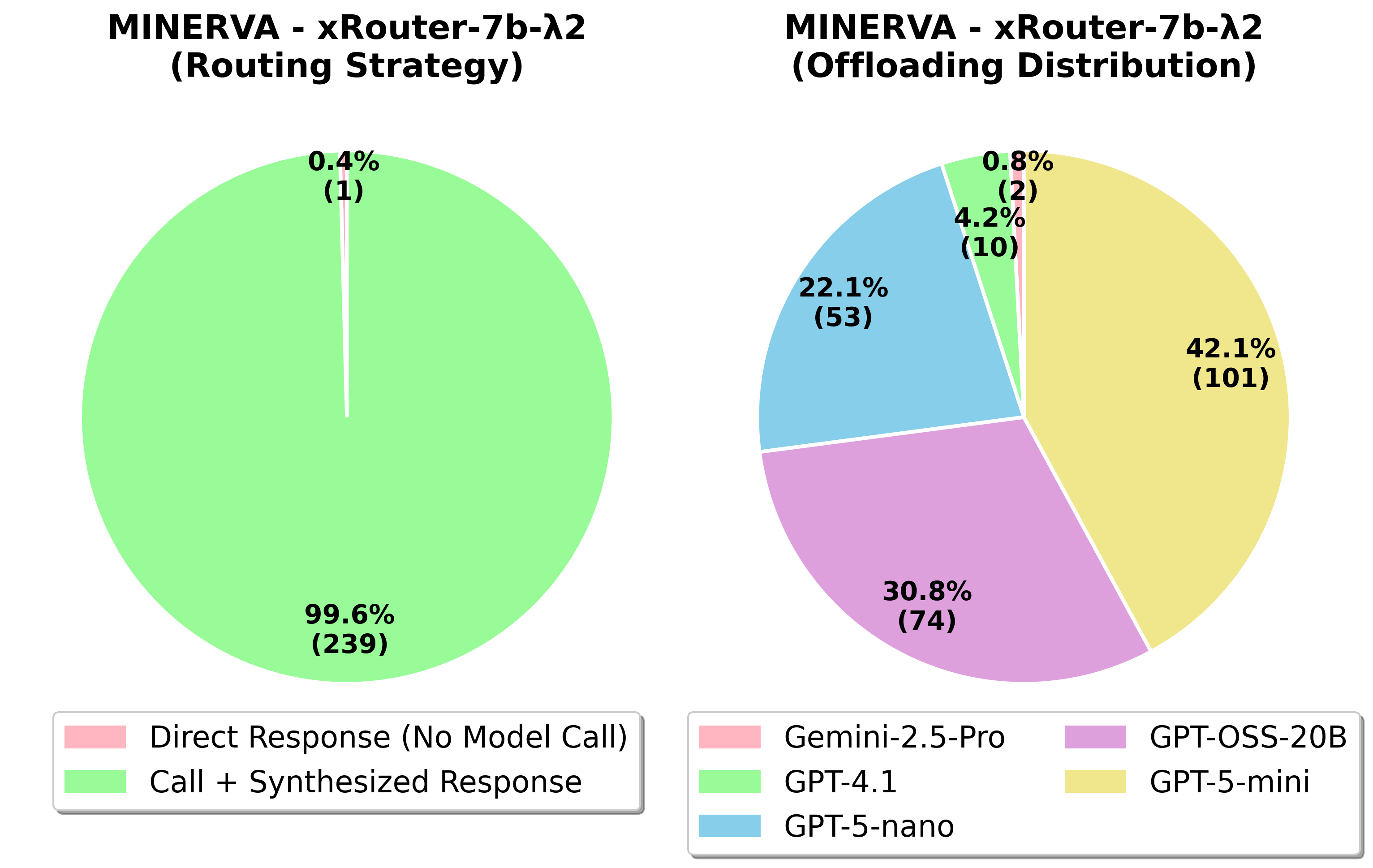} &
        \includegraphics[width=0.24\linewidth]{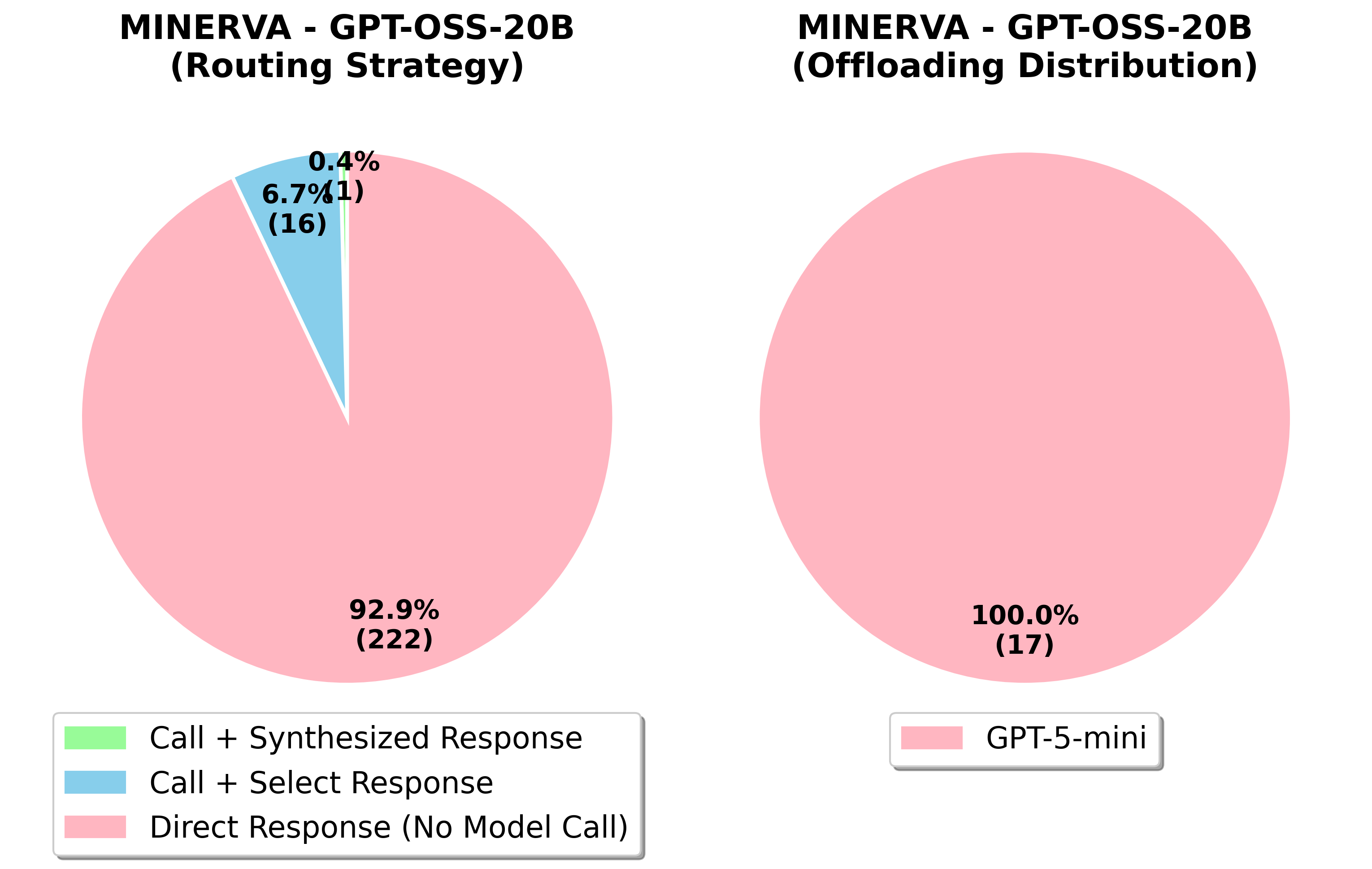} &
        \includegraphics[width=0.24\linewidth]{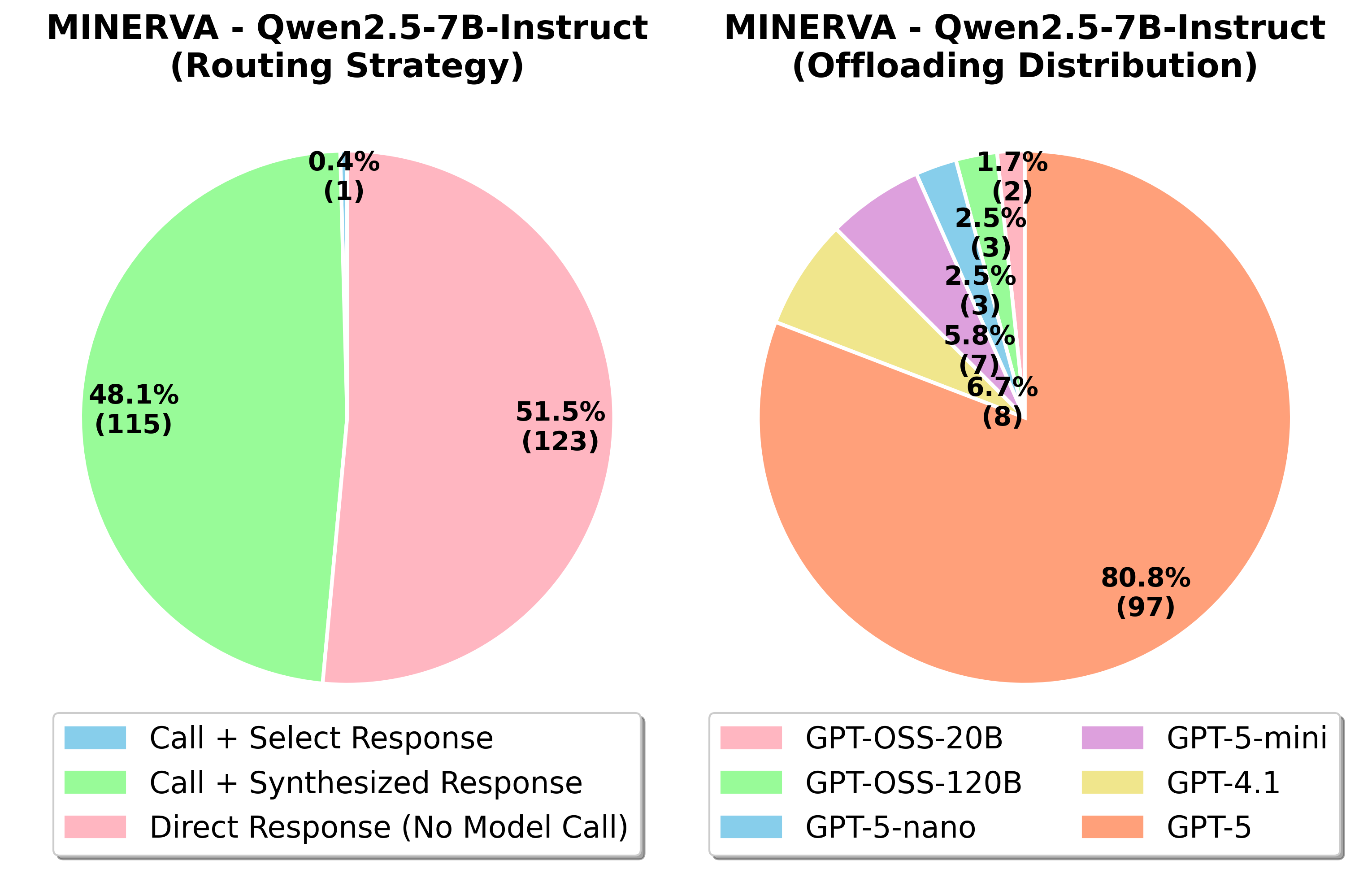} &
        \includegraphics[width=0.24\linewidth]{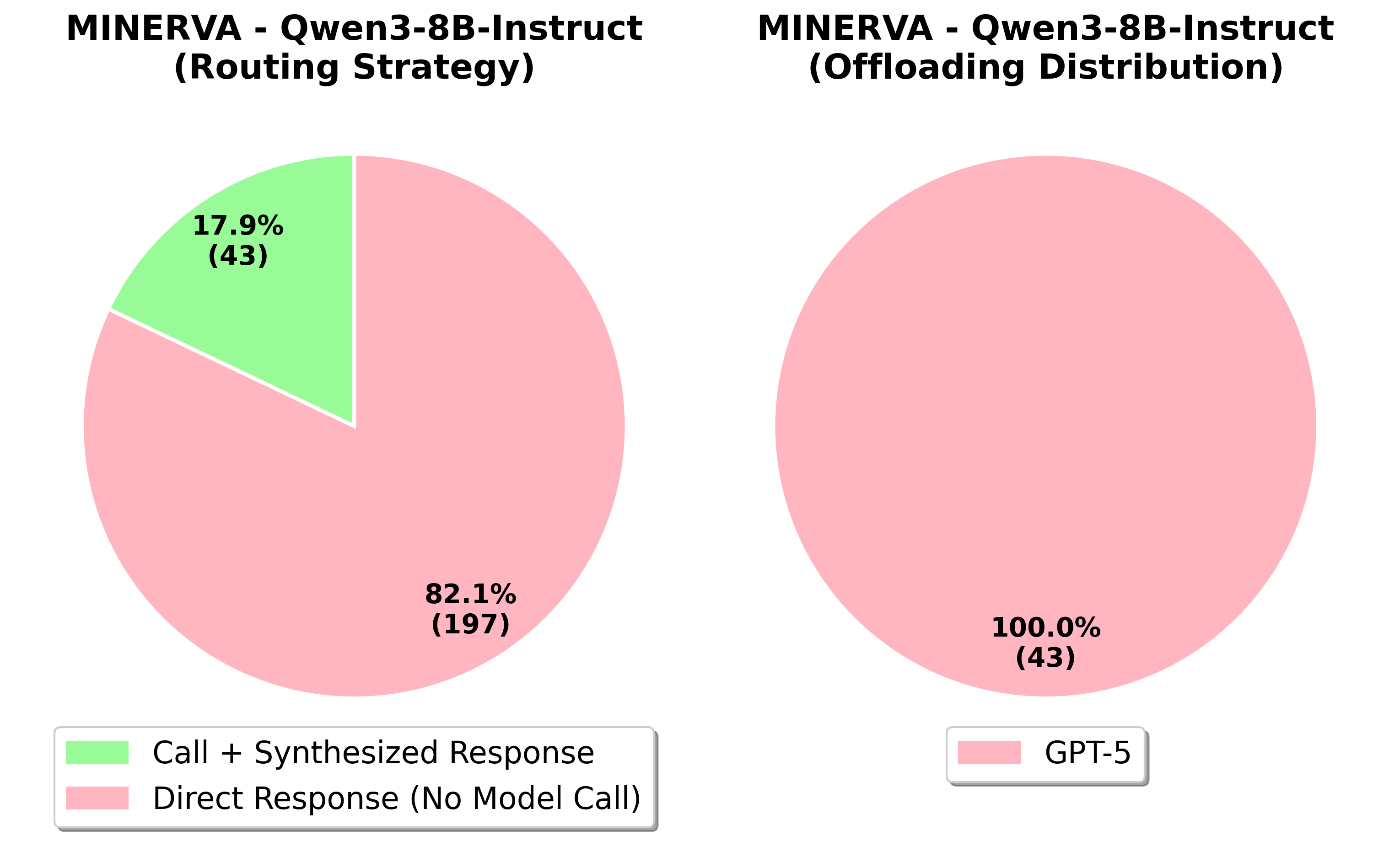} \\

        \includegraphics[width=0.24\linewidth]{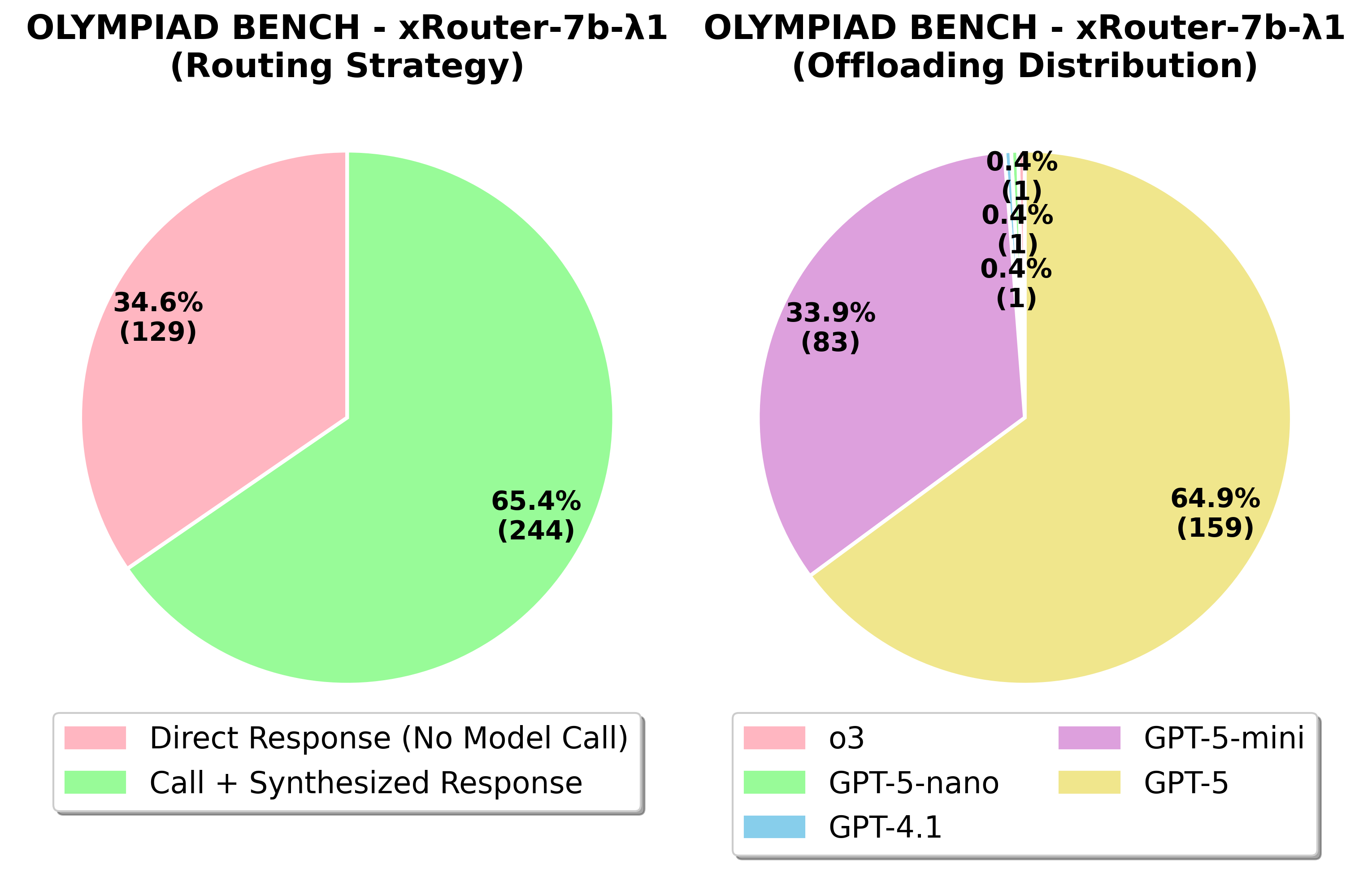} &
        \includegraphics[width=0.24\linewidth]{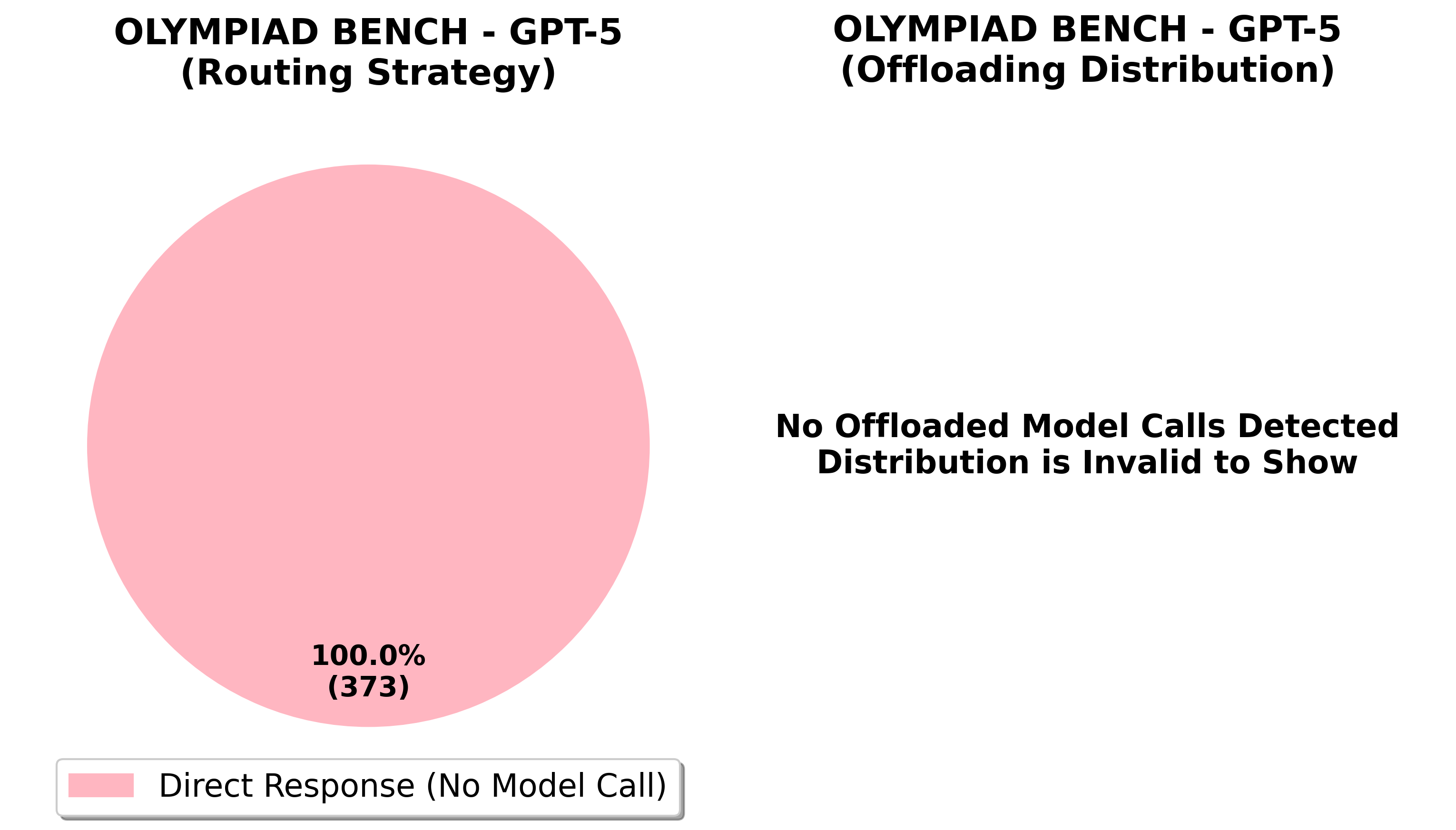} &
        \includegraphics[width=0.24\linewidth]{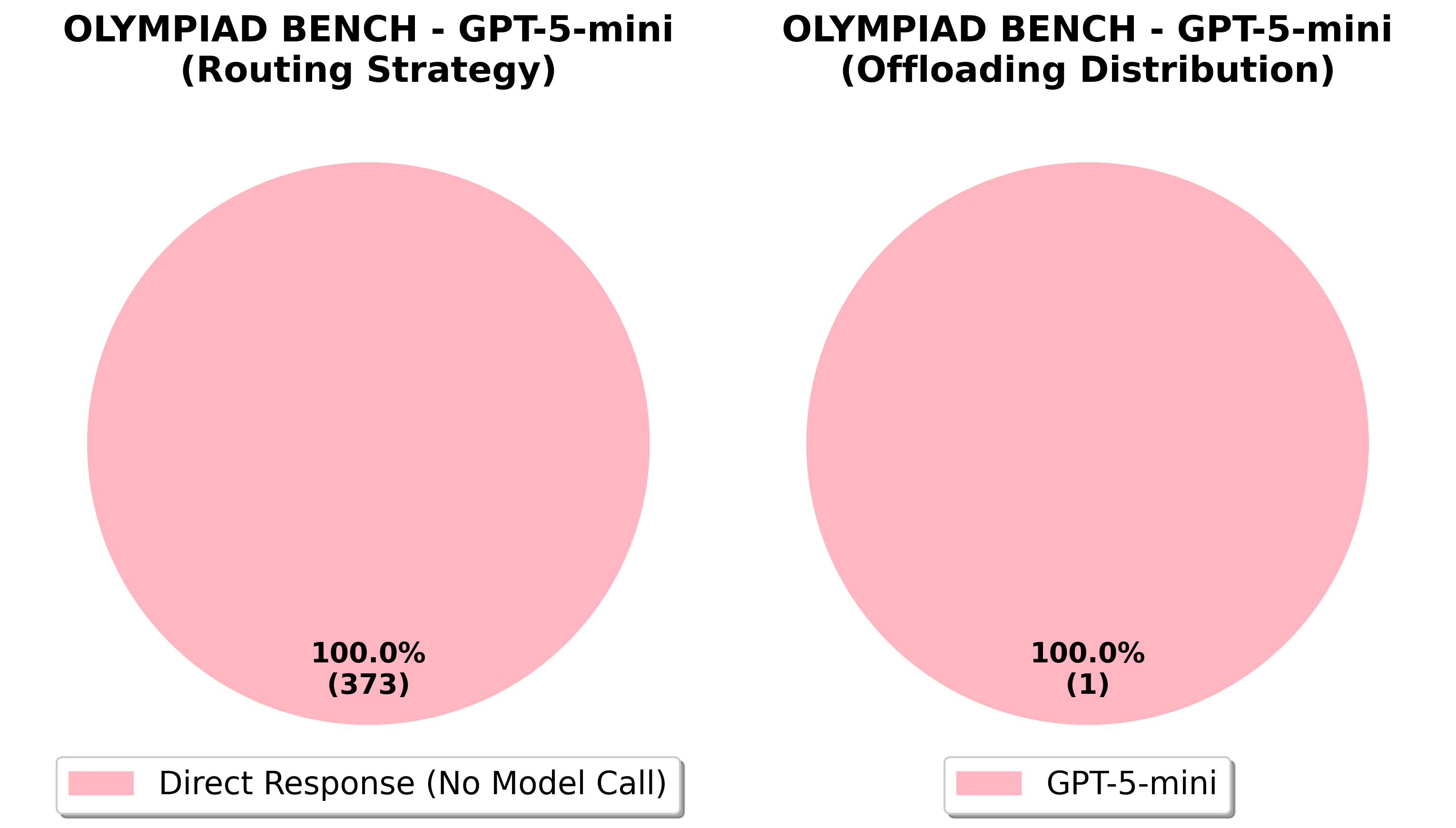} &
        \includegraphics[width=0.24\linewidth]{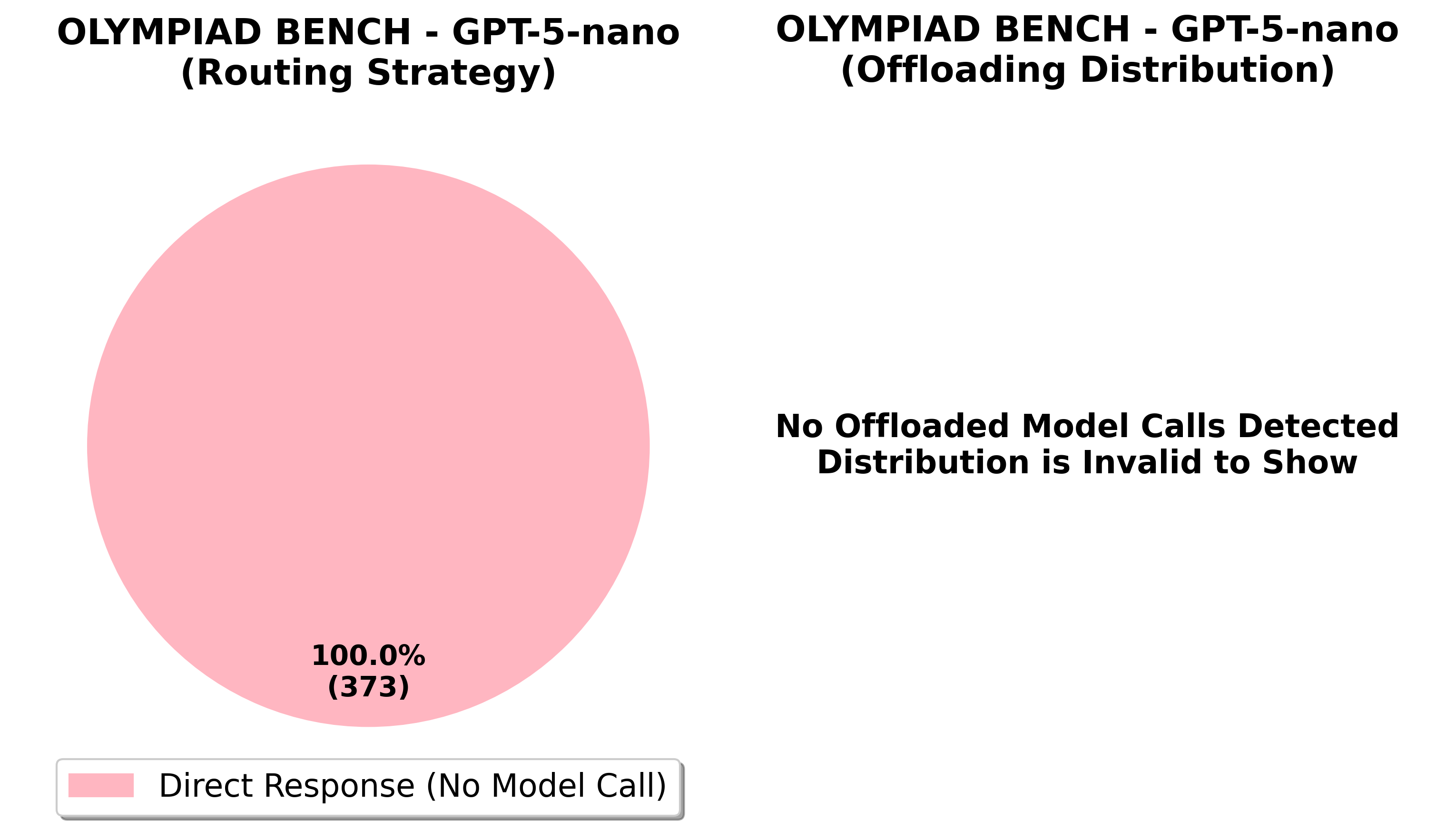} \\
        \includegraphics[width=0.24\linewidth]{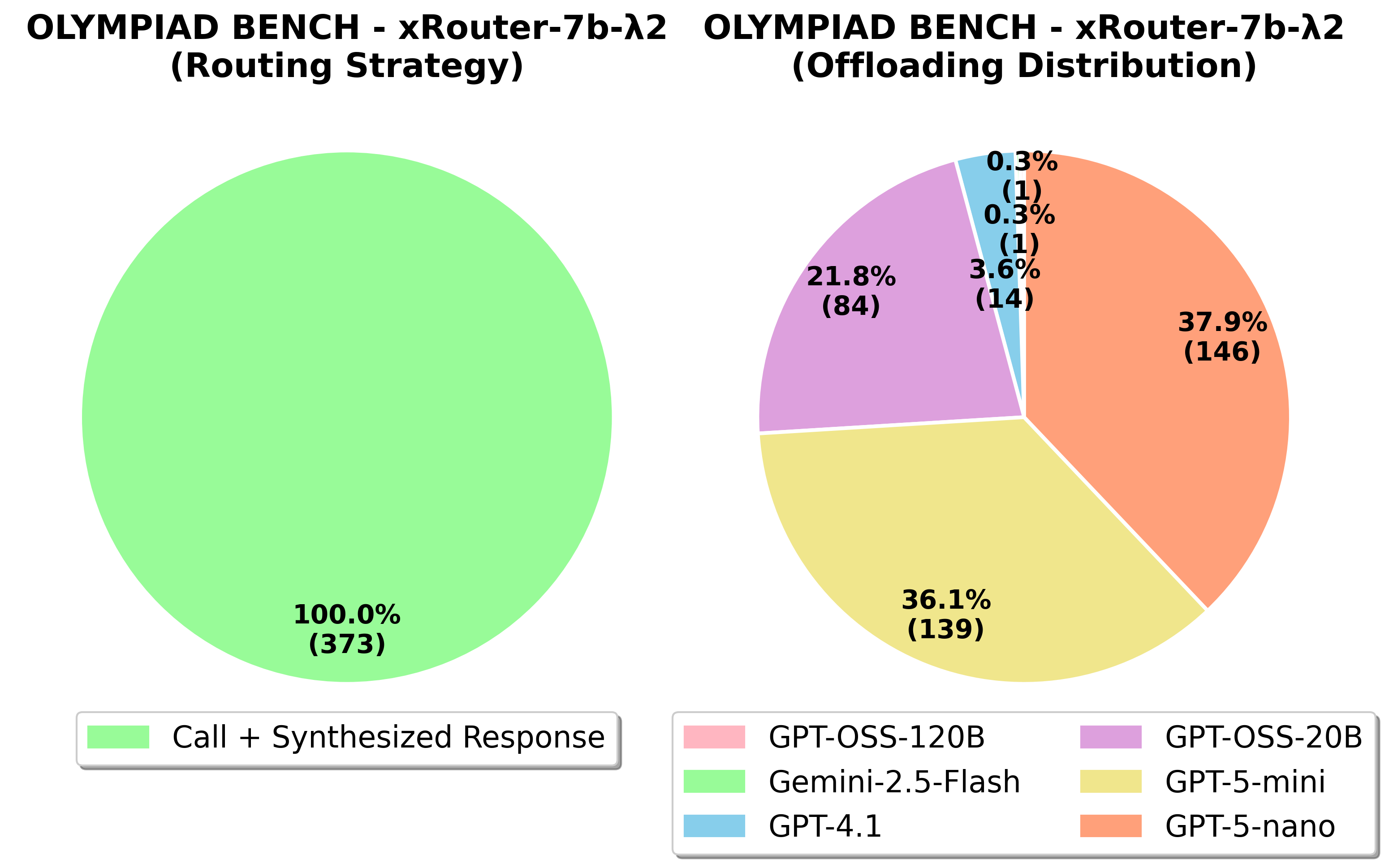} &
        \includegraphics[width=0.24\linewidth]{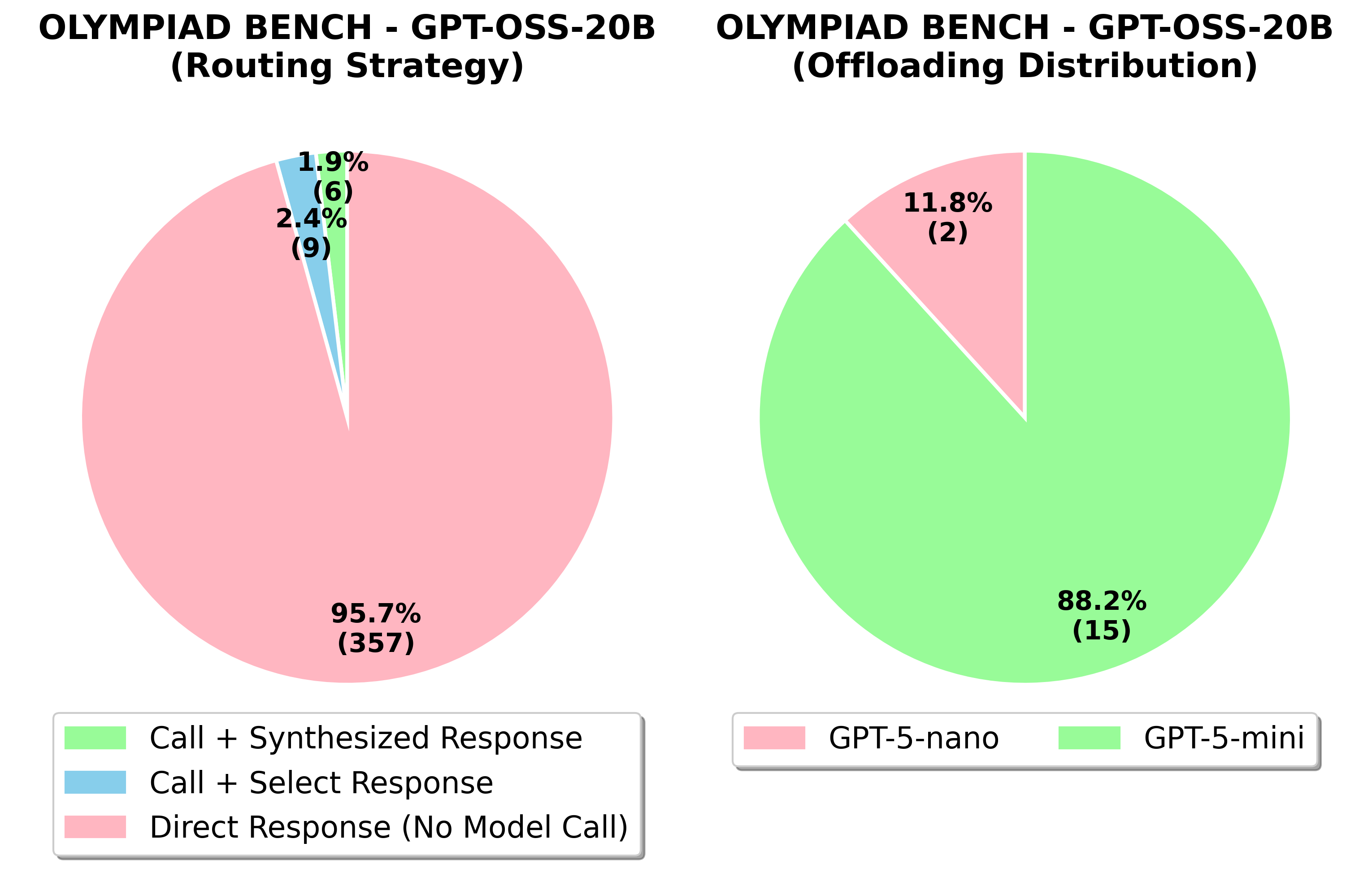} &
        \includegraphics[width=0.24\linewidth]{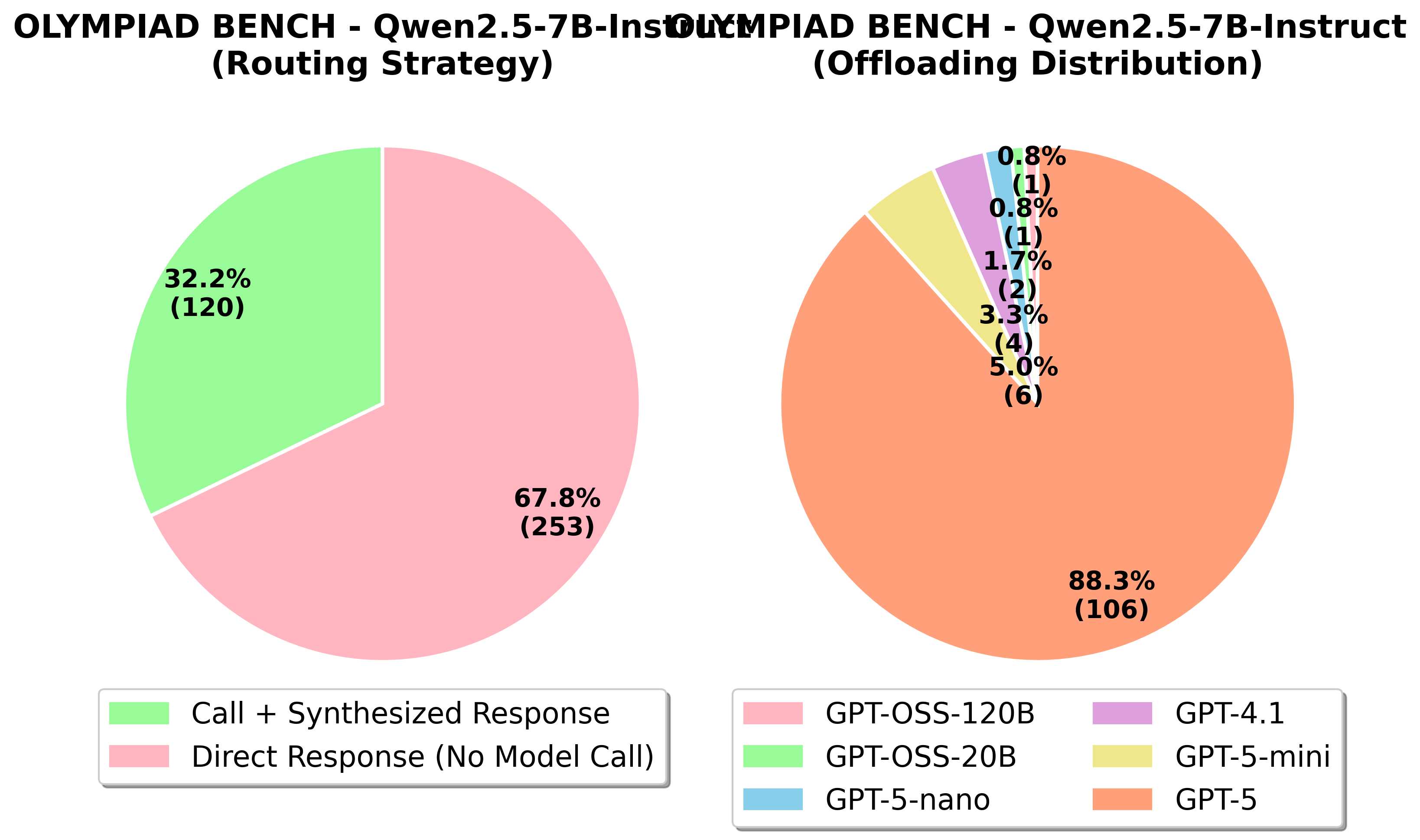} &
        \includegraphics[width=0.24\linewidth]{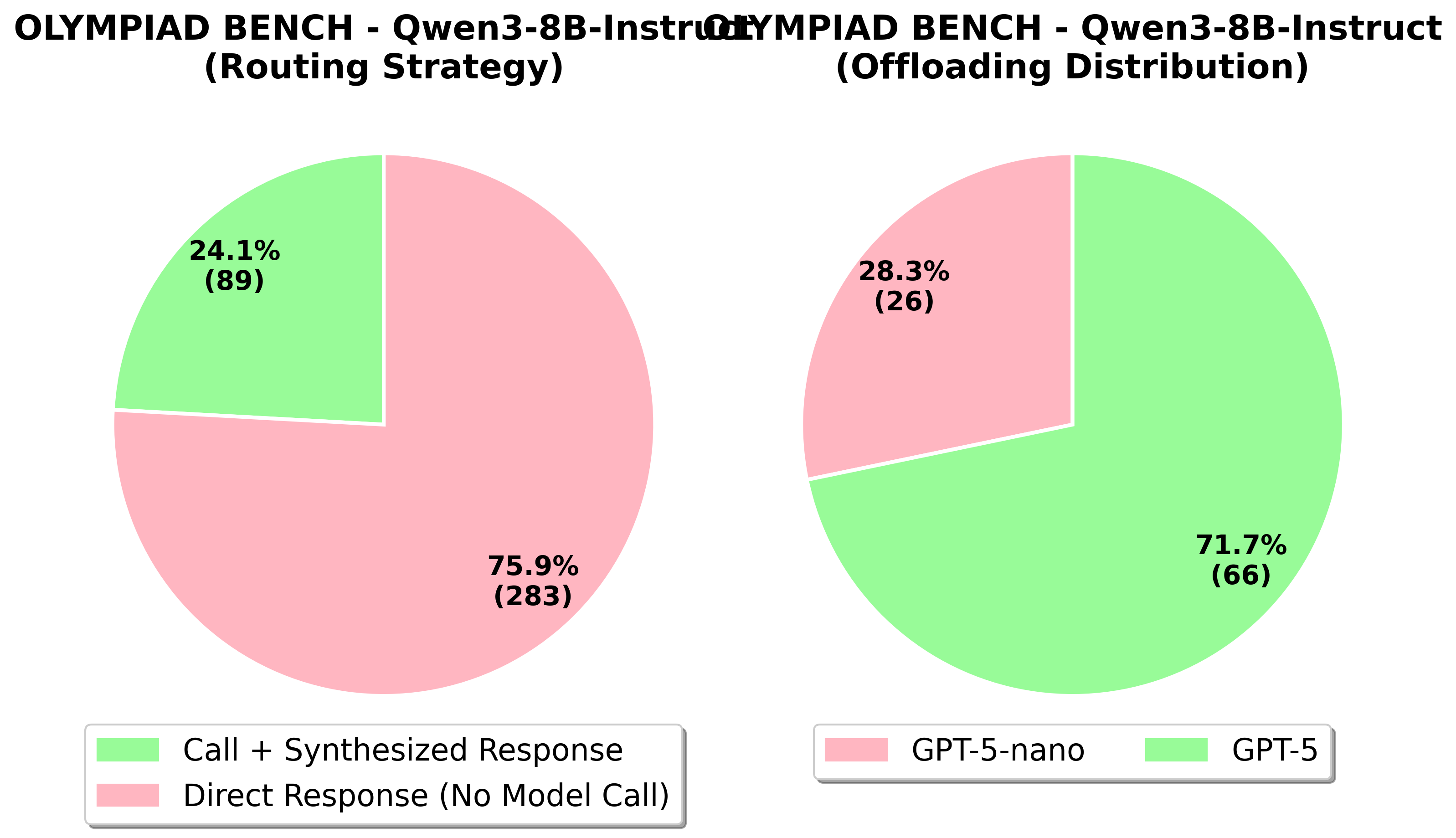} \\
    \end{tabular}

    \caption{Comparison of different router model's routing strategies and offloaded model distributions (\textit{AIME-24, Minerva, Olympiad Bench}).}
    \label{fig:distribution1}
    \vspace{-2mm}
\end{figure*}

\begin{figure*}[t]
    \centering
    \setlength{\tabcolsep}{1pt} 
    \renewcommand{\arraystretch}{1.5} 

    \begin{tabular}{cccc}
        \includegraphics[width=0.24\linewidth]{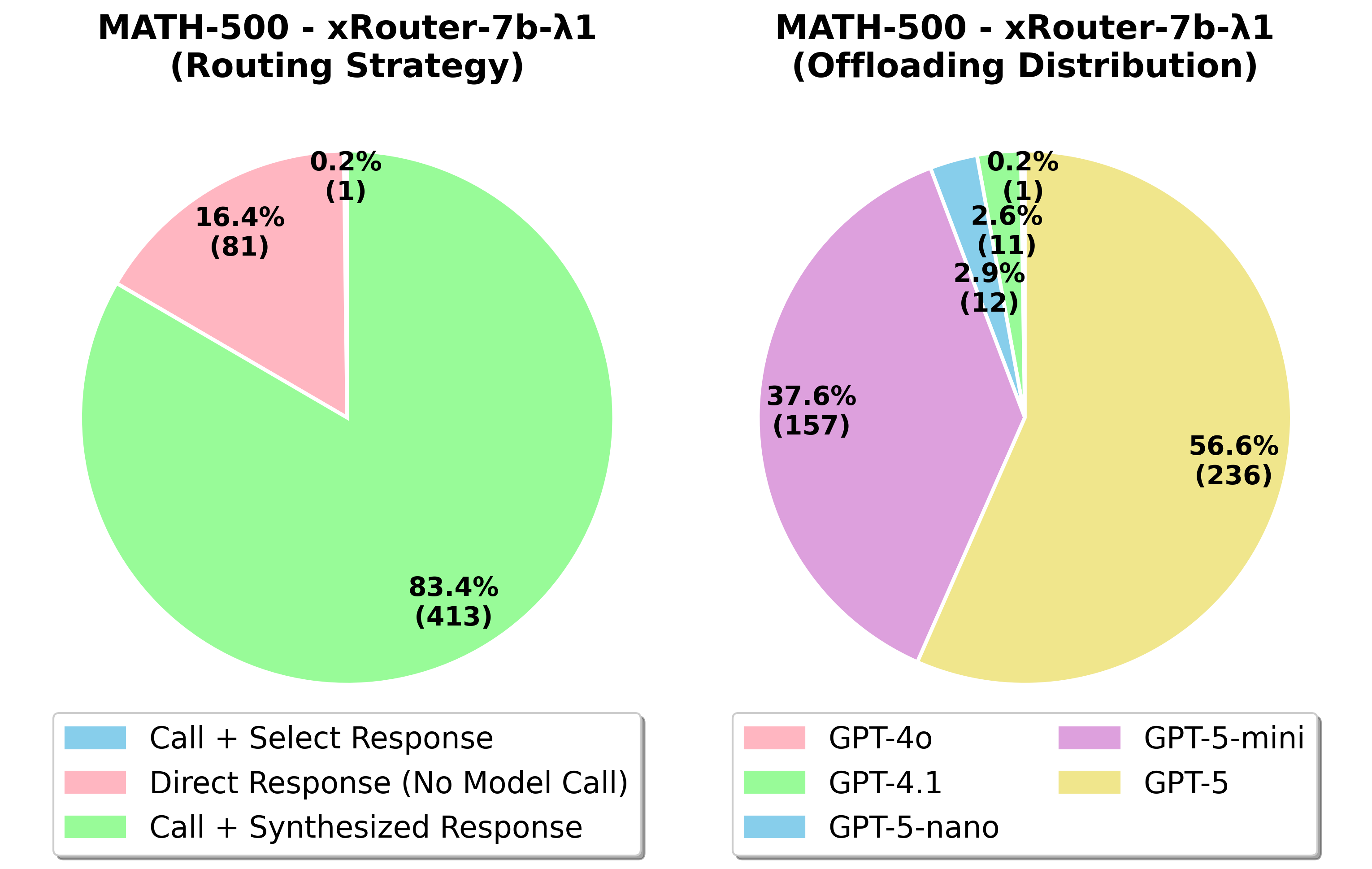} &
        \includegraphics[width=0.24\linewidth]{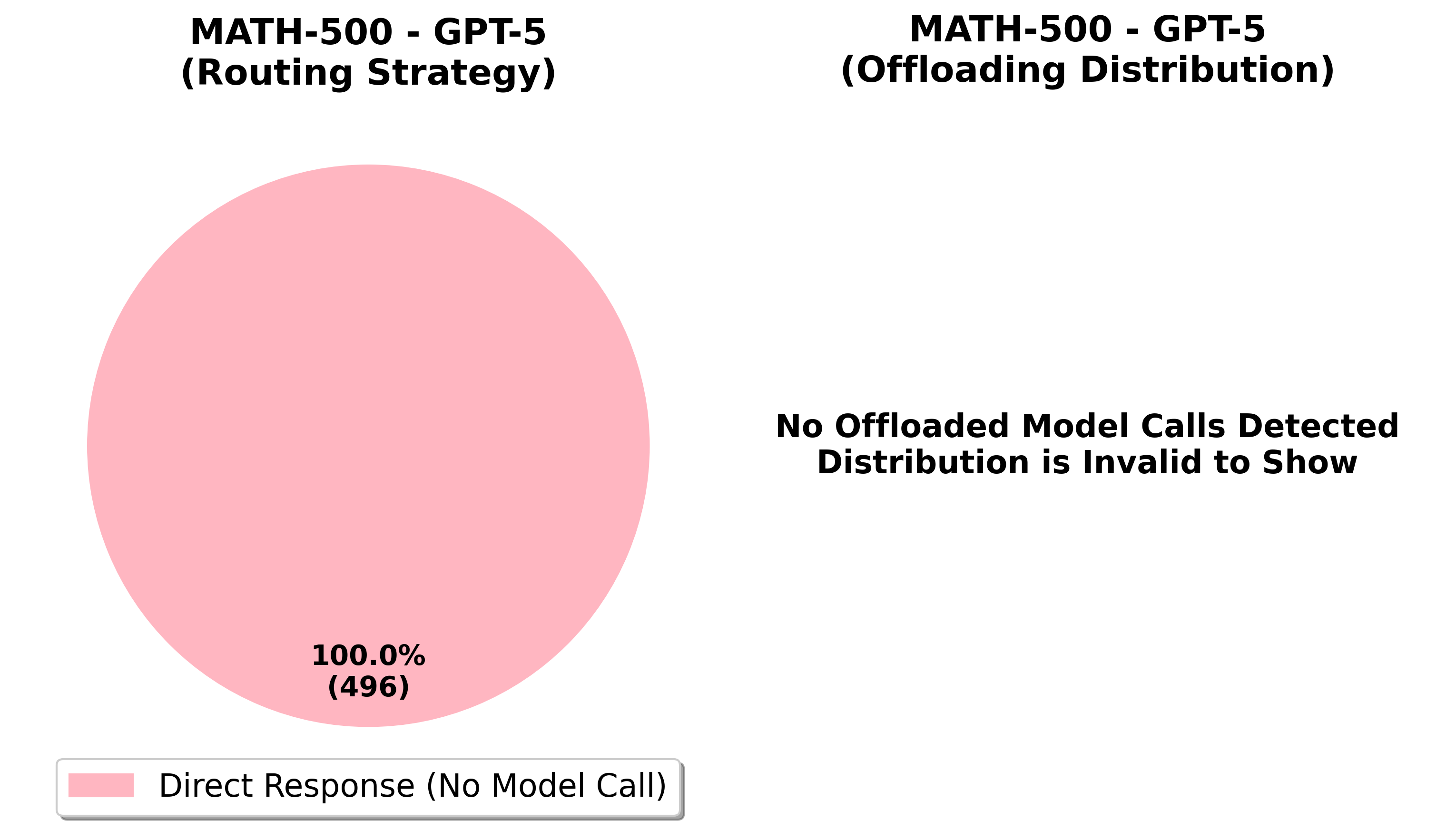} &
        \includegraphics[width=0.24\linewidth]{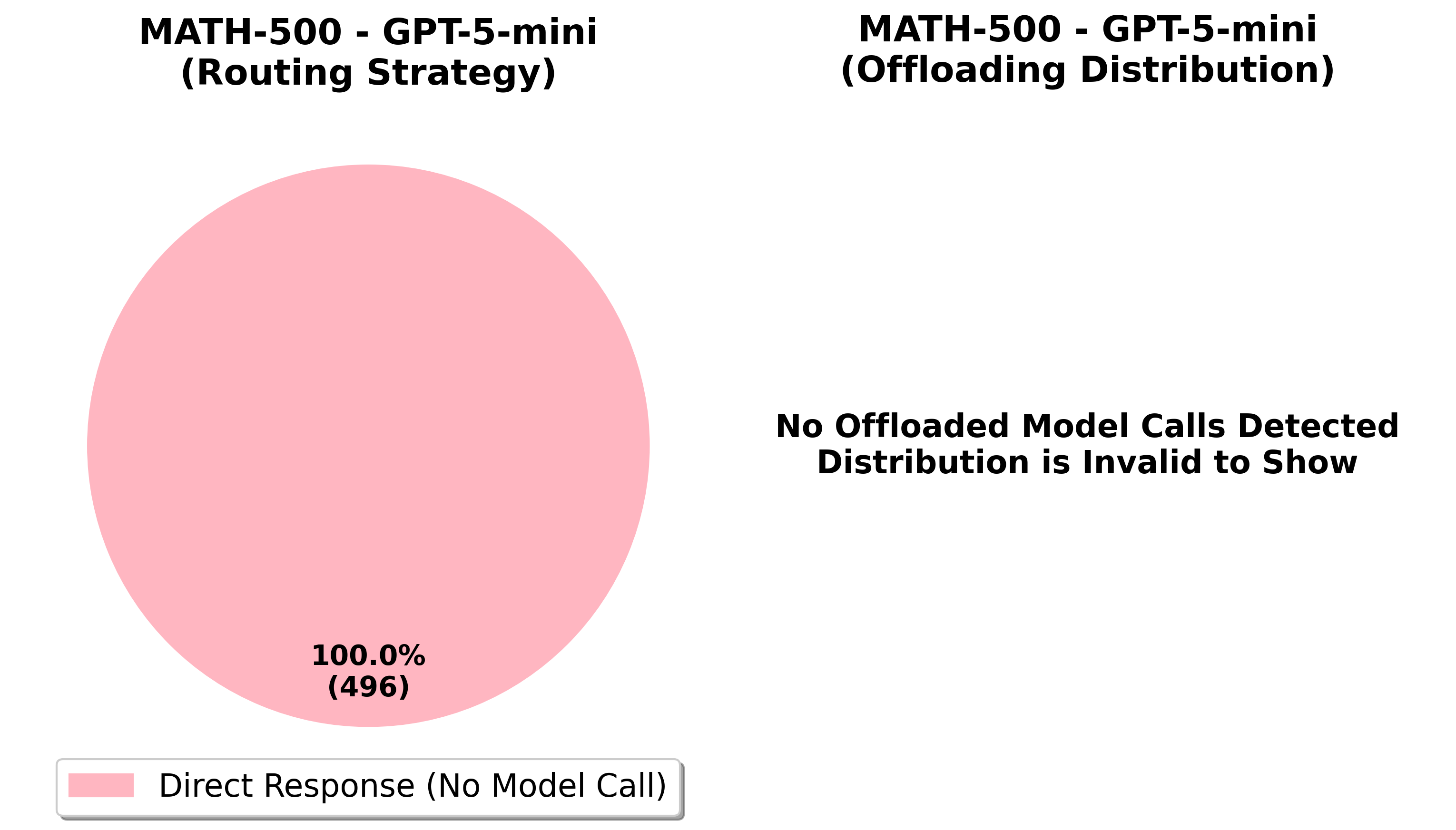} &
        \includegraphics[width=0.24\linewidth]{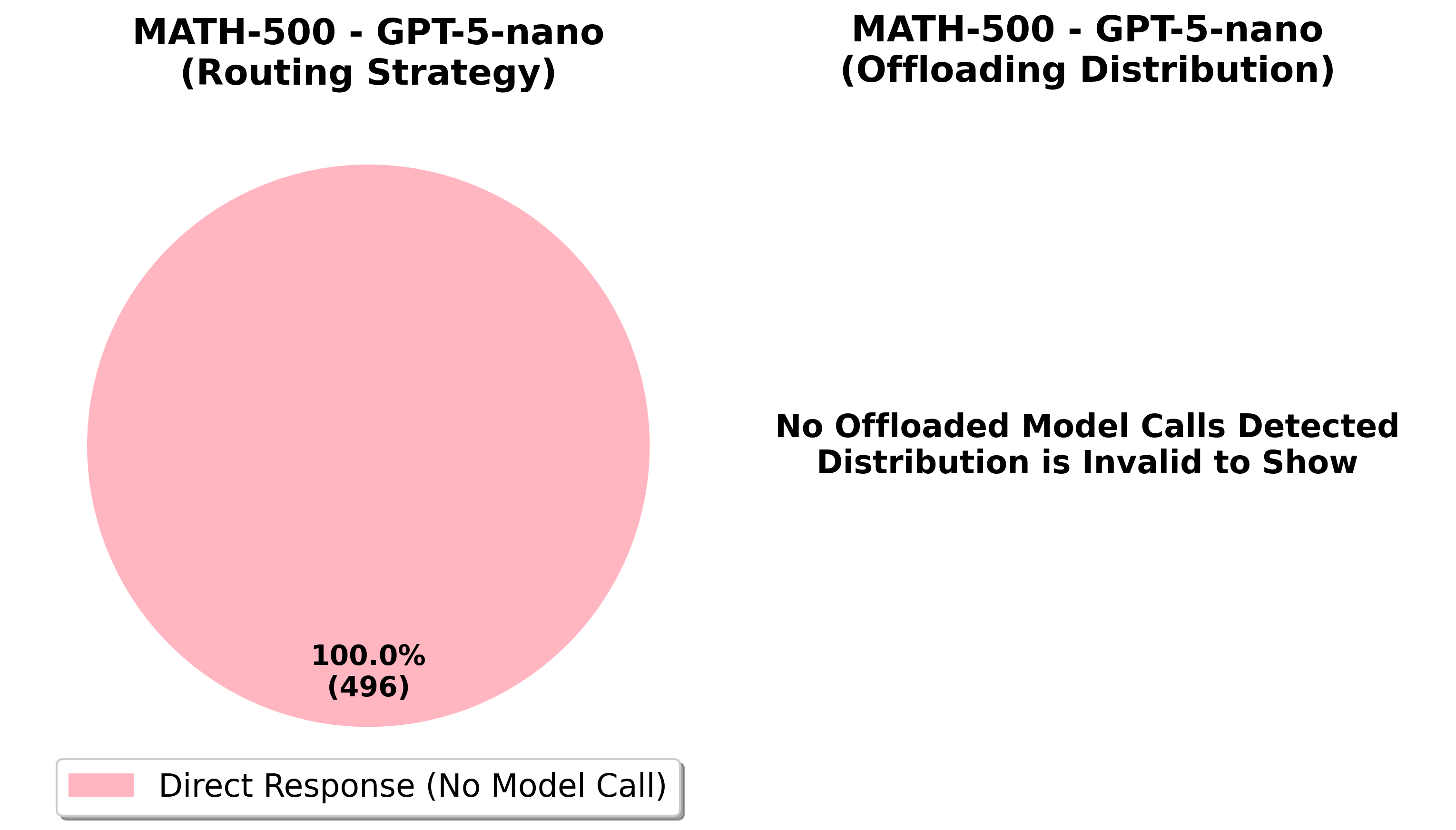} \\
        \includegraphics[width=0.24\linewidth]{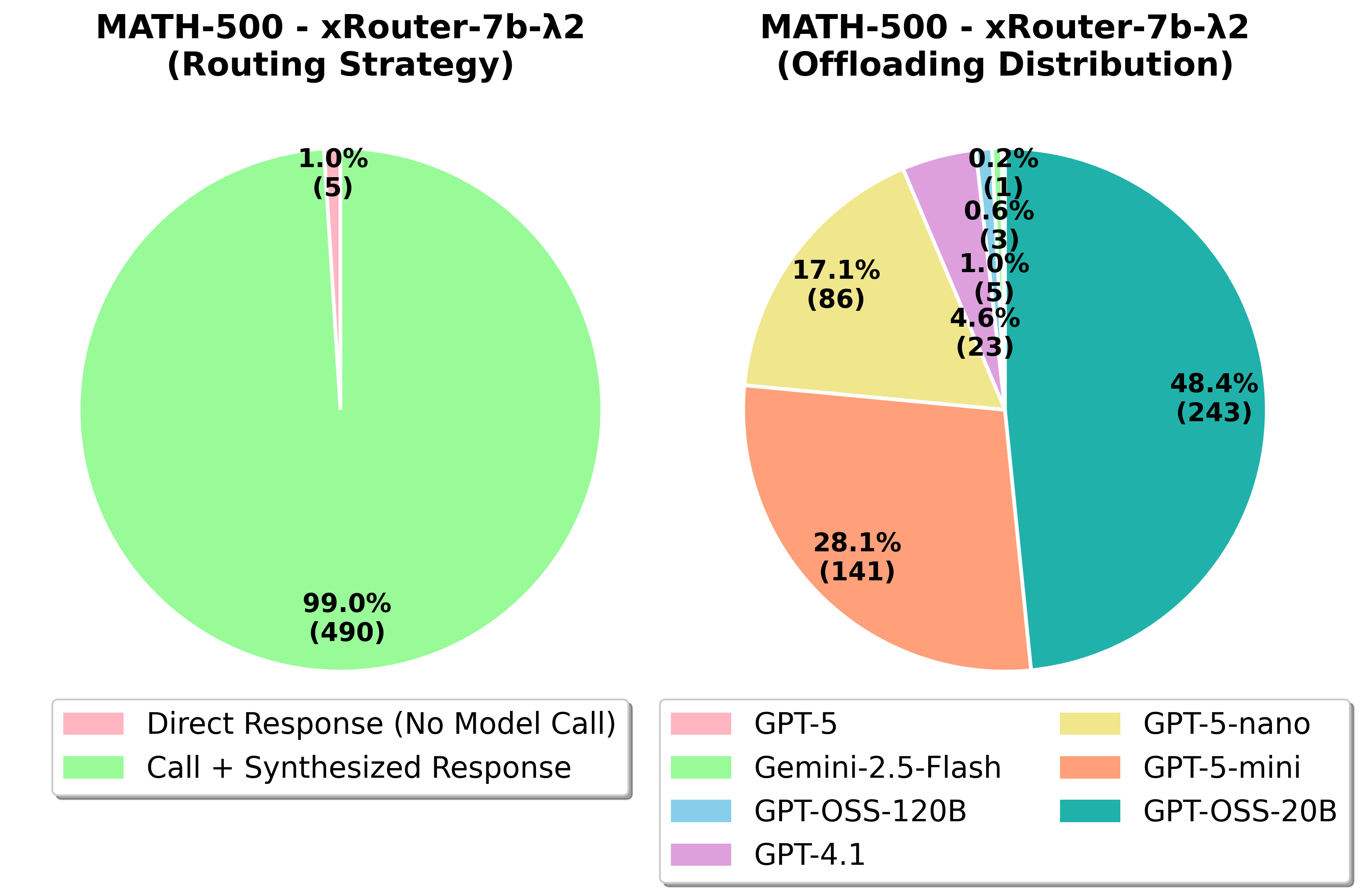} &
        \includegraphics[width=0.24\linewidth]{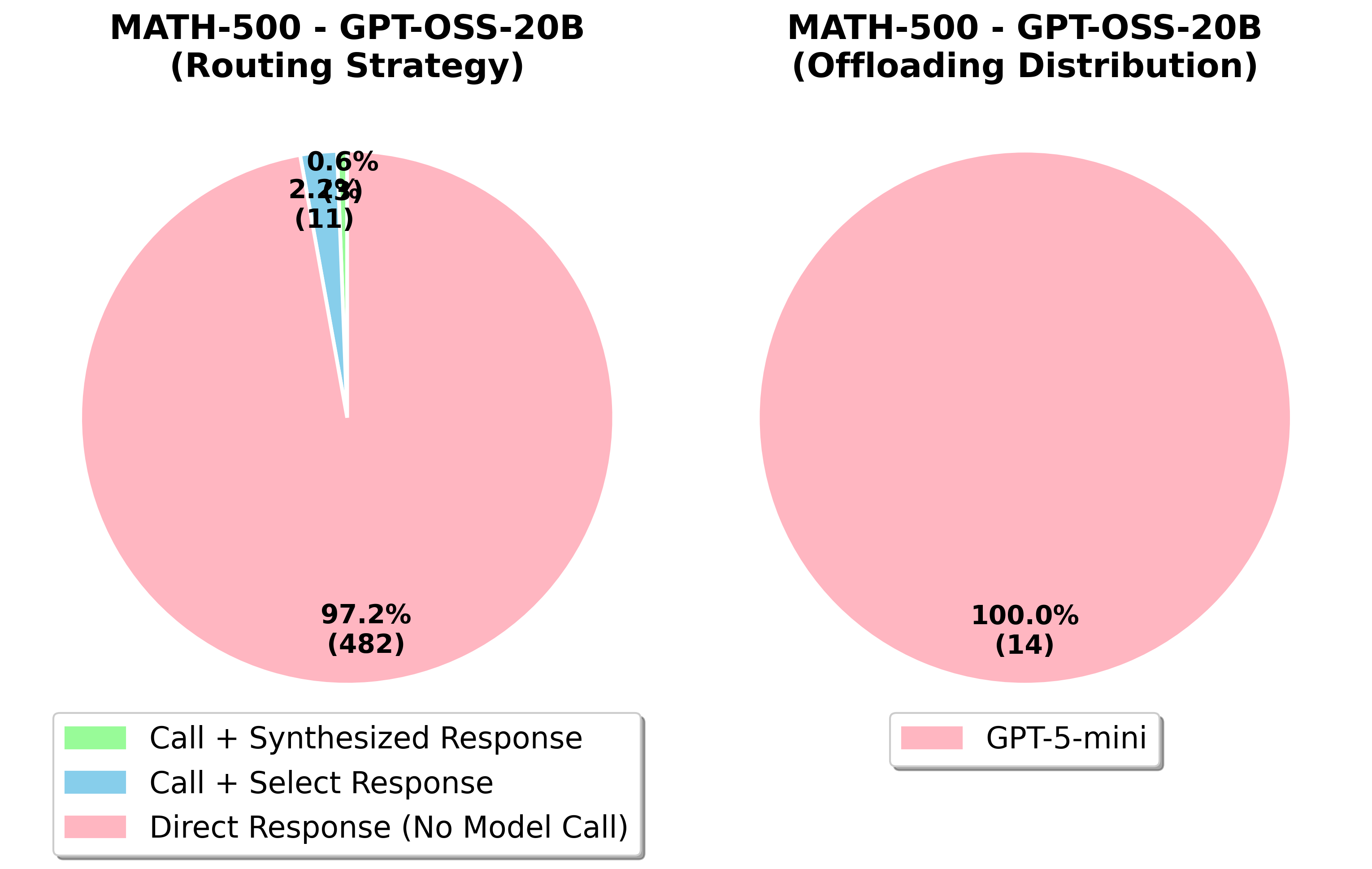} &
        \includegraphics[width=0.24\linewidth]{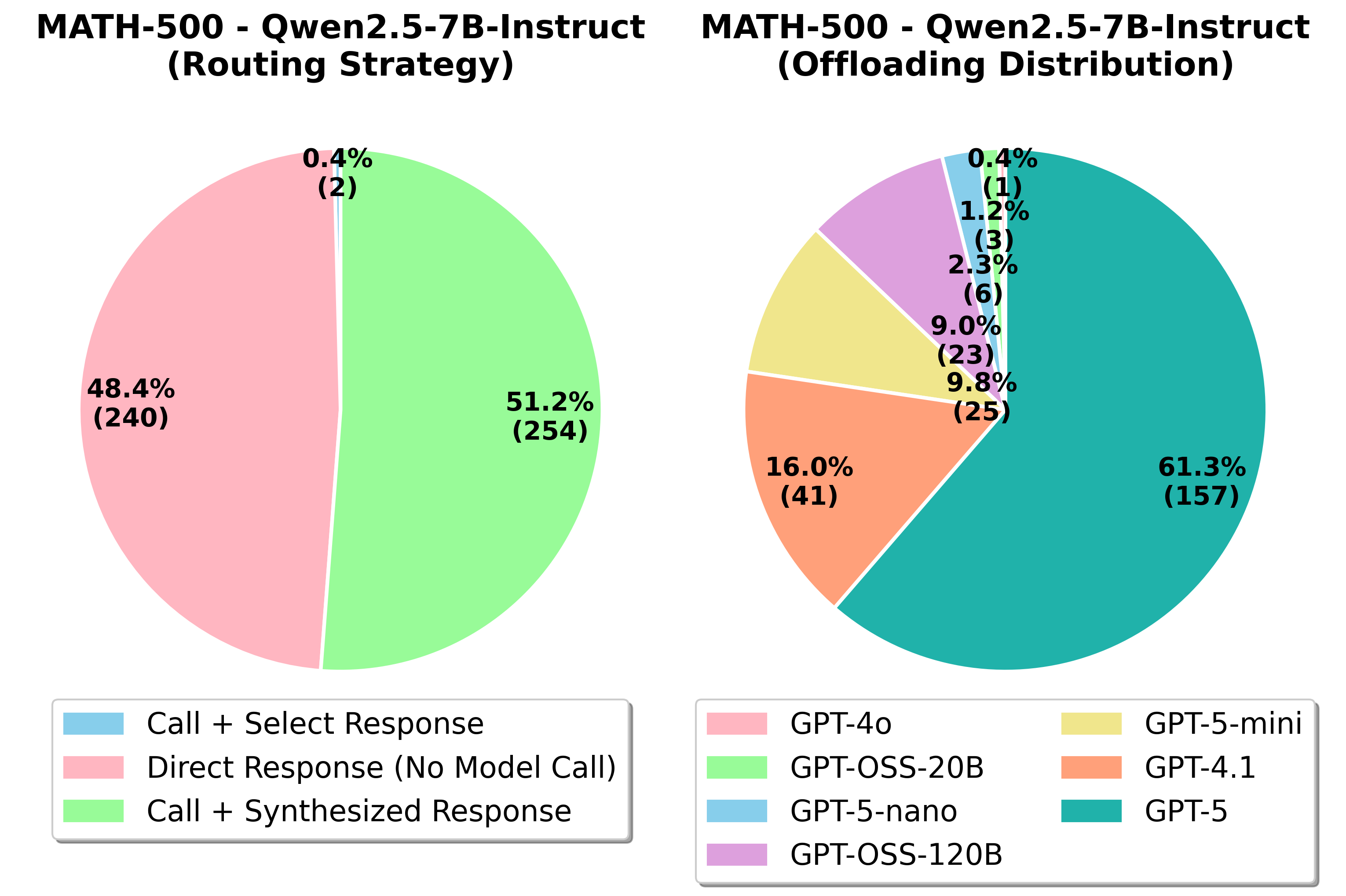} &
        \includegraphics[width=0.24\linewidth]{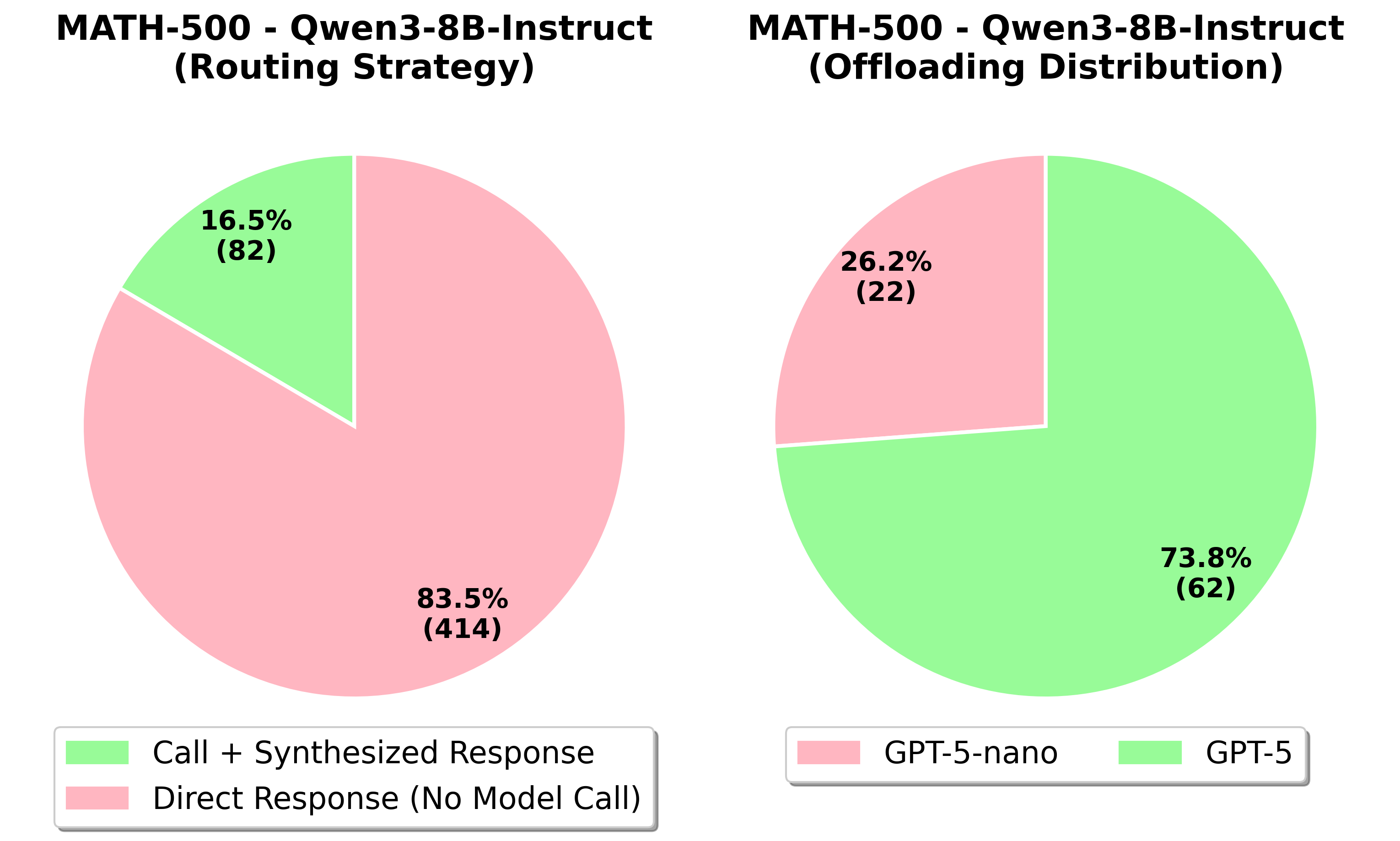} \\

        \includegraphics[width=0.24\linewidth]{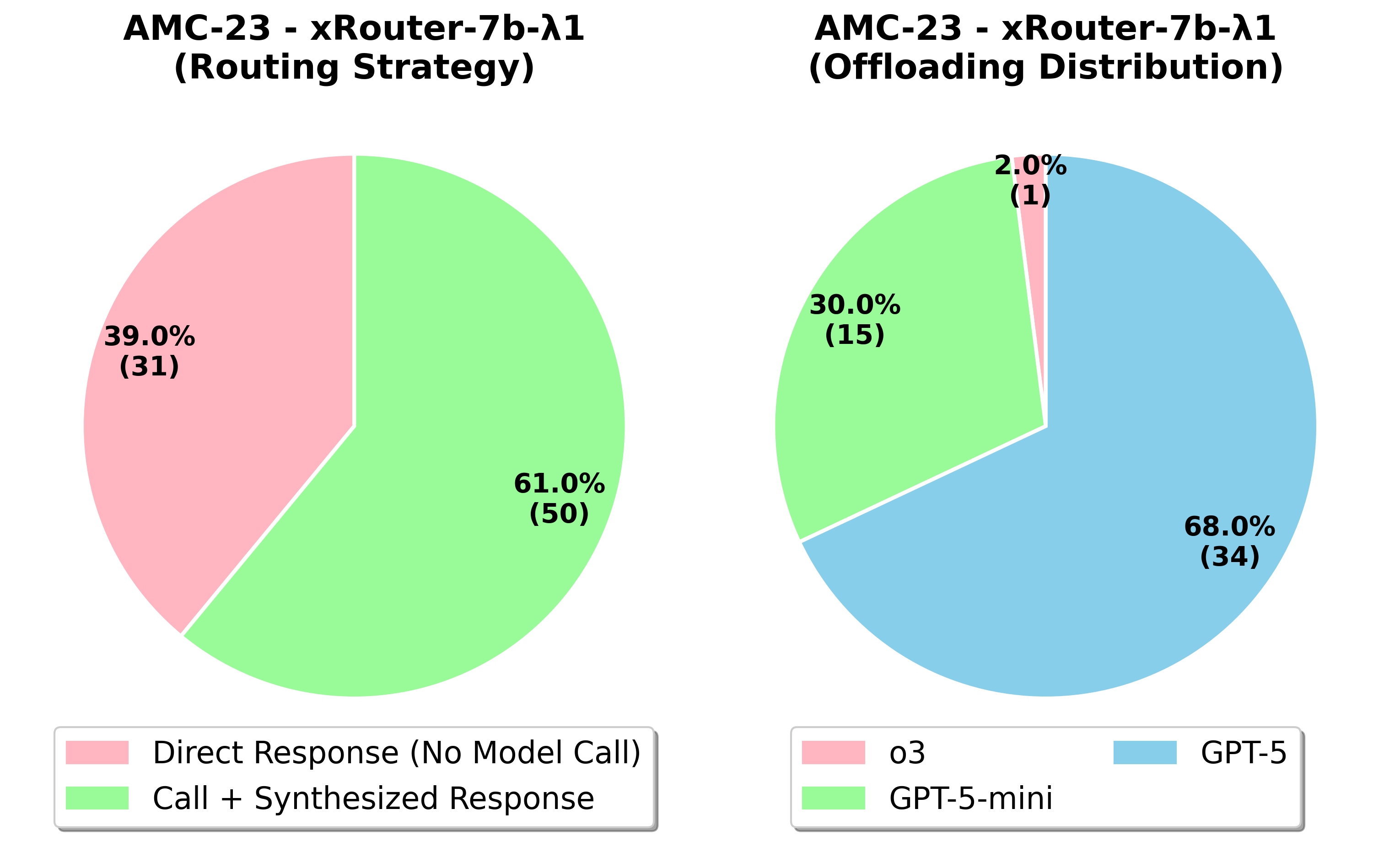} &
        \includegraphics[width=0.24\linewidth]{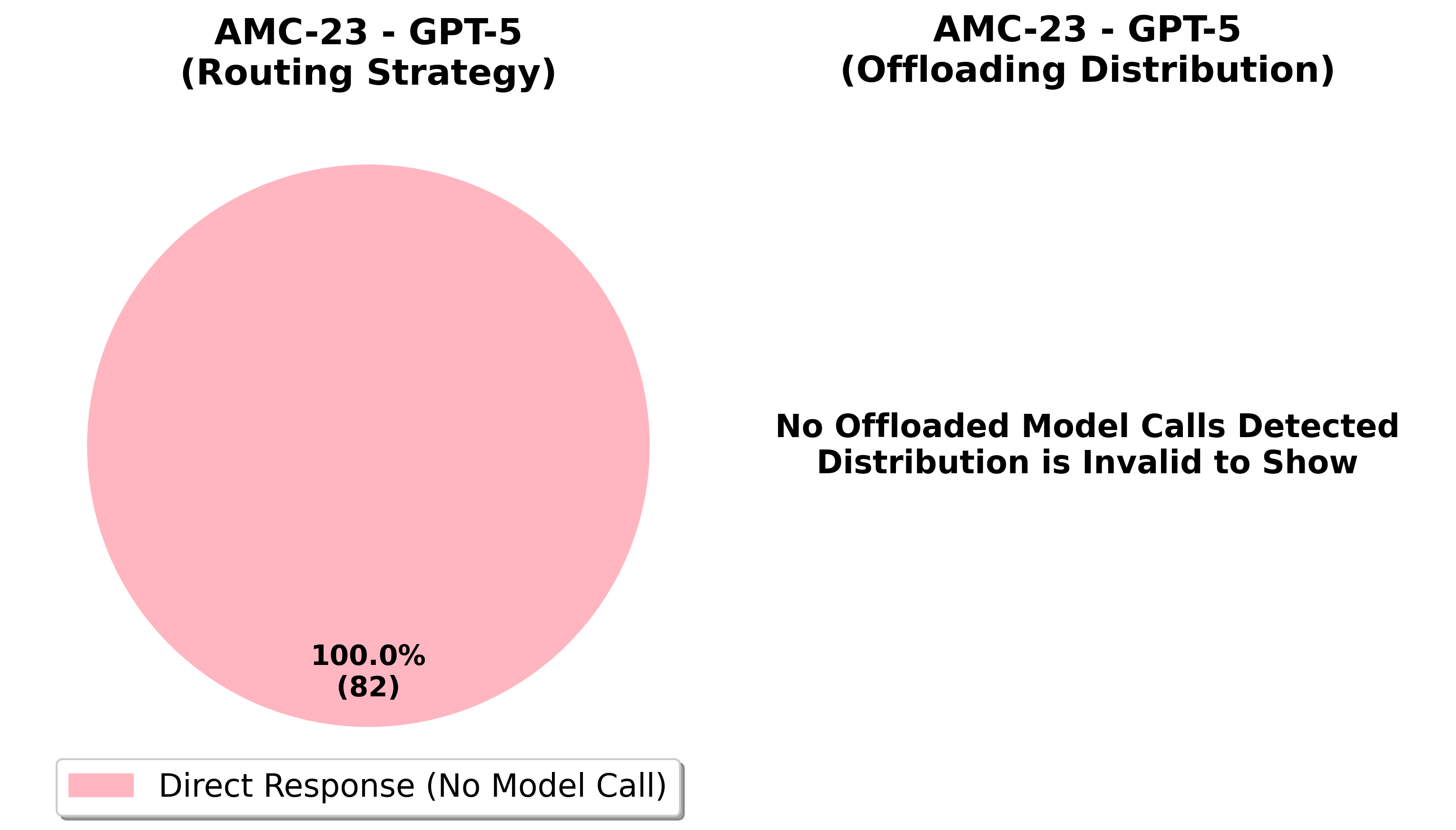} &
        \includegraphics[width=0.24\linewidth]{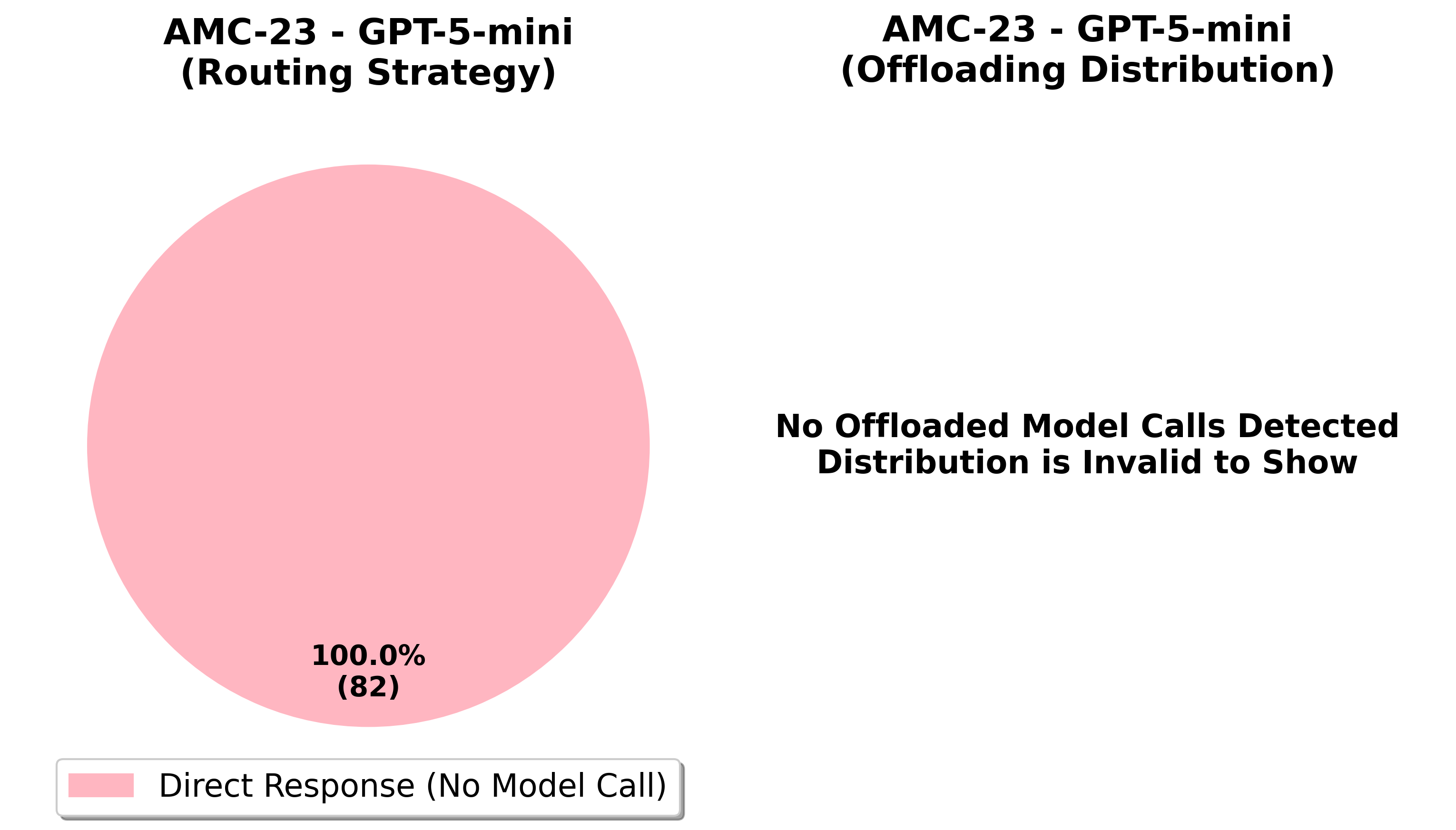} &
        \includegraphics[width=0.24\linewidth]{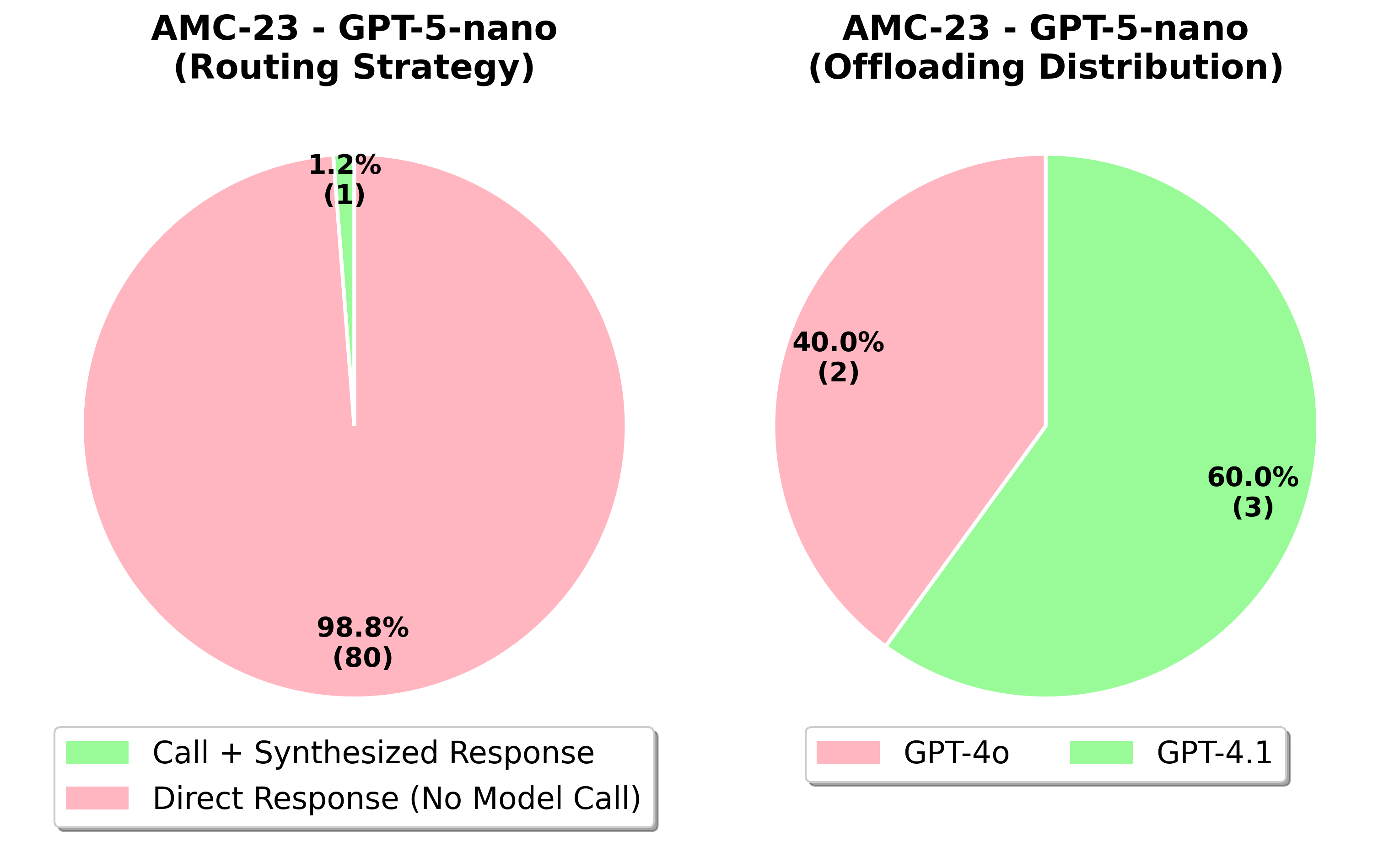} \\
        \includegraphics[width=0.24\linewidth]{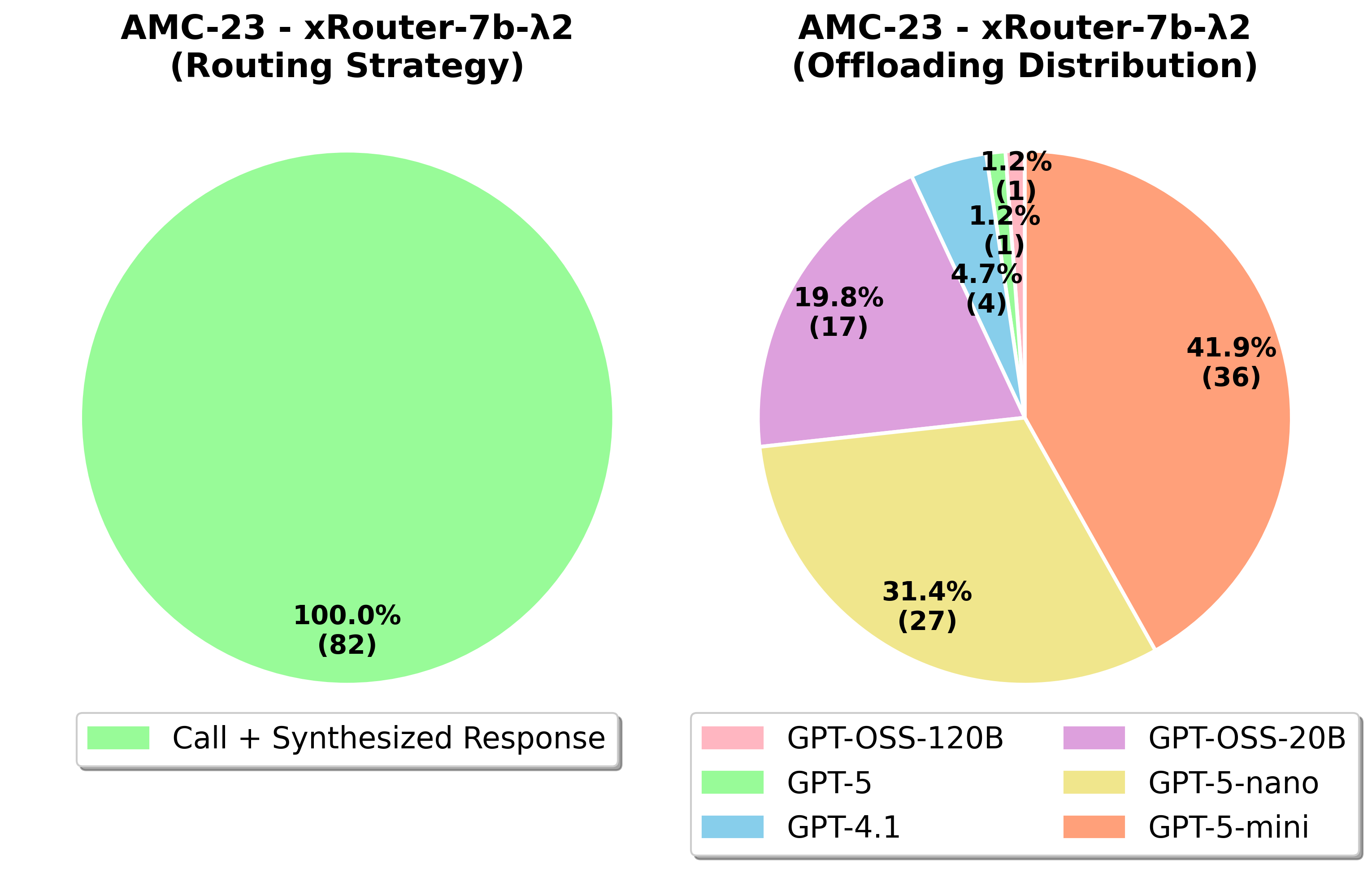} &
        \includegraphics[width=0.24\linewidth]{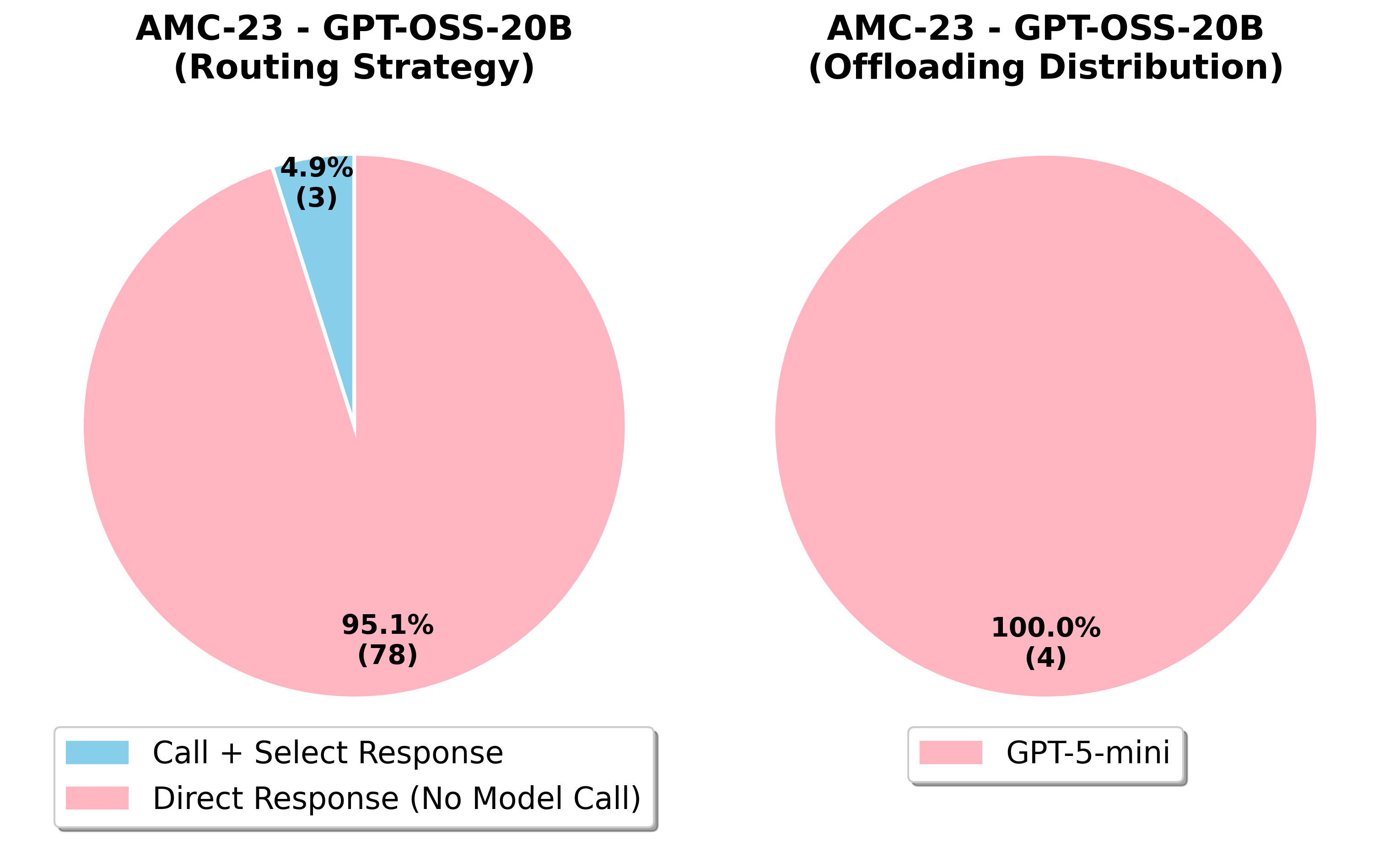} &
        \includegraphics[width=0.24\linewidth]{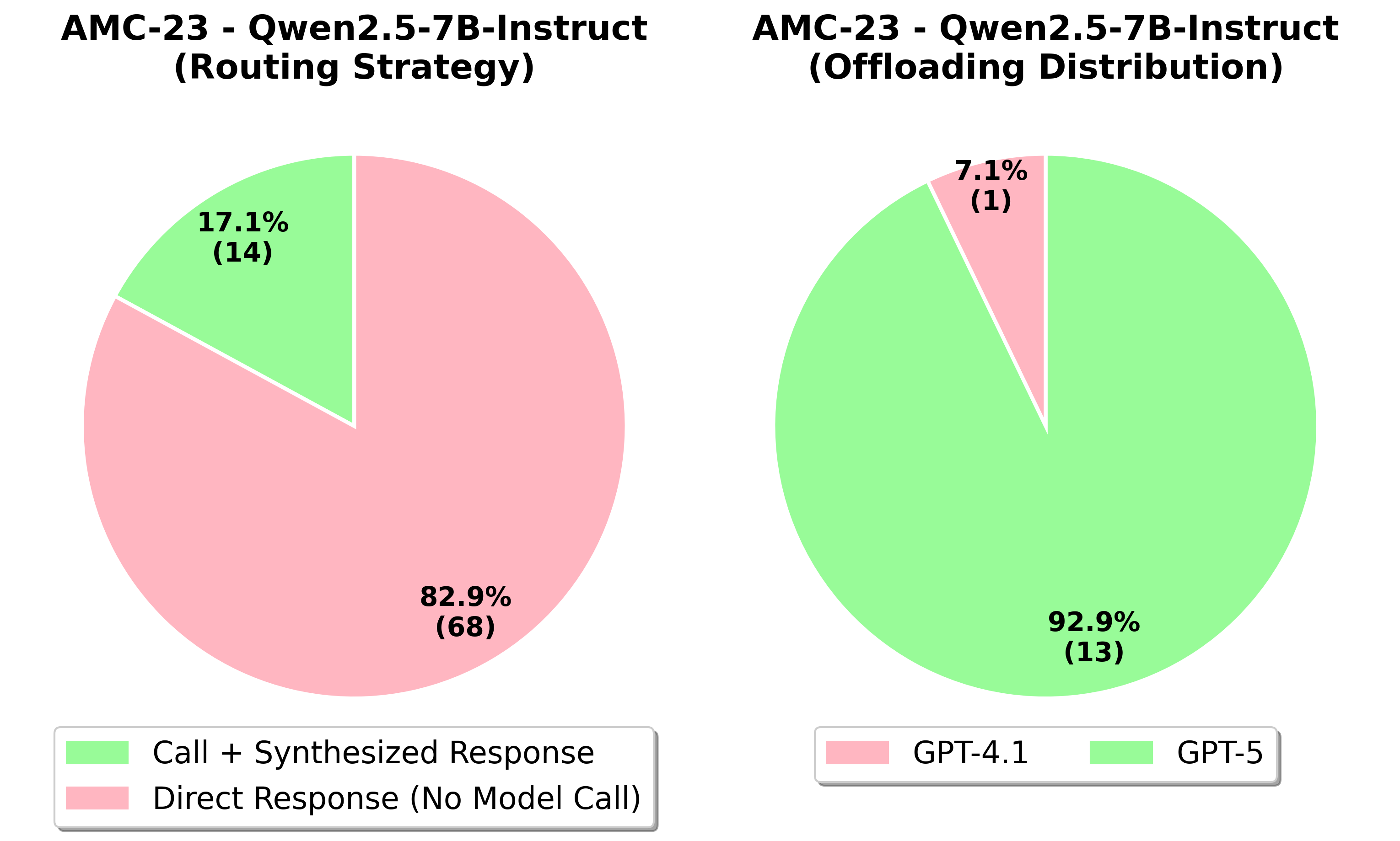} &
        \includegraphics[width=0.24\linewidth]{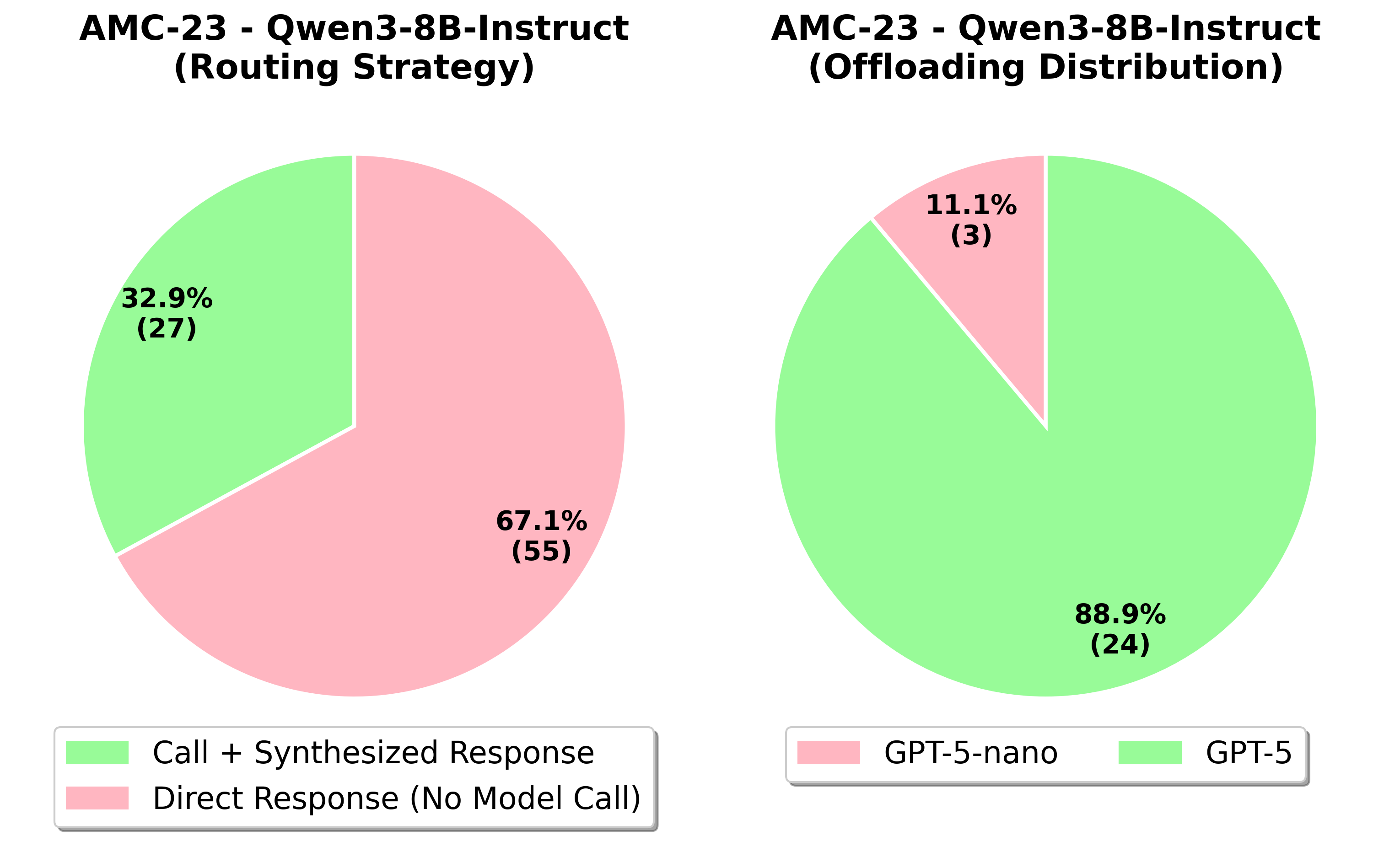} \\

        \includegraphics[width=0.24\linewidth]{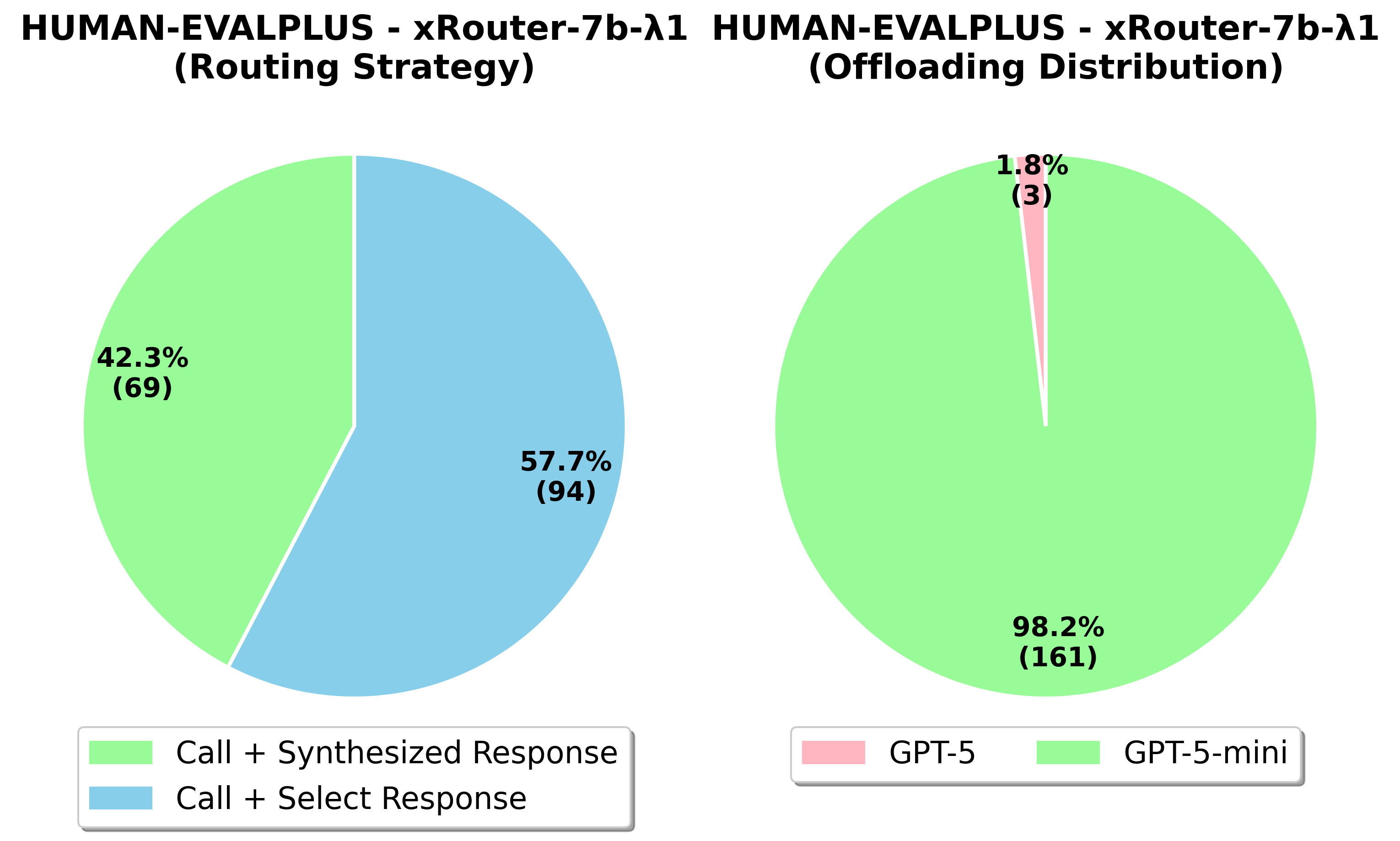} &
        \includegraphics[width=0.24\linewidth]{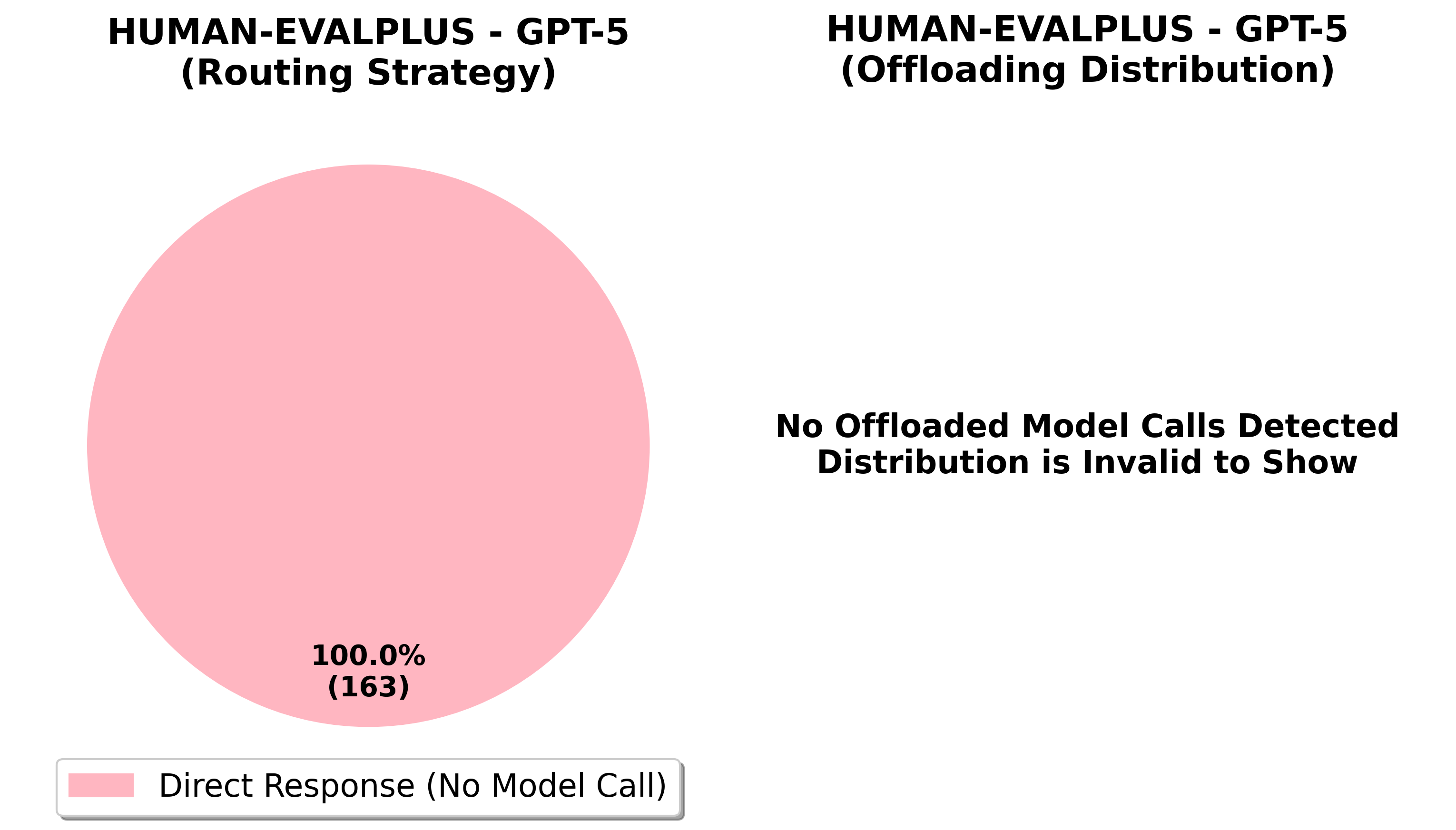} &
        \includegraphics[width=0.24\linewidth]{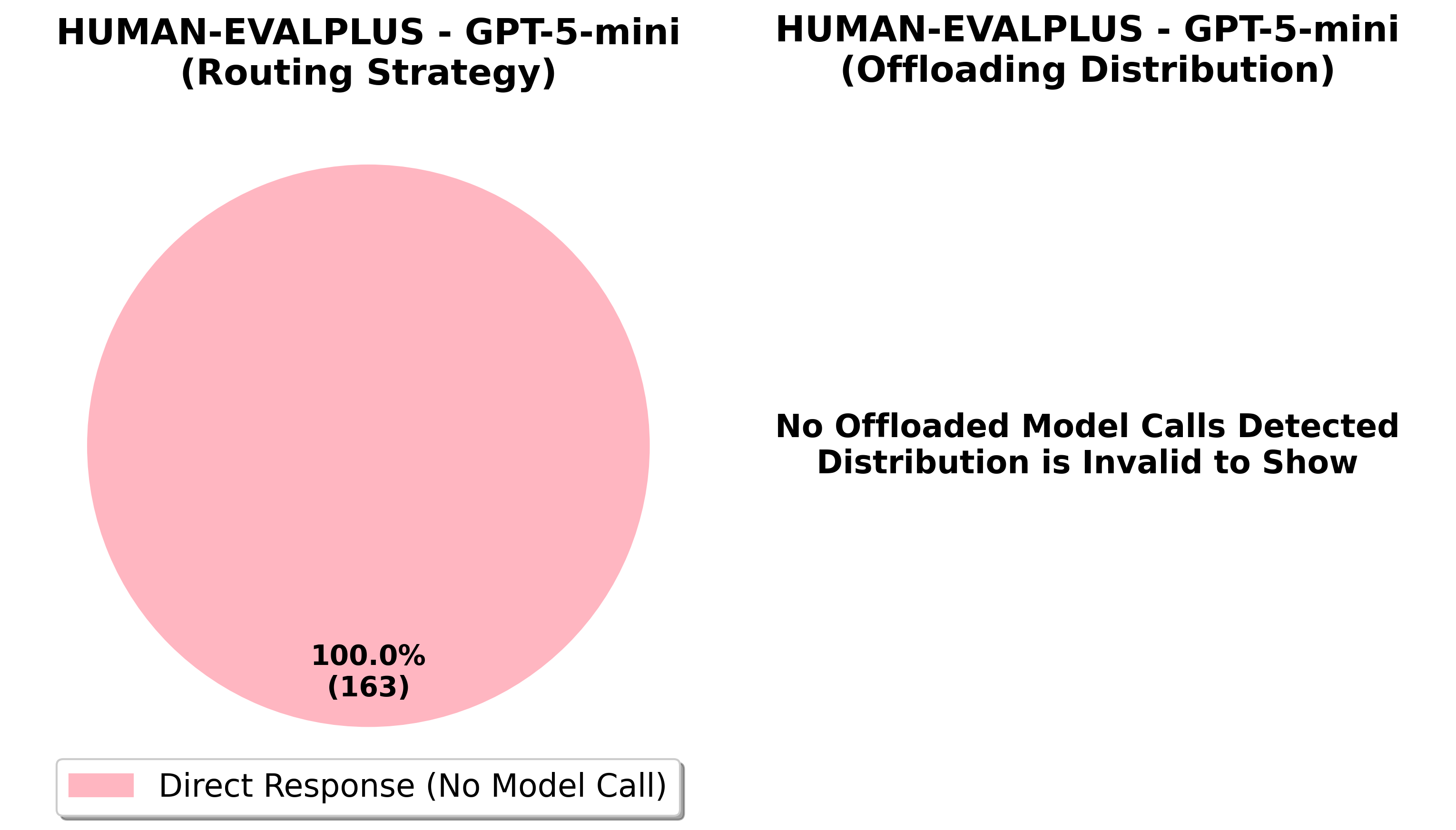} &
        \includegraphics[width=0.24\linewidth]{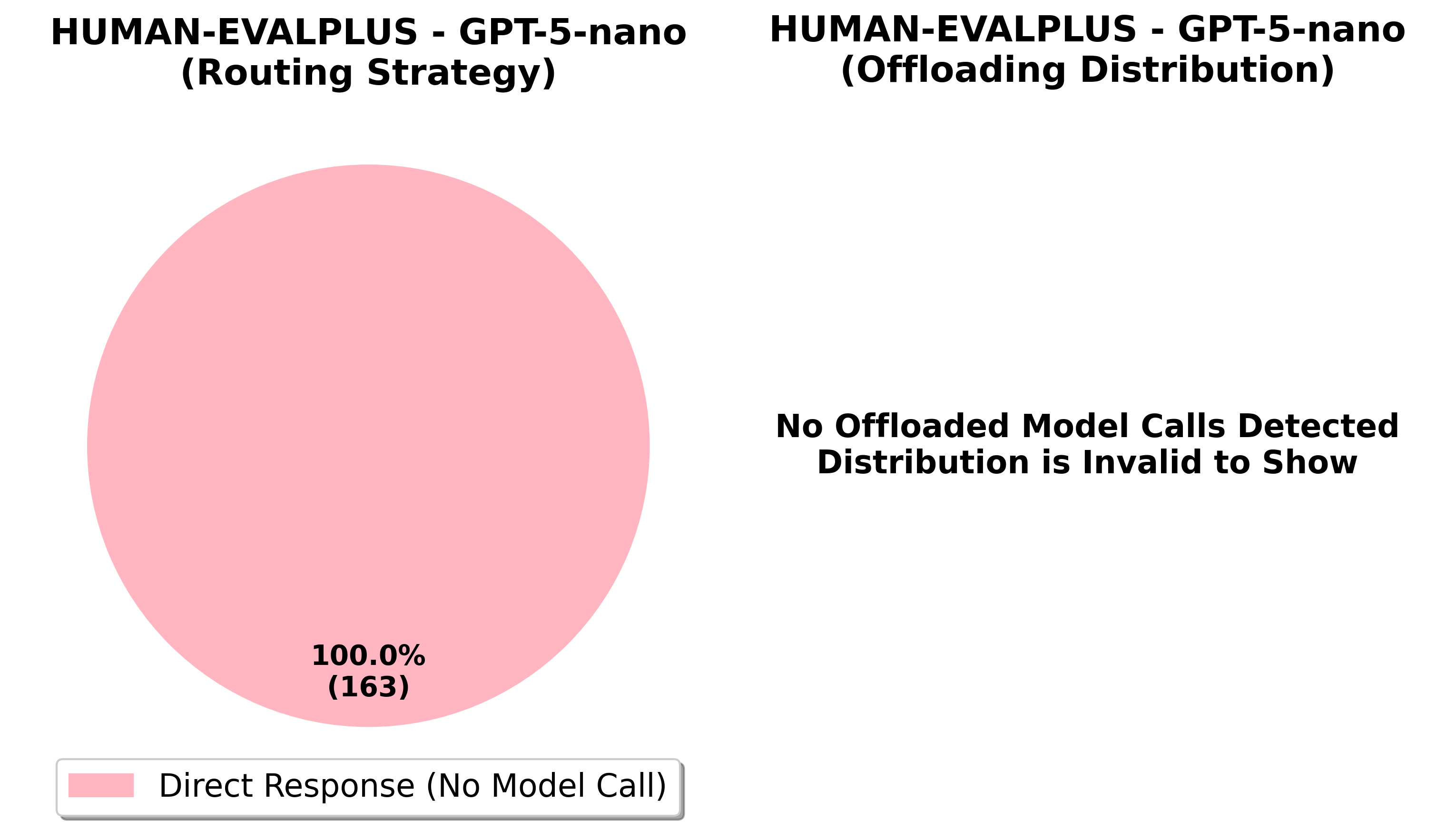} \\
        \includegraphics[width=0.24\linewidth]{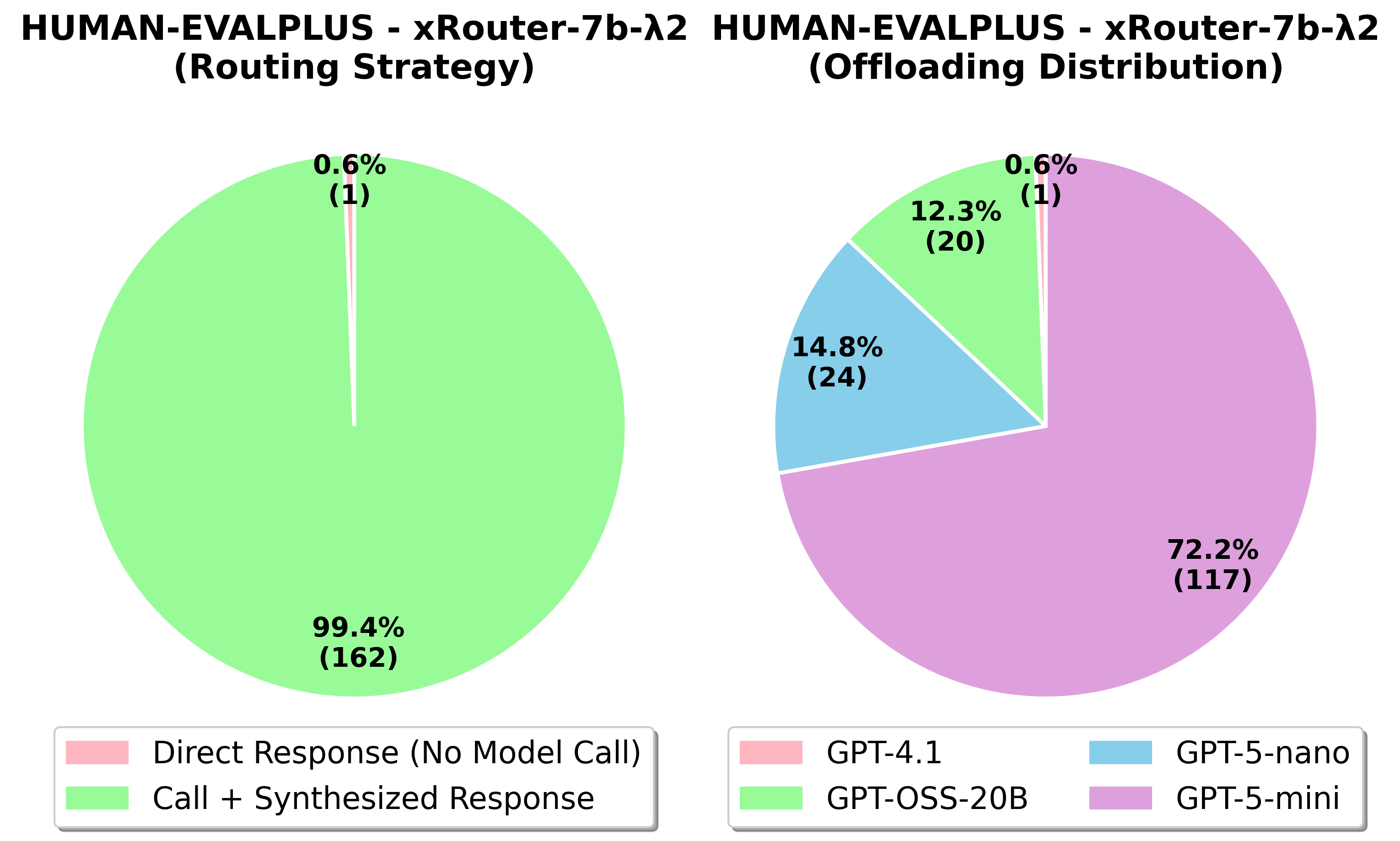} &
        \includegraphics[width=0.24\linewidth]{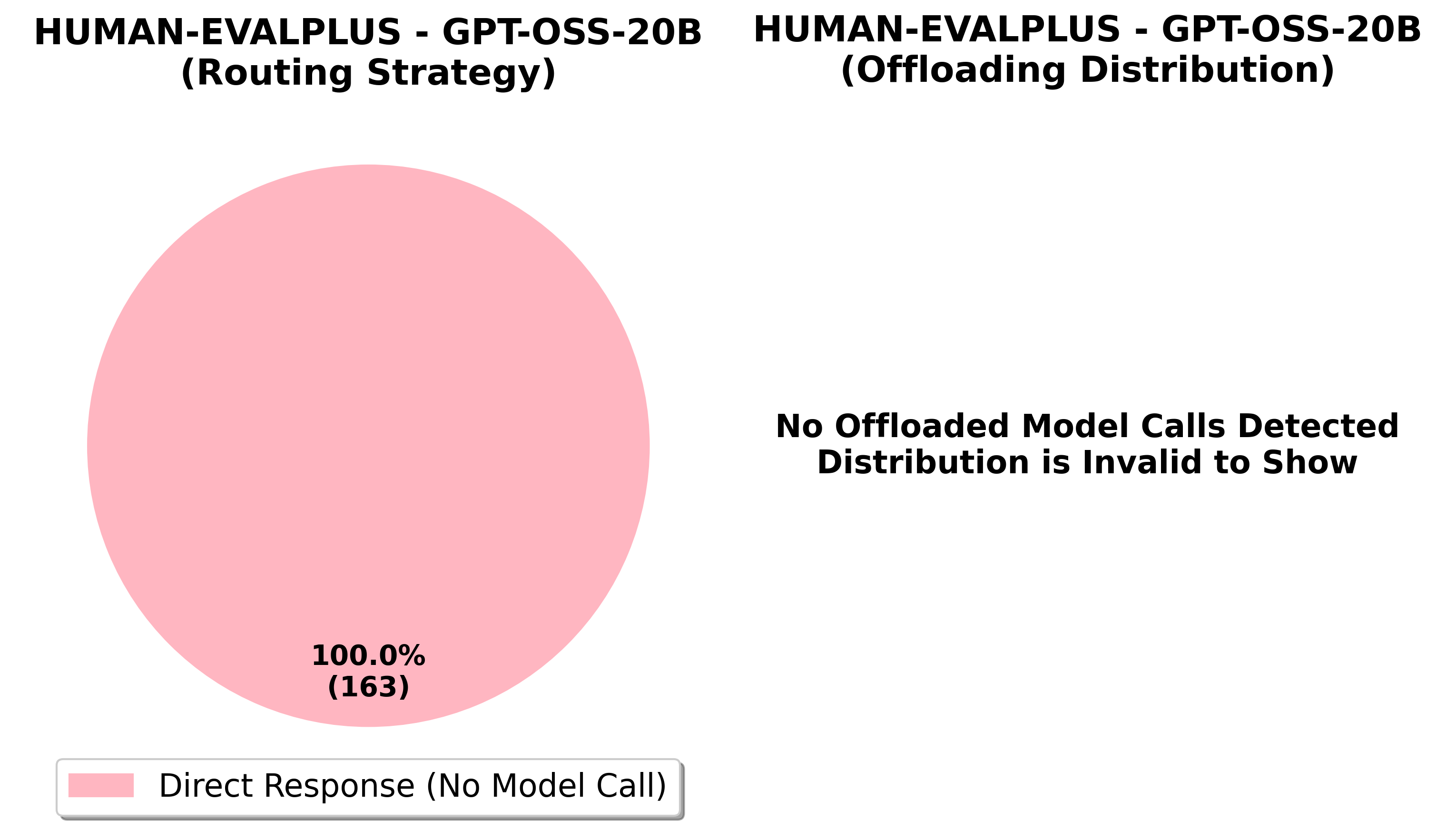} &
        \includegraphics[width=0.24\linewidth]{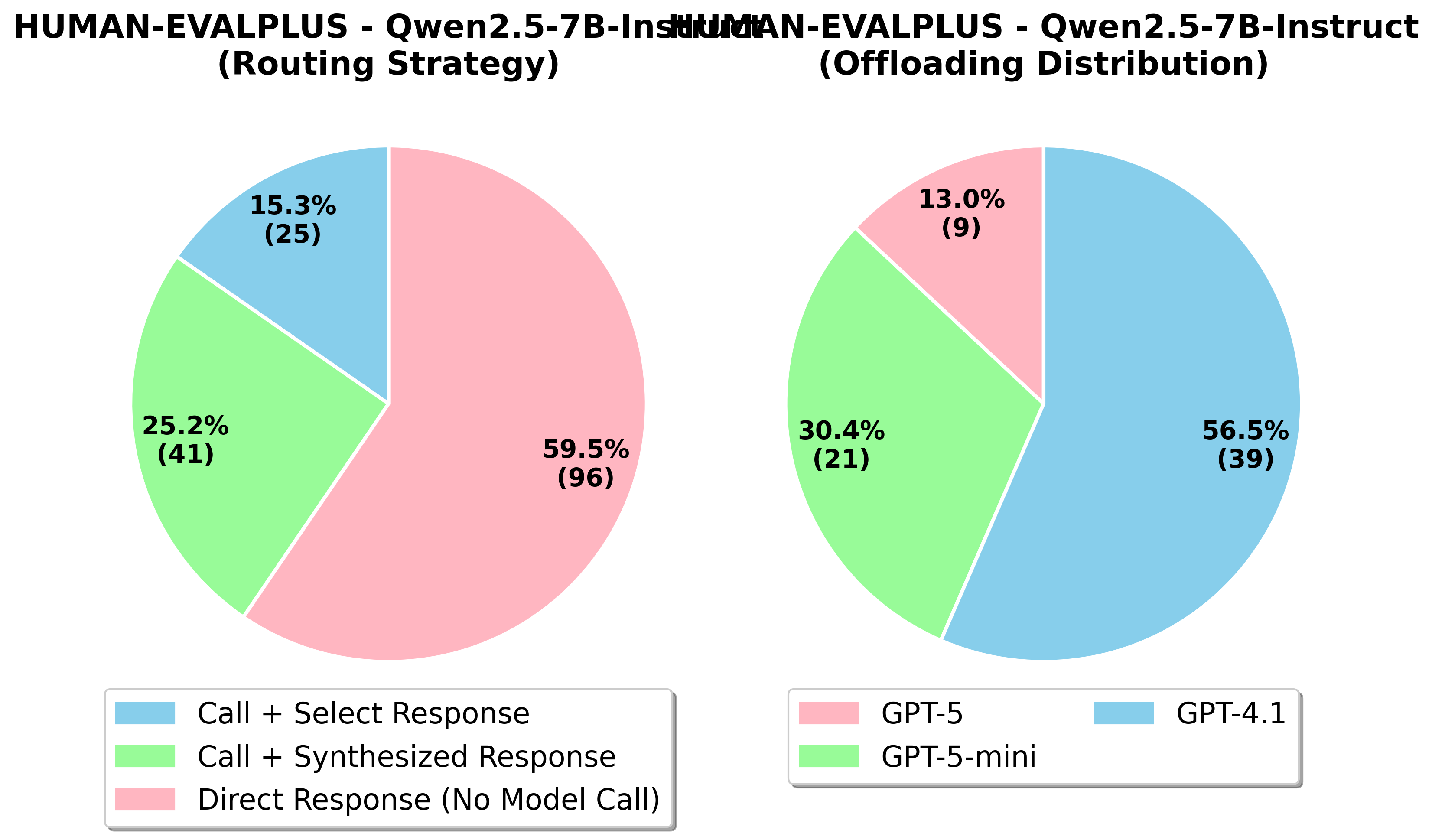} &
        \includegraphics[width=0.24\linewidth]{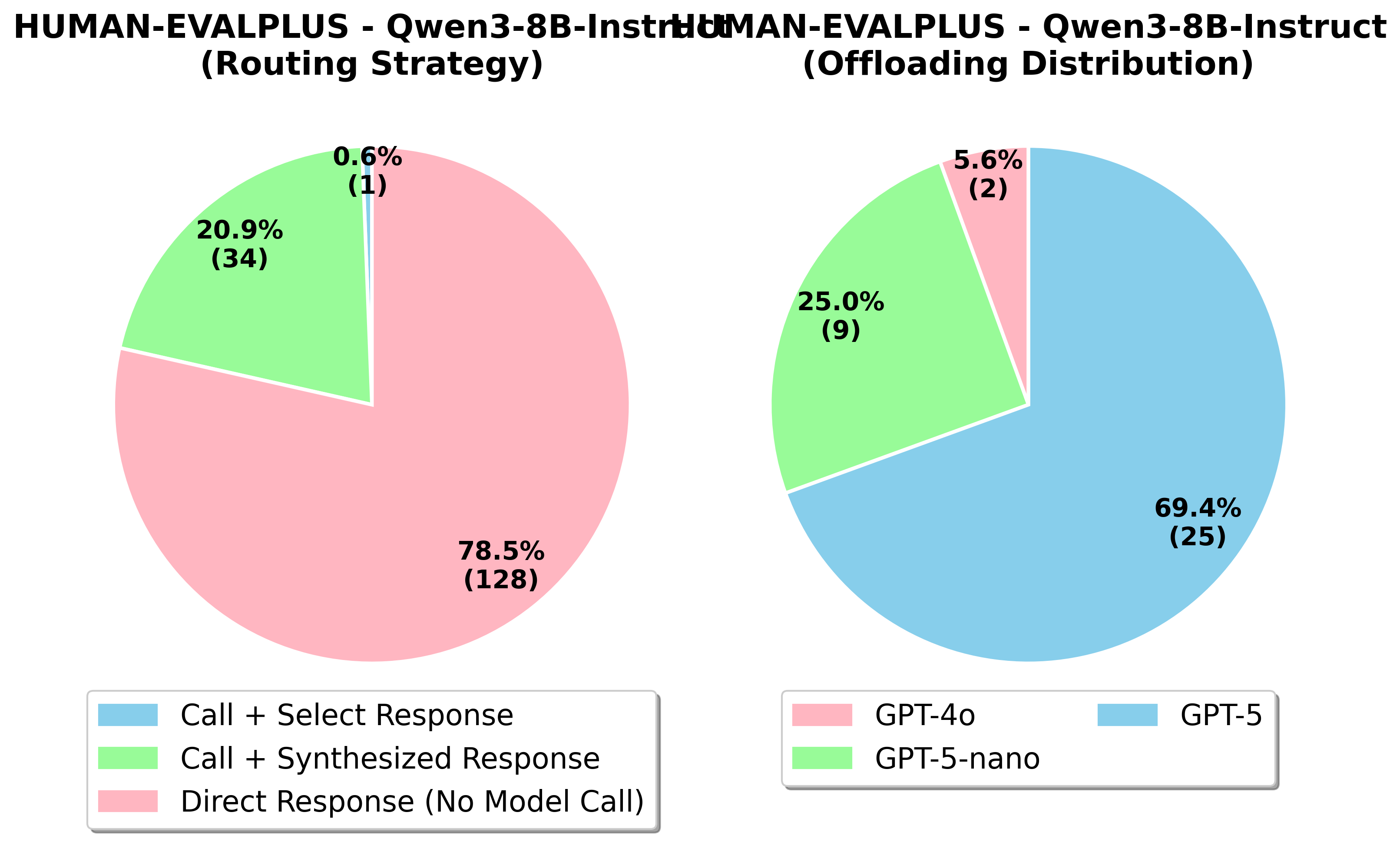} \\
    \end{tabular}

    \caption{Comparison of different router model's routing strategies and offloaded model distributions (\textit{MATH-500, AMC-23, Human-EvalPlus})}
    \label{fig:distribution2}
    \vspace{-2mm}
\end{figure*}

\paragraph{Distribution of Routing Strategies.}
Our evaluation supports multiple strategies for how a router model returns its final answer. Specifically, there are three distinct modes: 
(1) \textbf{Direct Response}, in which the router determines that it can handle the query independently without invoking any downstream models; 
(2) \textbf{Calling + Synthesized Response}, where the router delegates the query to downstream models, collects their responses, and then synthesizes a final answer based on its own reasoning; and 
(3) \textbf{Calling + Select Response}, in which the router also offloads the query but instead of generating its own summary, directly selects one of the downstream responses as the final output. The selection mechanism is implemented through a special function call that specifies which downstream model’s response should be adopted. When the evaluation system detects such a function call, it automatically retrieves the designated response and returns it as the router’s final submission.

As shown in \Cref{fig:distribution1} and \Cref{fig:distribution2}, the left pie chart in each subfigure illustrates the routing strategy distribution for different models when serving as routers. The leftmost subfigure in each row represents our trained router model. We observe several trends in the following: 
(1) our trained router demonstrates a more balanced and adaptive use of multiple strategies, primarily alternating between direct responses and synthesized responses; 
(2) in contrast, off-the-shelf models (both open- and closed-source) tend to favor direct responses, even when explicitly instructed to route queries when uncertain. This tendency is especially pronounced among the GPT series (GPT-4o, GPT-4.1, and GPT-5); and 
(3) all models rarely adopt the select-response strategy. This pattern was consistently observed during router training as well. One plausible explanation is that current models inherently prefer to construct their own answers rather than fully relying on the responses of downstream models. However, the select-response mechanism is potentially more \emph{energy-efficient}, as it avoids the additional computation and reasoning required for synthesis. Developing router models that can dynamically determine when to trust and adopt downstream responses could therefore further reduce computational cost and cognitive load. We leave the exploration of this adaptive trust and routing behavior to future work.

\paragraph{Distribution of Offloaded Models.}
We further analyze the distribution of downstream models that different routers choose to offload queries to. When a router decides to handle all queries within a task independently through direct responses, no downstream models are invoked. For these cases, we represent the corresponding pie charts as white regions in \Cref{fig:distribution1} and \Cref{fig:distribution2}. The right-hand side of each subfigure illustrates the distribution of offloaded models used by the router.

Several trends emerge from this analysis:
(1) Our trained router models exhibit notably more diverse offloading behaviors, often invoking a wide range of downstream models. In some tasks, a single router can distribute its calls across as many as seven different models, reflecting a more flexible and context-aware routing policy. 
(2) In contrast, GPT-based off-the-shelf routers tend to respond directly and thus rarely rely on any downstream models. The Qwen2.5-7B and Qwen3-8B routers display somewhat more variation, yet their downstream calls remain heavily concentrated on GPT-5, which accounts for more than 75\% of their offloading instances. 
(3) Our trained routers, however, show adaptive preferences across tasks, relying on different downstream models depending on the nature of the input. From the statistics, GPT-5, GPT-5-mini, and GPT-OSS-20B emerge as the most frequently selected downstream models, indicating that our routers' decisions are guided by task-specific reasoning rather than simply defaulting to the strongest or most expensive available models.

An additional observation is that nearly all routers, regardless of architecture, tend to favor the GPT-4 and GPT-5 model families as downstream choices, while seldom invoking other open- or closed-source reasoning models such as DeepSeek-R1 or o3-Pro. This behavior may stem from the higher computational cost associated with these models, which discourages their frequent use during training, or from their lower accuracy in certain tasks, which diminishes their reward signal in reinforcement-based optimization. It is also worth noting that our current benchmark focuses on general-purpose reasoning tasks and excludes most domain-specific settings (beyond math and code). Future work may extend both the router and the evaluation suite to include specialized domains such as medicine or chemistry, where different downstream distribution patterns may naturally emerge.
 

\subsection{Key Findings}

\begin{itemize}[topsep=-1pt, partopsep=-3pt, leftmargin=*, itemsep=0pt]
    \item \textbf{End-to-end router training significantly enhances decision quality.}
    Across all benchmarks, the trained {\xrouter} achieve better cost--performance balance across static and untrained router baselines, confirming that end-to-end optimization enables more accurate and context-aware routing decisions than heuristic strategies.

    \item \textbf{Cost-aware reward shaping yields efficient accuracy–cost balance.}
    Varying the cost penalty $\lambda$ demonstrates a trade-off between accuracy and efficiency. Moderate settings (e.g., $\lambda=2$) can achieve near-optimal performance while reducing inference cost by up to 80\%, illustrating the importance of calibrated economic regularization in multi-model orchestration.

    \item \textbf{Learned routing narrows the gap between open and proprietary systems.}
    {\xrouter} models based on mid-sized open-source backbones (e.g., Qwen2.5-7B-Instruct) achieve 80–90\% the accuracy of GPT-5 as off-the-shelf routing model but at only one-fifth of the cost, which redefines the efficiency frontier between open and closed ecosystems for large-scale reasoning.

    \item \textbf{Adaptive routing promotes domain generalization.}
    Without task-specific retraining, {\xrouter} maintains strong performance across heterogeneous domains such as mathematical reasoning, code generation, and general QA. This suggests that learned routing priors enable generalization across task types, even when downstream models vary in specialization.

    \item \textbf{Robustness to changing model pools underscores general adaptability.}
    When the available model pool is expanded with additional models, {\xrouter} can sustain or improve performance. This demonstrates {\xrouter}'s robustness to environmental changes and its ability to reason dynamically over varying resource sets.

    \item \textbf{Routing strategy diversity reflects reasoning maturity.}
    The trained {\xrouter} exhibits a balanced mix of direct and synthesized responses, unlike off-the-shelf models that overwhelmingly favor direct answers. This indicates a higher level of meta-cognitive reasoning—deciding when to solve independently versus when to delegate.

    \item \textbf{Offloaded model usage reveals adaptive specialization.}
    Our routers selectively offload to a diverse set of downstream models, choosing among GPT-5, GPT-5-mini, and various other ones based on input characteristics. In contrast, untrained or static routers display narrow or biased offloading behavior, often over-relying on a single strong model.
\end{itemize}

Overall, these findings collectively demonstrate that {\xrouter} is not merely a cost-control mechanism but a learned decision-making system that optimally balances accuracy, cost, and adaptability. Through dynamic model selection and cost-aware optimization, {\xrouter} advances the practical frontier of large-model orchestration for real-world reasoning applications.

\section{Discussion}
\label{sec:discussion}

Building {\xrouter} has taught us as much about the limitations of current approaches as about their potential. While our system demonstrates that learned routing can work in practice, the path to getting there revealed several uncomfortable truths about the current state of multi-model orchestration research.

\subsection{What We Got Right}

The core insight behind {\xrouter}, that routing decisions should be learned rather than programmed, appears sound. Our cost-aware reward mechanism successfully encourages models to find reasonable trade-offs between performance and expense, avoiding both the profligate use of premium models and the false economy of budget models on complex tasks. The system learns to route mathematical reasoning problems to specialized models while handling straightforward queries with cost-effective alternatives, demonstrating that the basic framework can capture meaningful distinctions.

More surprisingly, the integration challenges we expected across different model providers proved manageable. Despite significant variations in API formats, error handling, and response structures, the abstraction layer we developed successfully unified diverse models into a coherent ecosystem. This suggests that the technical barriers to multi-model orchestration, while non-trivial, are not insurmountable.

The production deployment aspects also exceeded expectations. The system handles the inevitable API failures, timeout issues, and rate limiting problems that plague real-world model deployments. Our distributed training infrastructure scaled reasonably well, and the OpenAI-compatible API integration made deployment straightforward for existing systems. These engineering contributions, while perhaps less glamorous than algorithmic innovations, prove crucial for practical adoption.

\subsection{Where We Fell Short}

However, several aspects of our approach revealed deeper problems that simple engineering cannot solve. The most significant disappointment concerns the sophistication of learned routing behaviors. Despite extensive training and carefully designed reward functions, our router consistently converges to disappointingly simple patterns: analyze the query, pick a model, format the response. The sophisticated orchestration strategies we envisioned, such as dynamic model switching based on intermediate results, intelligent parallel processing, or iterative refinement across multiple models, simply do not emerge naturally from standard RL training.

This limitation appears fundamental rather than incidental. The exploration mechanisms in current RL approaches seem insufficient for discovering complex multi-step strategies when simpler alternatives work reasonably well. The router learns to avoid failure rather than to excel, settling into safe patterns that prevent obvious mistakes while missing opportunities for genuine sophistication.

Our experience with different base model architectures proved equally humbling. The Qwen3-4B model, despite impressive capabilities in isolation, proved remarkably resistant to router training. The model's strong bias toward internal reasoning over tool utilization suggests that architectural choices during pre-training have profound implications for downstream task learning that we poorly understand. The counterintuitive finding that older Qwen2.5 models train more effectively than newer Qwen3 variants challenges common assumptions about model evolution and improvement.

\subsection{Implications for the Field}

These findings have broader implications for how we think about multi-model systems. The apparent trade-off between model sophistication and trainability suggests that the most capable standalone models may not make the best components in orchestrated systems. This creates a tension between the natural progression toward more capable individual models and the requirements for effective multi-model architectures.

The emergence problem, where sophisticated behaviors fail to develop naturally from RL training, points to fundamental limitations in how we approach complex system learning. Simply providing the right incentives appears insufficient; the system needs explicit guidance toward the types of behaviors we want to see. This suggests that future work may need to rely more heavily on supervised demonstrations of complex orchestration patterns before attempting RL refinement.

From a practical deployment perspective, our results suggest that the benefits of multi-model orchestration are real but more modest than initial expectations might suggest. The cost savings and performance improvements are measurable and valuable, but the sophisticated adaptive behaviors that motivate much of the research in this area remain elusive. Organizations considering multi-model deployment should calibrate expectations accordingly.

\subsection{The Cost of Complexity}

Perhaps most soberly, our work highlights the substantial overhead costs of multi-model systems. The engineering complexity of managing multiple API integrations, the computational costs of running routing models, and the operational challenges of monitoring and maintaining distributed model deployments all represent significant investments. For many applications, the benefits of intelligent routing may not justify these costs compared to simply using a capable single model.

This cost-benefit analysis becomes particularly complex when considering the hidden costs of multi-model systems. Training router models requires substantial computational resources and careful data curation. Monitoring system performance across multiple models introduces new classes of failure modes. The complexity of debugging issues that span multiple model providers can significantly increase operational overhead.

The question becomes not just whether multi-model routing can work, but whether it represents the best allocation of engineering and computational resources for most organizations. Our experience suggests that the answer depends heavily on deployment scale, cost sensitivity, and the diversity of workloads being served.

\subsection{Broader Context and Limitations}

Our work necessarily reflects the current state of the LLM ecosystem, which continues to evolve rapidly. The specific model capabilities, pricing structures, and API characteristics that shaped our design decisions may change significantly as the field progresses. The routing strategies that prove optimal today may become suboptimal as new models emerge.

Additionally, our evaluation methodology, while comprehensive within its scope, cannot capture all aspects of real-world deployment. User satisfaction, long-term system reliability, and the complex interactions between cost optimization and user experience remain difficult to quantify but crucial for practical success. The controlled experimental conditions under which we evaluated {\xrouter} may not reflect the messier realities of production deployment.

The generalizability of our findings also remains an open question. Our experiments focused on specific types of reasoning tasks and particular model ecosystems. Different task distributions, alternative model combinations, or novel orchestration requirements might yield different conclusions about the effectiveness of learned routing approaches.

These limitations suggest that while {\xrouter} demonstrates the feasibility of learned multi-model routing, much work remains to understand when and how such approaches provide value in practice. The field would benefit from more extensive real-world deployment studies, longer-term performance evaluations, and systematic investigation of the conditions under which multi-model orchestration justifies its complexity.

\section{Future Directions and Research Insights}
\label{sec:future}

Our development of {\xrouter} has revealed several critical insights that challenge conventional assumptions about multi-model orchestration and highlight promising avenues for future research. These findings emerge from practical experience with training, deployment, and the inevitable failures that accompany real-world system development.

\subsection{Rethinking Model Pool Composition}

One of our most significant discoveries concerns the relationship between model diversity and routing effectiveness. Contrary to initial expectations, expanding the API pool does not necessarily improve performance. Our experiments suggest that focusing on a smaller, carefully curated set of models (particularly within unified model families like the GPT-5 series) may yield superior results compared to diverse multi-provider ecosystems.

This finding has profound implications for system design. The complexity introduced by managing disparate API formats, varying response structures, and inconsistent error handling across providers often outweighs the theoretical benefits of capability diversity. Future work should investigate the optimal balance between model diversity and operational simplicity, with particular attention to how model families with consistent architectures and training methodologies might enable more stable routing behaviors.

The economic implications are equally important. Provider-specific optimizations, bulk pricing agreements, and reduced integration overhead could make focused partnerships more attractive than broad multi-provider strategies. Research into automated model family discovery and expansion within trusted ecosystems represents a promising direction for practical deployment.

\subsection{Fundamental Training Challenges with Modern Architectures}

Our initial experiments with different base router architectures have uncovered deep challenges that go beyond conventional training difficulties. The Qwen3-4B model, despite its impressive standalone capabilities, proved remarkably resistant to router training. The model exhibits a strong bias toward extended internal reasoning rather than tool utilization, suggesting that architectural choices fundamentally influence trainability for agentic tasks.

This phenomenon appears rooted in overfitting to specific reasoning patterns during pre-training. Qwen3 models consistently attempt to solve problems through prolonged internal deliberation rather than leveraging external model calls, even when explicitly trained to do otherwise. Interestingly, Qwen2.5 models demonstrate significantly better tool use behaviors, despite their earlier vintage. This suggests that newer is not necessarily better for router training, and that architectural evolution may inadvertently reduce flexibility for downstream fine-tuning.

The size dependency we empirically observed in initial trials, where Qwen2.5-3B fails while Qwen2.5-7B succeeds, points to critical mass effects in parameter space that enable effective behavior modification. Understanding these thresholds and their relationship to model architecture represents an important area for theoretical investigation. Future work should systematically characterize which model families and sizes are amenable to router training, potentially developing architectural modifications that preserve both reasoning capabilities and training flexibility.

\subsection{The Emergence Problem in Agentic Routing}

Perhaps our most important finding concerns the limited sophistication of learned routing behaviors. Despite extensive training and reward engineering, our router models often converge to simple behavior patterns even though the offloaded model shows diversity. Empirically, more sophisticated agentic strategies such as dynamic model switching and iterative refinement across models do not emerge from standard RL approaches.

This limitation appears to stem from fundamental constraints in how current RL methods explore complex action spaces. The router learns safe, predictable strategies that minimize risk of failure rather than discovering potentially superior but riskier orchestration patterns. Standard techniques like epsilon-greedy exploration or curiosity-driven methods prove insufficient for encouraging the deep exploration necessary to discover advanced multi-model workflows.

Addressing this challenge will likely require incorporating more diverse behaviors directly into supervised fine-tuning data before RL training. By explicitly demonstrating sophisticated routing patterns, including cases where complex orchestration strategies outperform simple selection, we may provide the foundation for RL to build upon rather than expecting such behaviors to emerge spontaneously. This suggests a two-stage approach where diverse synthetic routing episodes train the model to recognize when and how to employ advanced strategies, followed by RL that optimizes these behaviors for specific objectives.

\subsection{Infrastructure Limitations and Future Architectures}

Our current reliance on live API calls during training and inference has proven to be a significant bottleneck. The combination of high latency, occasional failures, and substantial costs creates instability that impedes both research progress and practical deployment. API timeouts and rate limiting become particularly problematic during high-concurrency training, while the financial costs of extensive experimentation can quickly become prohibitive.

A promising alternative architecture involves pre-computing and caching model responses along with associated metadata such as correctness probabilities, computational costs, and quality scores. This approach would enable rapid simulation-based training where reward functions operate on cached results rather than live API calls. Such a system could dramatically accelerate training iterations while providing more stable and reproducible experimental conditions.

The technical challenges of implementing such a cache-based system are substantial but surmountable. Key considerations include developing robust query normalization for cache hits, handling the exponential growth of cached responses, and maintaining cache freshness as models evolve. However, the benefits, including faster experimentation cycles, lower costs, and more reliable training environments, make this a high-priority direction for system development.

\subsection{Beyond Current Reward Mechanisms}

Our experience with cost-aware reward functions has highlighted fundamental limitations in how we formulate multi-objective optimization in routing scenarios. The simple linear combination of performance and cost metrics, while mathematically convenient, fails to capture the complex preferences that emerge in real deployment scenarios. Users rarely have uniform cost sensitivity across all task types, and the relationship between cost and acceptable performance varies significantly with context.

Future work should explore more sophisticated preference modeling that can adapt to user-specific and context-specific trade-offs. This might involve learning personalized utility functions that reflect individual cost sensitivities, developing contextual reward models that adjust expectations based on task complexity, or implementing multi-objective optimization approaches that can navigate Pareto frontiers dynamically.

The theoretical foundations for such approaches remain underdeveloped. Research into preference learning for multi-model systems, adaptive reward mechanisms that evolve with user behavior, and robust optimization techniques that maintain performance across diverse preference profiles represents an important frontier for both academic investigation and practical development.

\subsection{Towards Systematic Evaluation and Benchmarking}

Our work has underscored the absence of standardized evaluation frameworks for multi-model routing systems. Existing benchmarks focus on individual model capabilities rather than orchestration effectiveness, making it difficult to compare approaches or track progress in the field. The complex interplay between cost, performance, reliability, and user satisfaction requires more sophisticated evaluation methodologies than traditional accuracy metrics can provide.

Developing comprehensive benchmarks for routing systems requires addressing several challenges. These include creating realistic task distributions that reflect actual deployment scenarios, establishing fair cost models that account for varying provider pricing structures, and designing evaluation protocols that capture long-term behavior rather than single-turn performance. Future work should prioritize building shared evaluation infrastructure that enables rigorous comparison of routing approaches across different model ecosystems and deployment constraints.

This benchmarking challenge extends to the fundamental question of what constitutes success in multi-model orchestration. Traditional metrics like accuracy or cost efficiency may miss important aspects of user experience, system reliability, or long-term behavior. Research into holistic evaluation frameworks that capture the full spectrum of routing system performance is essential for guiding future development and enabling meaningful comparison across approaches.

\printbibliography[title={References}]



\end{document}